\documentclass[10pt,twocolumn,letterpaper]{article}

\usepackage{cvpr}
\usepackage{times}
\usepackage{epsfig}
\usepackage{graphicx}
\usepackage{amsmath}
\usepackage{amssymb}
\usepackage{graphicx}  \graphicspath{{figures/}}
\usepackage{algorithm}% http://ctan.org/pkg/algorithms
\usepackage{algpseudocode}% http://ctan.org/pkg/algorithmicx
\usepackage[normalem]{ulem}
\usepackage{makecell}
\usepackage{rotating}
\usepackage{booktabs}
\usepackage{gensymb}
\usepackage{enumitem}
\usepackage[toc,page]{appendix}
\usepackage[dvipsnames,svgnames,x11names]{xcolor}
\newcommand{\norm}[1]{\left\lVert#1\right\rVert}
\usepackage[pagebackref=true,breaklinks=true,letterpaper=true,colorlinks,bookmarks=false,citecolor=blue,linkcolor=blue]{hyperref}
\usepackage{multirow}
\usepackage{authblk}

\makeatletter
\renewcommand\AB@affilsepx{, \protect\Affilfont}
\makeatother

\cvprfinalcopy % *** Uncomment this line for the final submission

\def\cvprPaperID{100} % *** Enter the CVPR Paper ID here

\ifcvprfinal\fi
\begin{document}

%%%%%%%%% TITLE
\title{Putting Humans in a Scene:
Learning Affordance in 3D Indoor Environments}
%Learning 3D Human Affordance in Indoor Environment

\author[1]{Xueting Li\thanks{This work is completed during an internship at NVIDIA.}}
\author[2]{Sifei Liu}
\author[2]{Kihwan Kim}
\author[3]{Xiaolong Wang}
\author[1]{Ming-Hsuan Yang}
\author[2]{Jan Kautz}
\affil[1]{University of California, Merced}
\affil[2]{NVIDIA}
\affil[3]{Carnegie Mellon University}

\maketitle
%\thispagestyle{empty}

%%%%%%%%% BODY TEXT
%%%%%%%%% ABSTRACT
\begin{abstract}
%\JK{not entirely happy with it yet}
Affordance\footnote{Affordances are opportunities for interactions in a scene or environment. It represents what interactions an environment could provide for humans, e.g., a chair provides the opportunity to sit.} modeling plays an important role in visual understanding.
In this paper, we aim to predict affordances of 3D indoor scenes, specifically what human poses are afforded by a given indoor environment, such as sitting on a chair or standing on the floor.
In order to predict valid affordances and learn possible 3D human poses in indoor scenes, we need to understand the semantic and geometric structure of a scene as well as its potential interactions with a human.
To learn such a model, a large-scale dataset of 3D indoor affordances is required.
In this work, 
we build a fully automatic 3D pose synthesizer that fuses semantic knowledge from a large number of 2D poses extracted from TV shows as well as 3D geometric knowledge from voxel representations of indoor scenes.
With the data created by the synthesizer, we introduce a 3D pose generative model to predict semantically plausible and physically feasible human poses within a given scene (provided as a single RGB, RGB-D, or depth image).
We demonstrate that our human affordance prediction method consistently outperforms existing state-of-the-art methods. The project website can be found at \url{https://sites.google.com/view/3d-affordance-cvpr19}.
\end{abstract}

\begin{figure}[t]
\centering
\includegraphics[width=1\linewidth]{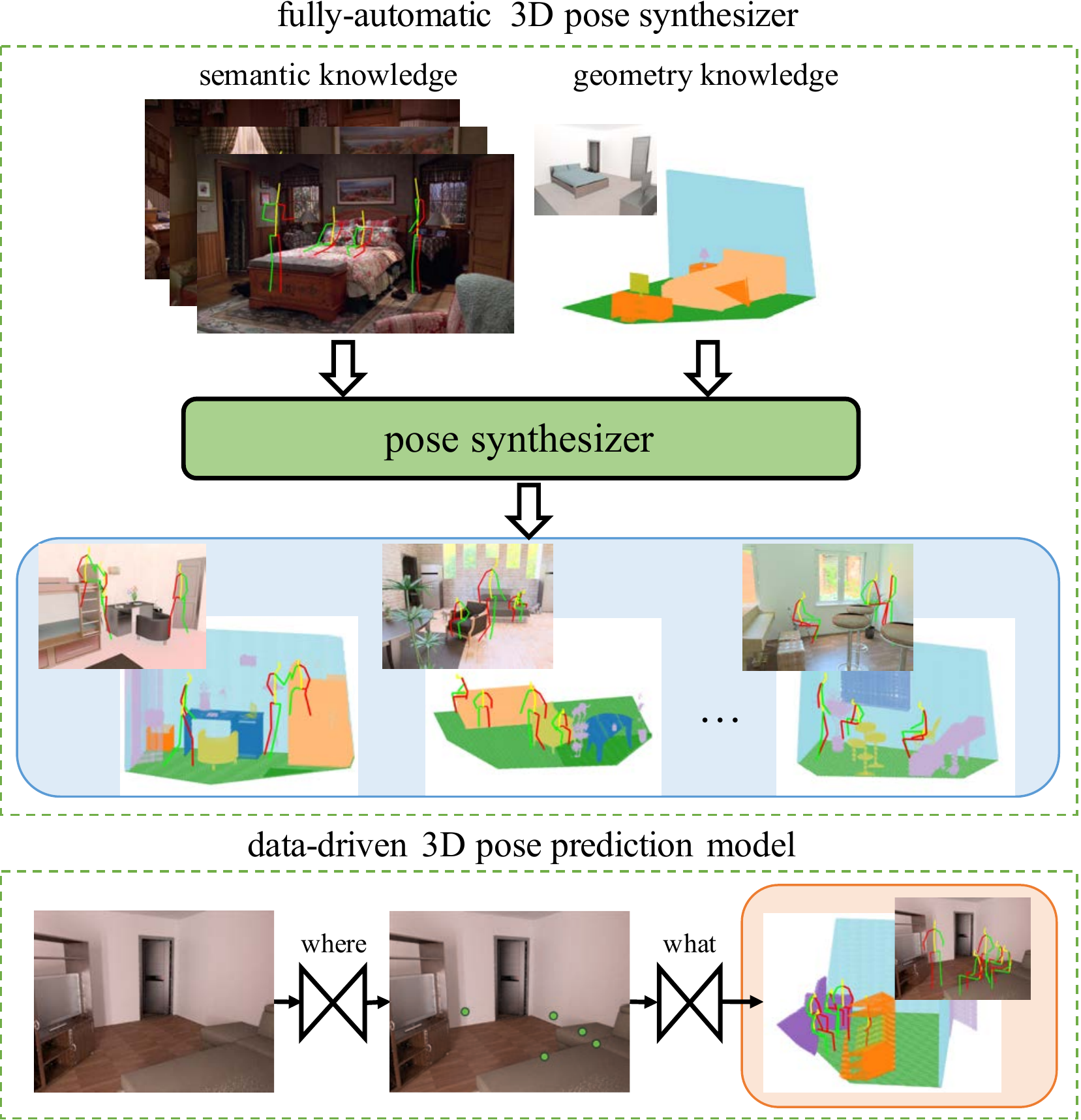}
\caption{\textbf{Overview of the proposed method.} 
Our method contains two stages. First, we propose a fully-automatic 3D pose synthesizer, which can synthesize an infinite number of 3D poses for indoor scenes (see Section~\ref{sec:poseproduce}). We illustrate synthesized pose samples in the \emph{light blue box}.
Second, we learn an end-to-end 3D affordance prediction model by jointly learning the distribution of locations and 3D poses (see Section~\ref{sec:3daffordance}). We show generated poses in the \emph{light orange box}. Zoom-in to see details.
}
\label{fig:overview}
\end{figure}

\vspace{-5mm}
\section{Introduction}
There is a long history of studies on functional reasoning of objects and scenes. Instead of focusing on the semantics of objects and scenes, Gibson proposes the idea of affordances~\cite{gibson79}, which can be seen as the ``opportunities for interactions'' with the environment. %The definition of affordance opens a door for researchers to use 3D geometry, scene layouts and functions to learn possible affordances.
%perceive the utilization of space and the meaning of objects. 

To infer the affordances of objects and scenes, researchers have studied the explicit modeling of physical interactions and contacts between human and the 3D scene through simulations~\cite{yixinzhu2016,qi2018human,grabner2011makes}. 
For example, Zhu et al.~\cite{yixinzhu2016} explicitly model sitting styles by inferring the forces and pressures from the interaction between humans and objects in a scene. 
However, explicit modeling suffers from the problem of generalization for other types of poses.
%
%However, the explicit modelling suffers from the problem of generalization for other types of poses such as ``opening a door'' and ``lying down on the bed''.
% such as \emph{opening a door} and \emph{lying down on the bed}.
%
%We might need to have different models for other cases like ``opening a door'' and ``lying down on the bed''. 
%The gap between current physics engine and the real physical world is negligible as well.  \khnote{This sentence supposed to be "not negligible?" In either case, not looking important. Add it if necessary later.}
%
To tackle the problem of generalization, researchers have proposed to directly infer affordances in a data-driven manner~\cite{FouheyDirect2015,fouhey2014people,Wang_affordanceCVPR2017}. Specifically, Wang et al.~\cite{Wang_affordanceCVPR2017} design a method to collect human-scene interactions by processing video frames of various TV shows and train CNNs for affordance reasoning. Though the method is able to generate semantically plausible human poses aligned with scene images, it is not able to follow the geometry of the 3D world and often produces results violating physics (e.g., first row in Fig.~\ref{fig:whatCom}) due to a lack of 3D geometric information of the scenes (as the data consists only of video frames and 2D poses).
%Given a reasonable amount of data, they are able to train ConvNets for affordance reasoning. They show that the ConvNets are able to generalize to new scenes and generate novel interactions. However, the data collection process still involves human labeling, and they can only collect limited training examples (around 20K). As the real world data follows the long-tailed distribution, the ConvNets are not fully able to capture interesting human behaviors from the TV series dataset. Without sufficient data, it is also hard for the ConvNets to follow the geometric constraints of the scene and often produce results violating the physics. 

% the approach should tackle these two questions: long-tailed distribution of interesting behaviors, and 3D geometric constraints

% put interesting interaction figures in teaser

% \JK{we need to be more clear on what the goal is and how we go from existing 2D models to 3D, and then learn another model from that. It's currently vague and confusing. Use Fig.~1 to explain.}

In this paper, our goal is to learn a model that is able to generate 3D human poses that not only follow natural human behaviors (e.g., humans should sit rather than stand on a chair), but also are physically feasible (e.g., humans should not collide with objects).
To achieve this goal, we need to synthesize an appropriate dataset containing human poses in various indoor scenes.
We first train a 2D pose prediction model using an existing real-world video dataset~\cite{Wang_affordanceCVPR2017}.
% , by predicting the possible pose locations, and generating poses at the predicted locations.
%
The trained model is then adapted to the indoor images in the SUNCG dataset~\cite{song2017semantic,zhang2017physically}, which contains complete 3D annotations, e.g., camera parameters and 3D geometry (we use a voxel representation).
Since there exist well-defined links between the 2D images and the 3D world, given these annotations, we can map the generated 2D poses into the 3D world.
%Since human height can be approximated as a Gaussian distribution and while the height of a pose in the image coordinate is known, the generated poses can be easily mapped into the corresponding 3D voxels.
% Since the height of a pose in the pixel coordinate system is known, we are able to map it into the corresponding 3D voxel by sampling real-world human height from a Gaussian distribution.
%
% The pose in the voxel is then further adjusted via a simple 3D correlation operator to its nearest feasible locations. 
We further adjust these mapped poses in 3D voxel space to make sure they are physically feasible (no intersections with objects and well supported by surrounding furniture).
Our dataset synthesis approach is fully automatic and can synthesize numerous, diverse ``ground-truth'' poses in different locations.

Given this large amount of data, we are able to train an affordance prediction model, which aims to generate 3D human poses given a single scene image. 
We model the pose distributions conditioned on the scene context, where the pose distributions are factorized into the distributions of (a) pose pelvis joint locations, and (b) pose appearance on top of sampled locations.
We name them the \emph{where} and \emph{what} modules, respectively.
The two modules are jointly trained using the pose pelvis joint locations as a differentiable bridge. 
Essentially, we propose a geometry-aware discriminator to encourage the model to better understand the geometry of the scene (see Fig.~\ref{fig:arch} (b)), even through a single RGB image.
%
% We evaluate our model according to both semantic plausibility -- whether the generated poses look like a real pose, and affordance consistency -- whether the poses are physically feasible.
%We evaluate the semantic plausibility of our generated pose by mapping them back to an RGB image,
We evaluate the plausibility of our generated 3D poses via user study as well as a trained classifier that aims to score the ``authenticity'' of generated poses. 
% \xlnote{``classifer'' is better?}
%
We also map generated poses back to the 3D voxel space to evaluate their physical correctness in the 3D world. 

%To this end, we present a complete scheme from automatically synthesizing 3D poses, to learning of a pose generator.
Our main contributions can be summarized as:
\begin{itemize}
    \itemsep0em 
    %\vspace{-2mm}
    \item We propose an efficient, fully-automatic 3D human pose synthesizer that leverages the pose distributions learned from the 2D world, and the physical feasibility extracted from the 3D world.
    %\vspace{-2mm}
    \item We develop a generative model for 3D affordance prediction which generates plausible human poses with full 3D information, from a single scene image.
    %\vspace{-2mm}
    \item We set a new benchmark for large-scale human-centric affordance prediction on the SUNCG dataset by leveraging the human pose synthesizer and the pose generator.
    % Our affordance prediction method outperforms the existing state-of-the-art affordance prediction methods w.r.t both semantic plausibility and physical feasibility evaluations.
    % %
    % Specifically, it can predict poses that are significantly more geometrically feasible than those models only based on 2D information, even with RGB images as the input.
\end{itemize}

 \vspace{-1mm}
 \section{Related Work}
 \vspace{-1mm}
\label{sec:related}
%\khnote{Added temporary references in the bib: feel free to remove or re-organize it.}\\
%~\cite{Hoiem08} Predict possible locations of pedstrain and car to assist detection.

\noindent {\bf Scene understanding.} In recent years, much progress has been made~\cite{xiao2018unified,chen2018deeplab,zhao2017pyramid} in the field of semantic scene understanding thanks to large-scale labeled datasets~\cite{zhou2017scene,lin2014microsoft}. A few methods~\cite{gkioxari2017interactnet,yao2010modeling,mallya2016learning} aim to specially model human-scene interactions. However, they focus on detecting human-object interactions rather than explicitly reasoning about object functionality in a scene.

\vspace{0.2mm}
\noindent {\bf Object functionality reasoning.} For deeper reasoning of objects in a scene beyond the conventional scene understanding techniques, several approaches~\cite{grabner2011makes,yixinzhu2016,zhao2013scene,zhu2015understanding} revisit the principle of affordance~\cite{gibson79} via explicitly modeling the functionality of objects in a scene. For instance, Grabner et al.~\cite{grabner2011makes} propose to detect a chair by considering its functionality (i.e. examining whether an imaginary human can sit on the object). Zhu et al.~\cite{zhu2015understanding} recognize tools and infer their functionality by analyzing RGB-D videos. However, these methods are hard to generalize to real-world scenarios because they rely heavily on complete 3D geometry information of a scene.
% or lack large-scale affordance dataset. 
% \khnote{This last sentence needs to be clearer.}
%\slnote{we should change Zhu's work to sth else since it is mentioned in sec 1.}\\

\vspace{0.2mm}
\noindent {\bf Human affordance prediction.} 
% Koppula13: learn object affordance by watching people
Other than explicitly modeling object functionality, several recent algorithms~\cite{koppula2014physically,cychuang2017learning,zhu2014eccv,Koppula13} exploit human affordance in a data-driven manner. Gupta et al.~\cite{GuptaSatkinEfrosHebert_CVPR11} manually associate human actions
with exemplar poses and search feasible locations for those actions in a scene by performing 3D correlation between poses and scene voxels.
%Gupta et al.~\cite{GuptaSatkinEfrosHebert_CVPR11} estimate human scene interaction by predicting the ``workspace'' of humans in a scene, i.e. the space where a human can perform certain physical actions.
Fouhey et al.~\cite{fouhey2014people} propose to estimate human-scene interactions and scene geometry by observing human actions in time-lapse sequences. Roy and Todorovic~\cite{roy2016multi} predict affordance segmentation maps for specific actions from single images by predicting and fusing mid-level visual cues.
Wang et al.~\cite{Wang_affordanceCVPR2017} collect human-scene and human-object interactions by scanning through millions of video frames in different TV series and train CNNs for human affordance reasoning, which partly motivated our work. However, the data collection process still requires manual effort, and can only collect limited training examples ($\sim$20K). Without sufficient data and geometric knowledge of scenes, it is hard for CNNs to follow the geometric constraints of a scene, leading to results that often violate the physics.

\vspace{0.2mm}
\noindent {\bf Instance placement in a scene.} Our affordance prediction method which puts humans into feasible locations in a scene can be seen as an instance placement task. Several recent approaches~\cite{lin2018stgan,ouyang2018pedestrian,donghoon2018} focus on predicting either location or appearance of an instance in a scene. For example, Lin et al.~\cite{lin2018stgan} propose to insert objects into feasible locations in a scene. However, this method requires a user provided template as the instance. Ouyang et al.~\cite{ouyang2018pedestrian} utilize a Generative Adversarial Network to in-paint pedestrians at given locations in a scene. Closest to our work, Lee~\cite{donghoon2018} jointly model a context-aware distribution of the location and shape of object instances given a scene. Nevertheless, their method focuses on inserting instances in 2D images and does not consider any physical feasibility in 3D scenes.  

\vspace{-2mm}

% In contrast to previous works~\cite{Wang_affordanceCVPR2017,fouhey2014people,GuptaSatkinEfrosHebert_CVPR11}, we aim to learn the interactions between human and \emph{3D scenes} in a \emph{purely automatic} and \emph{fully data-driven} manner. Our end-to-end model, which is trained on millions of synthesized data, is able to generate poses in 3D scenes that follow both semantic and geometry rules based on various input modalities such as RGB/RGBD images.

\section{3D Pose Synthesis}
\label{sec:poseproduce}
\vspace{-2mm}
\begin{figure}[th]
\centering
\includegraphics[width=1\linewidth]{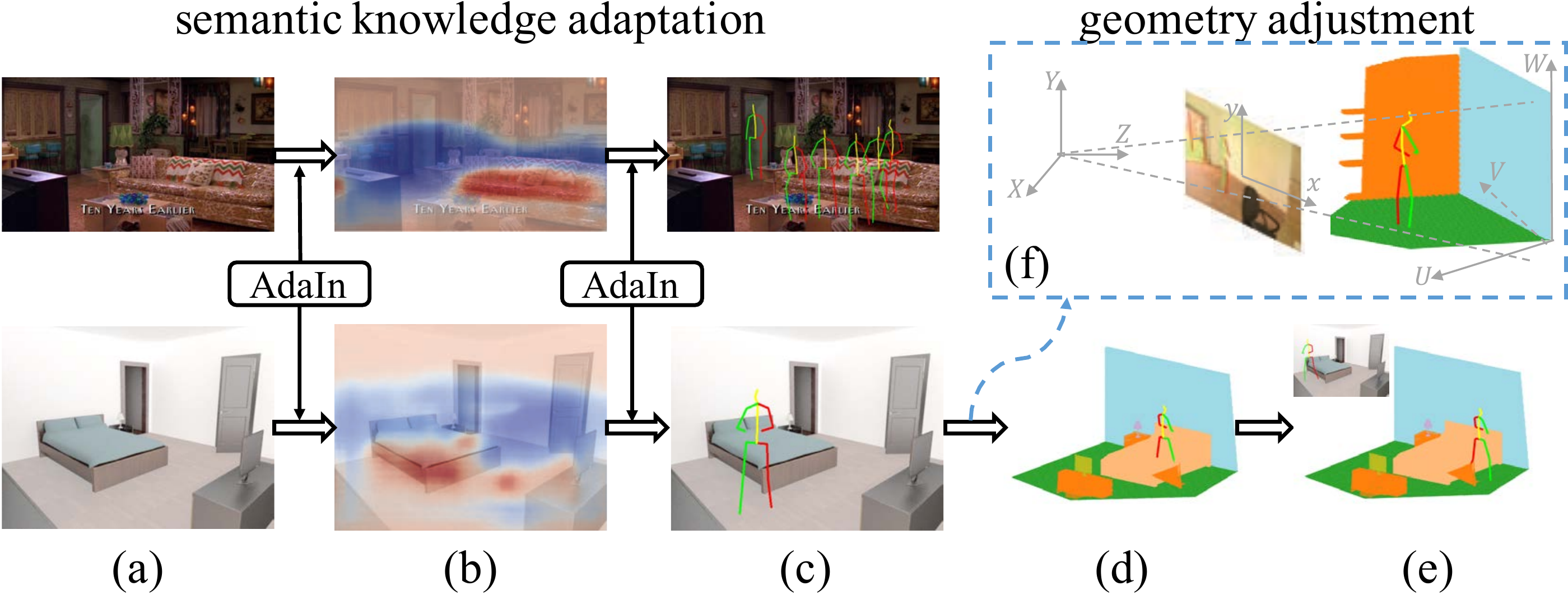}
%\caption{\textbf{Pose synthesis.} (a)$\sim$(c): Adapting semantic knowledge from Sitcom to SUNCG (Section~\ref{sec:2daffordance}). The blue and red regions in column (b) denote the areas suitable for standing and sitting, respectively.
%(d) and (e): Pose adjustment in voxels (Section~\ref{sec:3Dgc}). (f): Mapping the pose from pixel coordinate system to scene voxel (Section~\ref{sec:mapping}). \khnote{This still look really small and hardly read the caption inside the figure.}}
\caption{\textbf{Pose synthesis.} (a) Input image. (b) Location heat map. The blue and red regions denote the areas suitable for standing and sitting. (c) Generated pose. (d) Corresponding pose in the voxel. (e) Adjusted pose in voxel. (f) Mapping from image to voxel.}
\label{fig:stageI}
\vspace{-.2cm}
\end{figure}

%\khnote{Either adding overview of the paper structure (pointing Fig.1) or explaining what we are going to tell in each sectoin in the end of Section 1?}
%\subsection{Overview}
%We now describe how to learn the 3D human affordance for indoor scenes. Similar to Wang et al. ~\cite{Wang_affordanceCVPR2017}, we define the 3D affordance learning task as estimating the potential locations of human models in a given indoor scene, and generating plausible poses of them, which are aligned well with the context of the scene (e.g., sitting on a couch in the room). Training such a generative model requires a large number of annotated human poses in specific 3D scenes. However, such large-scale data set neither exists nor is trivial to annotate. 
%
%In this paper, we separate our framework into two stages. In the first stage we build a human pose generation engine that can create a vast number of human poses given a 3D scene, as shown in Figure~\ref{fig:overview} (Top). In stage two, we train an end-to-end generative model with data sampled from the 3D pose generation engine, as shown in Figure~\ref{fig:overview} (Bottom).

%\subsection{3D Pose Generation Engine}
% Manually annotating human poses in 3D scenes requires a large amount of human efforts, thus we opt t?o automatically generate ground truth human poses in 3D scenes. 
%
Collecting a large-scale dataset of human poses with 3D scene annotations is currently a tedious task~\cite{VirtualHome2018}.
%
% One one hand, it requires professional video capturing facilities, environments as well as actors to provide the data source.
% On the other hand, intensive annotating and pose-processing work is needed.
% %
% In addition, such data collection method may suffer from the problem of generalization.
%
In this section, we show how to automatically synthesize ``ground-truth'' 3D human poses in various indoor scenes. % from an existing 2D pose dataset \cite{Wang_affordanceCVPR2017}. 
To ensure the correctness of generated poses, we take two factors into account:
%
%The synthesizing process is constrained by two factors:
(i) semantic plausibility; the synthesized poses should follow natural human behaviours in typical indoor environments, and 
(ii) physical correctness; the human poses should not collide with objects in a scene or float in the air.
%
%In Section~\ref{sec:2daffordance} and Section~\ref{sec:mapping}, we show how to learn the poses from existing 2D examples~\cite{Wang_affordanceCVPR2017} (see Figure~\ref{fig:stageI}(a)$\sim$(c))  to satisfy constraint (i) and demonstrate how to map the generated 2D poses into the 3D scene (see Figure~\ref{fig:stageI}(f)).
%Then, in Section~\ref{sec:3Dgc}, we introduce an efficient way to adjust the pose in 3D scene (represented as voxels) to satisfy constraint (ii) (see Figure~\ref{fig:stageI}(d)$\sim$(e)).
To satisfy constraint (i), we learn a 2D human pose generative model that encodes the natural human pose distributions from existing 2D examples~\cite{Wang_affordanceCVPR2017} (see Fig.~\ref{fig:stageI}(a) to (c) and Section~\ref{sec:2daffordance}). 
Then, given the camera parameters, we map the generated poses into the 3D world represented as voxels. (see Fig.~\ref{fig:stageI}(f)) and Section~\ref{sec:mapping}). 
Finally, we introduce an efficient way to adjust the poses in the 3D scene to satisfy constraint (ii) (see Fig.~\ref{fig:stageI}(d) to (e) and Section~\ref{sec:3Dgc}).
%Then, we perform efficient search in the voxel to guarantee the physical correctness by further adjusting the pose, as introduced in Section~\ref{sec:3Dgc}. 
%

Overall, we use our pose synthesizer to produce around $1.5$ million ``ground-truth'' poses, which are then used in Section~\ref{sec:3daffordance}. 
%
%Note that we provide not only the generated dataset, but also a creation tool to further generate more examples in variouis indoor scenes. 
Fig.~\ref{fig:overview} (light blue box) shows samples of poses obtained by our pose synthesizer in 3D space and their projections onto 2D images. 

\vspace{-1mm}

\subsection{Affordance Prediction in 2D Scene Images}\label{sec:2daffordance}
%\slnote{I would suggest to reduce this part to "Finding potential locations for humans" only. The others are not important.}
% \slnote{we need to discuss the subsection title.}
% \JK{This is convoluted in terms of what happens in 2D vs. 3D. We need to state somewhere early on exactly what is input and output at each stage. Fig.2 sort of shows it, but in this section it's not clear.}
We synthesize 3D human poses by first generating poses in 2D images, then projecting them into the 3D world as shown in Fig.~\ref{fig:stageI}. 
%This is based on the well-defined links between the 2D images and the 3D world.
%\khnote{This sentence could be vague..}
%
%For example, since human height varies between a small range, it is easy to estimate how far away a person is from images given corresponding camera parameters.
%d
%In this work, we use SUNCG dataset~\cite{song2017semantic} which offers the camera parameters, images and 3D models that are used for rendering the images, and generate poses for the dataset. With the known camera parameters and scene models, we can roughly estimate the potential distance of human in the scene because the potential human height lies in a certain range especially in the indoor scene.
%Generating semantically correct poses for the 3D scene data is not trivial. 
%Note that pose samples captured from videos (sitcom or movies) are already semantically correct sources for the 3D human poses.
%Because such operation requires camera parameters and 3D annotations (e.g. voxels) corresponding to the images, we choose to use the SUNCG dataset~\cite{song2017semantic}.
% Though our ultimate goal is to model human affordance in 3D scenes, the lack of existing 3D human affordance data motivates us to first explore the affordance learning in 2D scenarios and then adapt the learned affordance model into 3D scenes using rendered images and camera parameters as a link. 
%How can we generate human poses on the SUNCG synthetic images while there is no pose data contained in the dataset?
%
To this end, we utilize the Sitcom dataset~\cite{Wang_affordanceCVPR2017} which contains pose samples captured from sitcom videos and train a human pose prediction model. Then we adapt the trained model onto the SUNCG images to generate poses that follow natural human behaviors.
The work by Wang et al.~\cite{Wang_affordanceCVPR2017} only focuses on predicting the most plausible human pose at a feasible location in 2D scene images. 
However, the annotations of such feasible locations are not available in the SUNCG dataset.
Therefore, we need to learn a network that predicts locations to put humans in a scene, before utilizing the method in~\cite{Wang_affordanceCVPR2017} to generate human poses at each predicted location.
%\khnote{I still had hard time to edit this section, please review if it is okay or not. I tried as explicit as possible.}
%\noindent \textbf{Finding potential locations for humans.}
%
% We represent the location of human as the position of pelvis joint denoted by the coordinates $(x,y)$.
% Learning a model that predicts possible pose locations in a scene is a challenging task. 
% %
% Because there are multiple valid poses that can be plausible at the same predicted location. Also, existing pose annotations in 2D scenes~\cite{Wang_affordanceCVPR2017} are highly sparse (typically only few poses per scene). 

We represent each pose location by its pelvis joint coordinates. A typical technique~\cite{roy2016multi} for predicting human pose locations is to learn a pixel-wise probability map of a scene. 
%
%However, the existing 2D pose annotations are highly sparse (typically only few poses per scene). Besides, the pose prediction model in~\cite{Wang_affordanceCVPR2017} also depends on pose class information at each pose location.
%
%To address these issues, we first assign each pose to one of 20 pose clusters obtained by k-mediod clustering over all poses and represent pose class by a 20 dimensional one-hot vector. We further augment such sparse  annotations by extending pose class label at a single point to a local square patch, assuming nearby areas can afford same poses. %\khnote{set the annotated pose classes to one -- correct?}
%
%
However, the existing 2D pose annotations are highly sparse (typically only a few poses per scene). To address this issue, we augment the annotation from a single point to a local square patch, assuming the nearby area can afford the same pose.
%
%Furthermore, the pose prediction model in~\cite{Wang_affordanceCVPR2017} also requires pose class information at each pose location. 
%
%Thus we first assign each pose to one of 20 pose clusters obtained by k-means clustering over all poses. \JK{clustering comes out of nowhere}
%Thus we first cluster all poses into 20 clusters by the k-means method and assign each pose to one of these clusters, we take the cluster index as the pose class label.
Furthermore, Wang et al.~\cite{Wang_affordanceCVPR2017} cluster all poses into 30 clusters according to their gestures and feed the cluster center corresponding to each pose as a condition to their pose prediction model.
Thus to utilize their pose prediction model, we not only need to find feasible locations for human poses, but also predict the most likely pose class at each predicted location.

To this end, for each location that has a pose annotation, we use a 31-dimensional binary vector to represent the corresponding pose class. Locations without pose annotations are labeled as background (the 31$^\mathrm{st}$ class).
This results in a $31\times h \times w$ pose location map as the ground truch heatmap for each scene, where $h$ and $w$ are the height and width of the scene image. 
%This results in a pose location heat map for each scene, with each pixel indicating pose class suitable for this location (or background if the location is not suitable for any pose class).
%
We learn a CNN that takes a scene image as input and predicts the corresponding heat map.
During the testing process, we sample from the heat map and output both locations possible for human poses as well as the most likely pose class at these locations. 

% \slnote{I think we can move the 2D to 3D pose mapping to here.}
% \xlnote{why we need to do this}
Since our ultimate goal is to generate 3D poses, we first map 2D pose annotations in the Sitcom dataset to 3D poses in the Human3.6M dataset~\cite{Ionescu36mpami} and then train the pose generation model in~\cite{Wang_affordanceCVPR2017} to generate 3D poses. Detailed mapping process can be found in the appendix.
In this way, we extend the pose prediction in~\cite{Wang_affordanceCVPR2017} from generating 2D poses in given ground truth locations, to generating 3D poses at sampled locations. 
Fig.~\ref{fig:stageI}(b) and (c) illustrate location heat maps and poses predicted by our model respectively.

To narrow the domain gap between the SUNCG and the Sitcom dataset, we perform domain adaptation~\cite{huang2017adain}
when applying the trained model onto the SUNCG images, via matching the second-order statistics of image features for both location and pose prediction models. More details about the domain adaptation can be found in the appendix. \\
\vspace{-5mm}
\subsection{Mapping Poses into 3D Scenes} 
\label{sec:mapping}
%Due to limited data and scene geometry information, it is hard for the models trained on 2D scenes to follow the geometry constraints in a scene so it often produces poses violating the physical constrain (e.g. poses collide with objects in the scene). 
%To guarantee the geometry feasibility of generated poses, we map them into 3D scenes represented by voxel cubic to perform geometry adjustment described in the next section. 
% Since the generative model originally trained from 2D scenes do not have access to scene geometry information, it may generate poses collide with surrounding objects in a scene. Thus we map generated poses back into 3D scene voxels and by moving or rotating them, we guarantee the feasibility of the 3D poses in a scene. We denote this process as \enquote{geometry adjustment} and describe in details in the next section. 
Mapping a pixel from the image coordinates to the 3D world requires its depth value and the camera parameters. %and its depth value of the corresponding point in the 3D world. \xlnote{Please check this sentence. I don't know if it should be ``the depth value of a point'' or ``the depth value of a surface''.}. 
Unfortunately, depth values are not known for the generated human poses. However, we circumvent this problem by estimating these depth values from the known real-world distribution of human heights.
%
%However, the depth values are not known for the generated human poses.
%
%But fortunately, it can be estimated since we know how tall a human might be in the real world!
%
We sample the height of a human for standing pose from $\mathcal{N}(1.65,0.1)$, and for sitting pose from $\mathcal{N}(1.20,0.1)$.
% We map each generated pose into the 3D world by finding the location of each joint in the 3D world given its coordinates in the 2D image and camera parameters of the scene. 
% %
% %Note that direct mapping of an object into 3D scene requires depth information, which is not available for the generated 2D poses.
% Mapping a joint from 2D image into the 3D world requires not only its coordinates in the 2D image but also its depth value. However, depth information is not available in our generated poses. So how can we map each joint into the 3D world? 
% Fortunately, real human height in 3D world lies within a small range of average human height and can be sampled from a Gaussian distribution (e.g. human height for standing pose can be sampled from $\mathcal{N}(1.65,0.1)$). 
Given the sampled human height in 3D world, we can estimate the depth $d$ of each pose by $d = \frac{H \times f}{H_p \times r_{32}}$, where $H_p$ is the pose height at pixel coordinate system, $H$ is the sampled human height mentioned above, $f$ is focal length and $r_{32}$ is a specific parameter in camera extrinsic matrix. A detailed derivation is available in the appendix. 
Fig.~\ref{fig:stageI}(f) illustrates the mapping process. We take the resulting pose depth as the depth of the pelvis joint and calculate the depths of other joints by their offsets w.r.t. the pelvis joint. Then, we map each joint into the 3D world using intrinsic and extrinsic camera matrices. %, given the camera parameters.

\subsection{Affordance Constraint in the 3D World}
\label{sec:3Dgc}
\vspace{1mm}
%\slnote{Let's reduce this part.}
% The heat map and pose prediction models trained on Sitcom is solely based on the interaction between human and 2D scenes. Therefore, even when a pose appears plausible in a 2D image, it may violate geometric rules when mapped into 3D scenes. e.g. a human floating in the air. 
% %
% To insure the geometric feasibility of generated poses in 3D scenes, we map them into 3D scenes and adjust pose locations. % to guarantee their geometry correctness. 
% %
% For instance, if the generated human sits slightly above the surface of a sofa, with the assistance of geometry information, we are able to adjust the pose location of this human so that the human can sit comfortably on the surface of a sofa.

% To make geometric constraint specific,
\begin{figure}[t]
\centering
\includegraphics[width=1\linewidth]{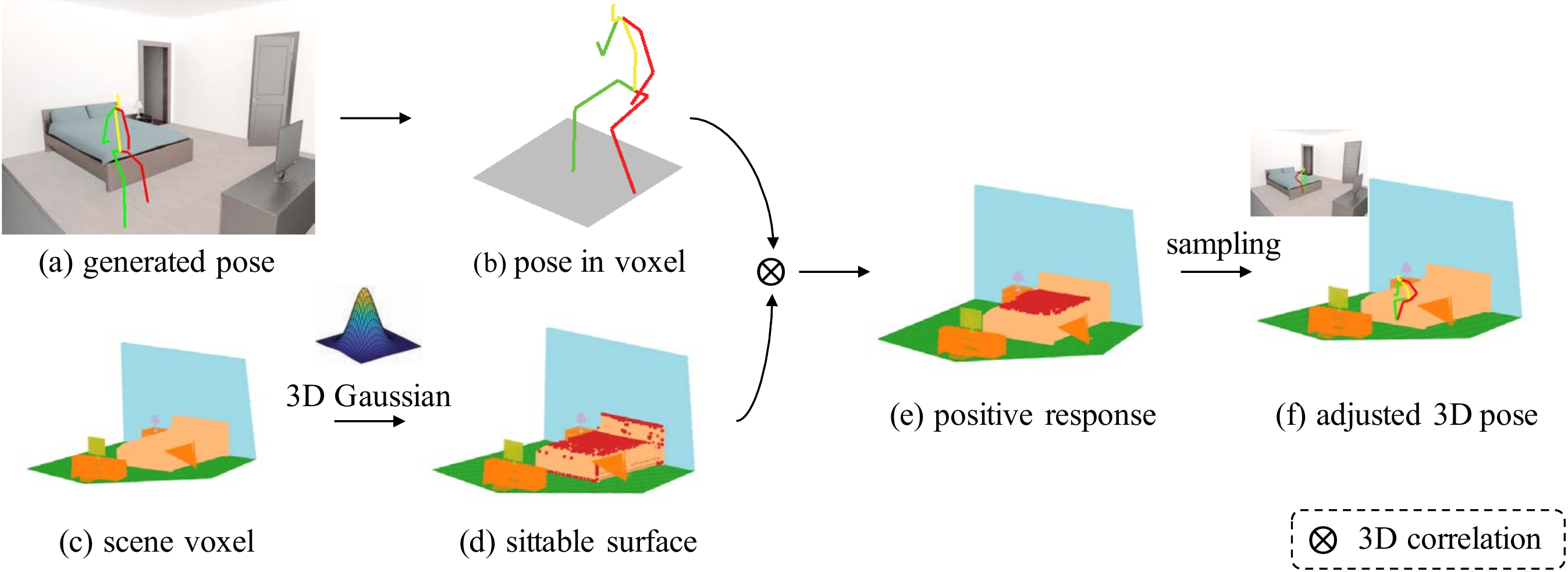}
\caption{\textbf{Affordance adjustment.} (a) Generated pose by the model described in Section~\ref{sec:2daffordance}. (b) Corresponding pose in the voxel space. (c) A scene voxel. (d) The surface of a bed (colored in red) detected by a 3D Gaussian kernel. (e) Positive responses indicating locations suitable for the given sitting pose (colored red). (f) Adjusted pose at the location with the highest positive response.}
%For each generated pose, we first perform 3D correlation between scene voxel containing only sittable objects (chair, sofa etc) with a 3D Gaussian kernel to generate sittable surface (as shown in the right area in \enquote{sittable surface}). Then we perform one more 3D correlation between voxelized sitting pose with sittable surface. Red area in Positive Response shows 10000 points with highest response. Next we select a location with highest response and move the predicted pose to this location as well as adjust human orientation and scale according to location information.
\label{fig:gsearch}
\end{figure}
\vspace{-2mm}

Since the pose prediction model is trained with only 2D information, a plausible generated pose may not be physically feasible when mapped to 3D, e.g., the pose collides with the bed as exemplified in Fig.~\ref{fig:stageI}(d).
%e.g. a human floating in the air.
%
Therefore, we adjust it locally to make the pose physically feasible.
For example, we can adjust pose locations to avoid collision as shown in Fig.~\ref{fig:stageI}(e), or adjust a sitting pose right onto the surface of a bed as shown in Fig.~\ref{fig:gsearch}(f). 

%Specifically, we ensure the physical feasibility of poses by obeying the \emph{free space constraint} and \emph{support constraint} described in ~\cite{GuptaSatkinEfrosHebert_CVPR11}. To this end, we perform 3D correlations between scene voxel and poses to find suitable locations for the generated poses with no manual annotation needed. \\
The method by Gupta et al.~\cite{GuptaSatkinEfrosHebert_CVPR11} manually associates each action
with an exemplar pose and searches locations valid for the pose by satisfying the \emph{free space constraint} and the \emph{support constraint}. 
However, such a manual solution is not feasible in our case since our poses are generated, rather than selected from a set of fixed poses. 
We explain next how to extend the method in~\cite{GuptaSatkinEfrosHebert_CVPR11} to search for locations satisfying both constraints in an efficient and fully-automatic manner.
%
%
%Specifically, we adopt the constraints described in ~\cite{GuptaSatkinEfrosHebert_CVPR11} and divide the affordance constraint into \emph{free space constraint} and \emph{support constraint}, respectively. 
%
%The free space constraint states that no human body parts could intersect with any object in the scene, such as furniture, desk or walls.
%
%The support constraint states that the human pose should be supported by a surface of surrounding objects (e.g., floor, bed).
%
% For instance, a standing human must be supported by the floor and a sitting human must be supported by a chair or sofa. 
%
%We denote the adjustment process as \emph{affordance adjustment}. To make it fully automatic, as shown in ~\cite{GuptaSatkinEfrosHebert_CVPR11}, we perform 3D correlations between scene voxel and poses to find suitable locations for the generated poses.\\
%
%Note that we process the standing poses and sitting poses respectively, while poses are attributed to them according to their classes among the 20 pose clusters (see Section~\ref{sec:2daffordance}).\\
%

%\vspace{-1mm}
\noindent \textbf{Free space constraint.} The free space constraint states that no human body parts can intersect with any object in the scene, such as furniture or walls. To satisfy this constraint, we perform a 3D correlation between poses and a voxel representation of the scene. 
We denote the voxelized 3D pose as $p$, with all voxel valued as one. We binarized the original voxel (Fig.~\ref{fig:gsearch}(c)) with the free space as zero, and the occupied ones as one, denoted as $V_f$.
The free space constraint is satisfied in the locations where $R_f$ below a threshold $T_f$:
\vspace{-2mm}
\begin{equation}
\label{eq:free_space_constraint}
    R_f = p * V_f
    \vspace{-1mm}
\end{equation}
where $*$ indicates a 3D correlation operation.
%
%Any location with zero response indicting no intersection between human body parts and objects in the scene. 
%
Necessary contacts between human and objects (e.g., a human touches the chair when sitting, or the floor when standing) should be considered. 
Thus we mask out these body parts that have to contact with objects, including thigh and pelvis for sitting poses and feet for standing poses, when performing the 3D correlation.

%\vspace{-1mm}
\noindent \textbf{Support constraint.} 
%According to the support constrain, humans should be supported by surrounding objects such as chair or floor to be physically stable. 
%
The support constraint states that the human pose should be supported by a surface of surrounding objects (e.g., floor, bed). 
%Gupta et al.~\cite{GuptaSatkinEfrosHebert_CVPR11} manually create a set of interaction blocks beneath the joints that need support forces to indicate locations where support surfaces must be present. 
%However, such solution is not feasible in our case since our poses are generated, other than selected from a set of fixed poses. 
%\xlnote{may ``improve the method in xxx'' is better?}
We search locations that satisfy this constraint by performing two 3D correlations. 
The first correlation is performed between scene voxels $V_s$ and a 3D Gaussian kernel to detect voxel cells on the surfaces of affordable objects (e.g., the bed in Fig.~\ref{fig:gsearch}(d)).
The $V_s$ is produced by marking all voxels of affordable objects (chair, sofa, floor etc.) to zero, and the other voxels (including unoccupied voxels or objects that can not support a human pose) to one.
After correlating with a 3D Gaussian kernel, all voxels except voxels on the boundaries will be either zero or one. Masking them out would leave us only voxels on affordable objects boundaries.
We further mask out boundary voxels that do not have an upward surface normal.

% \xlnote{I'm trying to make it clear here...}
% \xlnote{To be specific, we mark all voxels of affordable objects (chair, sofa, floor etc.) to zero and other voxels (including unoccupied voxels or objects that can not support a human pose) to one. After correlating with a 3D Gaussian kernel, all voxels except voxels on the boundaries will be either zero or one, masking them out would leave us only voxels on affordable objects boundaries. We further mask out boundary voxels that do not have an upward surface normal using gradients.}
%first evaluating a 3D correlation between a 3D Gaussian kernel with the scene voxel containing only objects that can afford poses (e.g. sofa, chair or floor). 
%
%We only keep voxel cells on the surfaces of affordable objects by examining gradients at upward direction of each voxel cell and filtering out boundaries that do not have an upward surface normal. 
%
%With the filtered voxel, 
%\slnote{what is "evaluate a correlation"}
Next, we perform another 3D correlation between poses and the object surfaces (see Fig.~\ref{fig:gsearch}(e)) and take the location with the maximum correlation score as the optimal location for putting the pose (see Fig.~\ref{fig:gsearch}(f)). 
Similar to the free space constraint discussed above, we denote the voxelized 3D human pose and pre-processed affordable object boundary voxel as $p$ and $V_s$, the Gaussian kernel as $G$, then the \textit{support} constraint $R_s$ can be expressed as:
\begin{equation}
 \label{eq:constraint}
    R_s = p * (G * V_s)   
\end{equation}
We adjust a pose to the ``best location'' where the person can comfortably lay or sit with maximal contacting area with the support surface. The location can be explicitly obtained through localizing at the point with $\max \left(R_s\right)$. 
%
%Note that poses are adjusted in a local region which do not violate too much of the semantic information. 
Note that poses are adjusted in a local region to preserve the semantic information.
Poses that do not find a valid location are discarded, i.e., the support constraint is satisfied in the locations where $\max \left(R_s\right)$ is above a threshold $T_s$.

\begin{figure*}[t]

\centering

\includegraphics[width=16cm,scale=0.5]{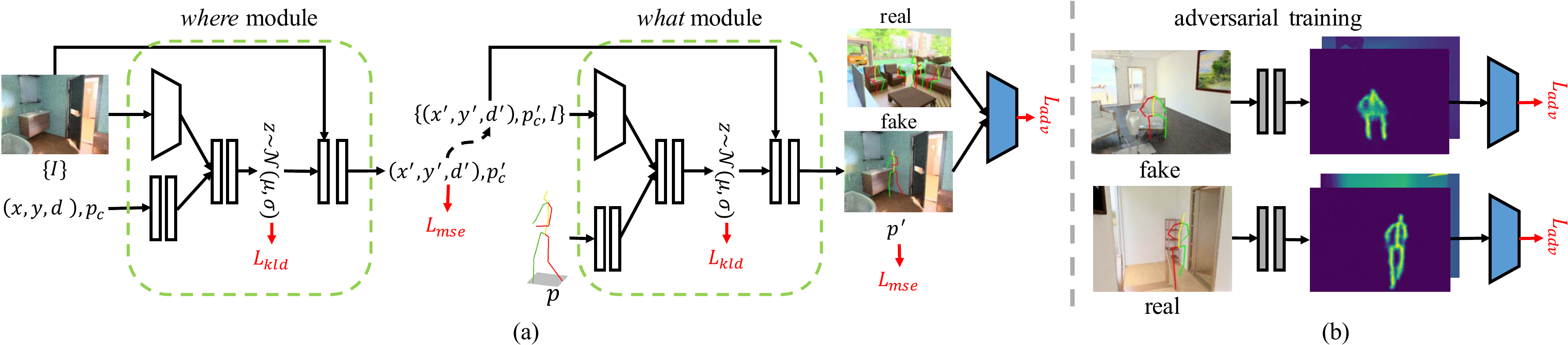} 

\caption{\textbf{Overview of the 3D affordance learning model.} (a) Our end-to-end framework consists of a \emph{where} (Section~\ref{sec:where}) and a \emph{what} (Section~\ref{sec:what}) component for pose location and gesture prediction respectively. (b) Detailed illustration of our adversarial training (blue block in (a), detailed in Section~\ref{sec:joint}). Grey blocks convert joint coordinates and depth to a ``depth heat map'', which are pretrained and fixed when jointly training the \emph{where} and the \emph{what} module. Blocks with same color share parameters.}
\label{fig:arch}
\end{figure*}

\section{3D Affordance Generative Model}
\label{sec:3daffordance}
%\slnote{I have the biggest concern with this section: I suggest we should start with supervised/unsupervised path and why do we designing it, the VAE-GAN loss, and where/what modules.
%Now the big picture is not clear and the two paths are not even mentioned.
%}

%\slnote{Compared to the dataset part, this section is more important and deserve more space. Dataset has many details, however, people often less concern about it, and will assume it is already there as the ground-truth. 
%}
%

In this section, we show how to generate 3D human poses conditioned on a single scene image using the synthesized data described in Section~\ref{sec:poseproduce}.
Generating human poses in 3D scenes requires modeling the joint distribution of human \textit{scale}, \textit{pose}, \textit{location} and \textit{interactions} with objects in 3D, which is very challenging.
A typical solution is to use a single network to model the joint distribution of pose locations and gestures. This approach, however, will result in a huge solution space and poor performance, as analysed in Section~\ref{sec:ablation}.
In contrast, we break it down to two jointly learned sub-tasks, where the generative model for each sub-task is much easier to learn, 
%and (b) the two models are still connected and jointly optimized.
%In this work, we instead factorize the problem into two jointly learned tasks. 
%
To be specific, we first predict the plausible locations in a scene (see the \emph{where} module in Fig.~\ref{fig:overview} and Fig.~\ref{fig:arch} (a)) and then predict the suitable human poses that are aligned with their surrounding context (see the \emph{what} module in Fig.~\ref{fig:overview} and Fig.~\ref{fig:arch} (a)) of the predicted locations. %
Both modules are jointly trained using the pose location as a differentiable link, which allows the two modules to mutually benefit from each other, 
% and be regularized simultaneously by
as well as from the discriminator described in Section~\ref{sec:joint}. 

We take two factors into consideration when designing both the \emph{where} and the \emph{what} modules. First, both modules should be able to understand the semantics of scene context to generate poses that follow natural human behaviors (e.g., sit rather than stand on a sofa). To this end, we model the distributions of pose locations and gestures by two VAEs conditioned on the scene context. We explain them in detail in Section~\ref{sec:where} and Section~\ref{sec:what} respectively.
Second, both modules should be able to hallucinate 3D geometry of the scene to generate poses that obey physical rules in a scene (e.g., poses should be well supported by objects rather than float in the air). To achieve this goal, we introduce a geometry-aware discriminator that further regularizes the two modules to generate physically correct poses, which we discuss in Section~\ref{sec:joint}. Fig.~\ref{fig:arch} illustrates the complete pipeline of our pose prediction model.

\subsection{The \emph{Where} Module: Pose Locations Prediction}
\label{sec:where}
%
%The goal of the \emph{where} module is to model the spatial distribution of valid human pose locations (i.e., the pixel coordinates and the depth of the pelvis joint) in 3D scenes conditioned on scene context.
%
%To achieve the goal, two important factors need to be considered: 
%
%First, the locations of human should follow the semantic and geometric constraint within a scene. For example,  locations inside an object are not valid. This requires our model to be able to understand the scene context. 
%
%Second, multiple locations are possible to place a human pose in a scene, which requires our model to learn the distribution of pose location rather than infer a single location.
%
Given a scene image $I$, we build a \emph{where} VAE to encode pose location in the 3D scene, by simultaneously reconstructing 
%\slnote{should we say "reconstructing" rather than "predicting"?} 
pose pelvis joint coordinates $(x,y)$ and depth $d$, as well as the most likely pose class $p_c$ at the predicted location. The standard variational equality is represented as:
\begin{eqnarray}
\label{eq:whereVAE}
    &\log{P(Y|I)} - KL(Q(z|Y,I)||P(z|Y,I))\\ \nonumber &= E_{z\sim Q}(log{ P(Y|z, I)}) - KL(Q(z|Y,I)||P(z|I))
\end{eqnarray}
$P(z|I)$ and $Q(z|Y,I)$ are two normal distributions $\mathcal{N}(0,1)$ and $\mathcal{N}(\mu(Y,I),\sigma(Y,I))$ and $KL$ represents the Kullback-Leibler divergence. 
%\khnote{Isn't it common to use $D_{\mathrm{KL}}$?}

The pose class $p_c$ provides a clue for the likely pose appearance (e.g., sitting or standing), which can be obtained by assigning each pose to one of the 30 pose clusters described in~\cite{Wang_affordanceCVPR2017}.
%\slnote{did we carefully described it?}.
%
Note that~\cite{Wang_affordanceCVPR2017} uses an one-hot vector to represent the pose class,  which does not consider the similarities of different pose typologies between classes. 
Here we directly represent $p_c$ by the normalized center pose of each cluster so that similar pose classes also have similar representations, i.e., each $p_c\in\mathcal{R}^{3\times 17}$ (each pose contains 17 joints).
%\xlnote{not sure about ``their''} 
%\khnote{may be  ``$p_c$ in the normalized coordinates"?} \\

%The encoder learns to map scene context, pelvis location and pose center to normal distribution $Q$ by estimating $\mu(Y,I)$ and $\sigma(Y,I)$ while the decoder learns to reconstruct pelvis location and pose center by modeling log-likehood $log{P(Y|z, I)}$.
%\JK{unconnected equations}
%

\noindent \textbf{The structure of the \emph{where} module.}
As illustrated in Fig.~\ref{fig:arch} (a), the encoder extracts image features using an 18 layer ResNet~\cite{he2016deep} and concatenates them with the location features and pose class features extracted by two fully connected layers. 
The final concatenated feature is then fed into four fully connected layers to predict $\mu(Y,I)$ and $\sigma(Y,I)$ for distribution $Q$. 
The decoder takes a latent variable $z$ sampled from $Q$ and the scene context features shared with the encoder to predict $\{x,y,d, p_c\}$. 
%
%Note that it is challenging for the model to associate numerical coordinates with image spatial by directly regressing on the coordinates
Because it is challenging for the model to associate numerical coordinates with the exact location in the image,
%~\xlnote{I rephrase the sentence, is it better now?}
%~\khnote{I don't get this sentence}. 
we predict a heat map in the decoder to indicate possible locations for a pose and adopt one Differentiable Spatial to Numerical Transform (DSNT)~\cite{nibali2018numerical} layer to convert the heat map to pose location coordinates.

\noindent \textbf{The objectives of the \emph{where} module.}
We use three losses in training the \emph{where} module. 
First, we minimize the Euclidean distance on the estimated pose class, depth and pelvis coordinates by $L_{mse} = \norm{Y^{\ast} - Y}$. 
Second, we minimize the KL-divergence between the estimated distribution $Q$ and the normal distribution $\mathcal{N}(0,1)$ by $L_{kld} = KL[Q(z|\mu(Y,I), \sigma(Y,I))||\mathcal{N}(0, 1)]$.
In addition, to better associate predicted pelvis joint depth and pixel coordinates, we minimize the Euclidean distance between ground truth and predicted pelvis coordinates under the world coordinate system using camera parameters for each scene. We refer this loss as \emph{geometry loss} and represent it as $L_{geo} = \norm{M_eM_i[x^{\ast},y^{\ast},d^{\ast}] - M_eM_i[x,y,d]}$, where $M_e$ and $M_i$ are camera extrinsic and intrinsic matrices. Our final objective is:

\vspace{-6mm}
\begin{eqnarray}
\label{eq:where_loss}
    L = {\lambda}_{mse}L_{mse} + {\lambda}_{kld}L_{kld} + {\lambda}_{geo}L_{geo},
\end{eqnarray}
where ${\lambda}_{mse}$, ${\lambda}_{kld}$, ${\lambda}_{geo}$ are the weights that balance the three objective terms.

\begin{figure}[t]
\centering
\includegraphics[width=.24\linewidth]{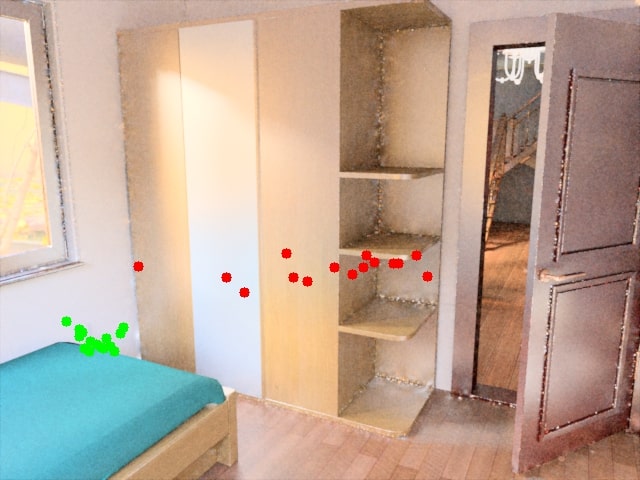}
\includegraphics[width=.24\linewidth]{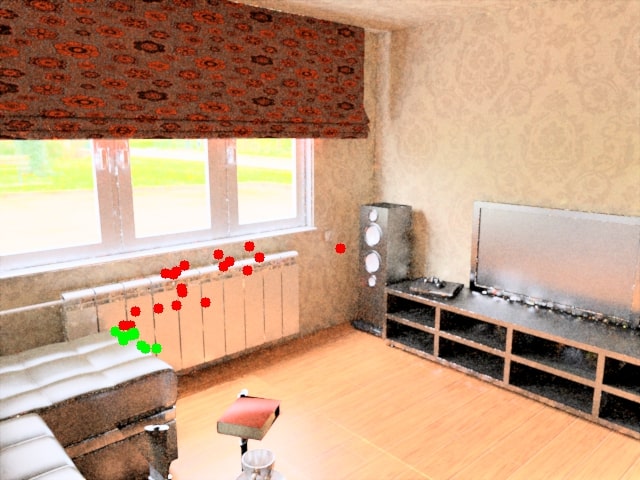}
\includegraphics[width=.24\linewidth]{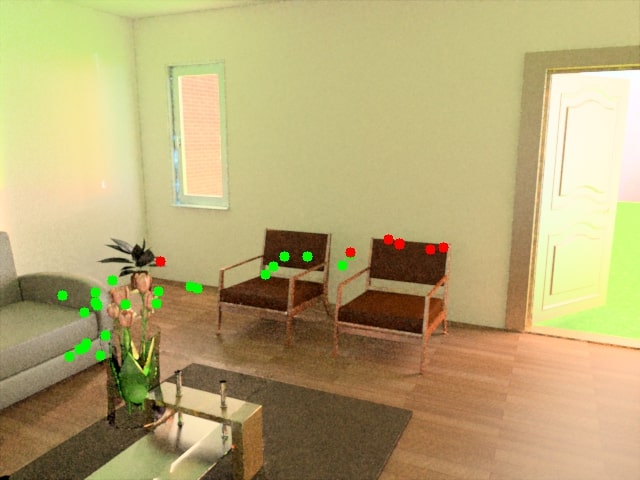}
\includegraphics[width=.24\linewidth]{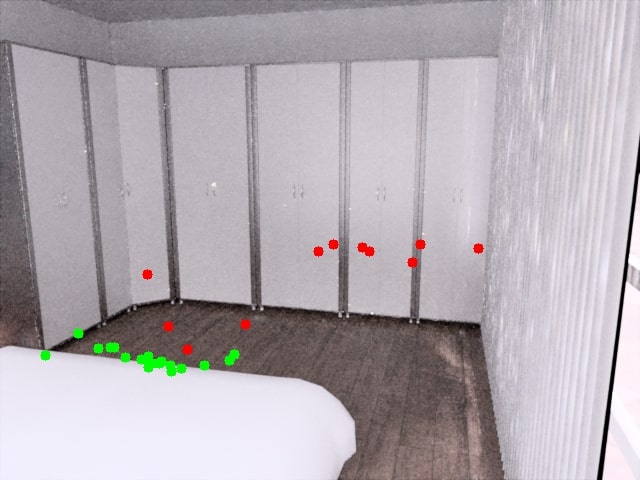}\\
\includegraphics[width=.24\linewidth]{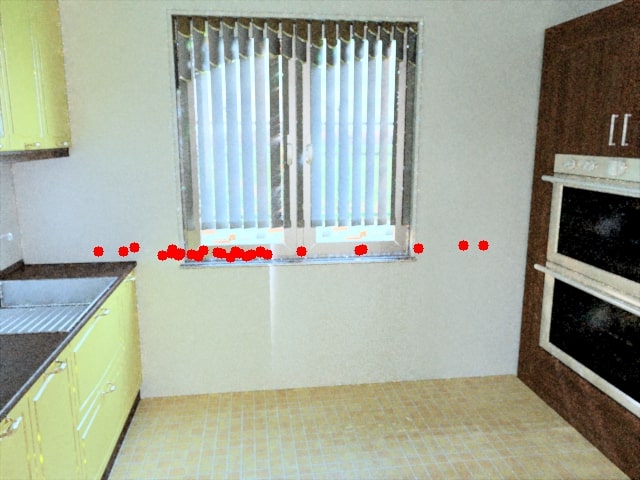}
\includegraphics[width=.24\linewidth]{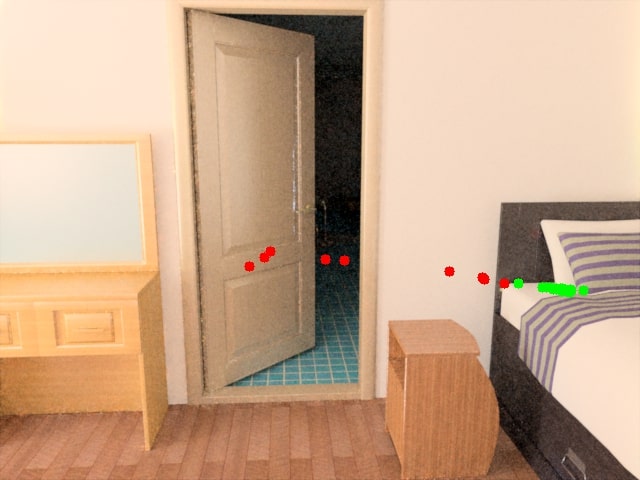}
\includegraphics[width=.24\linewidth]{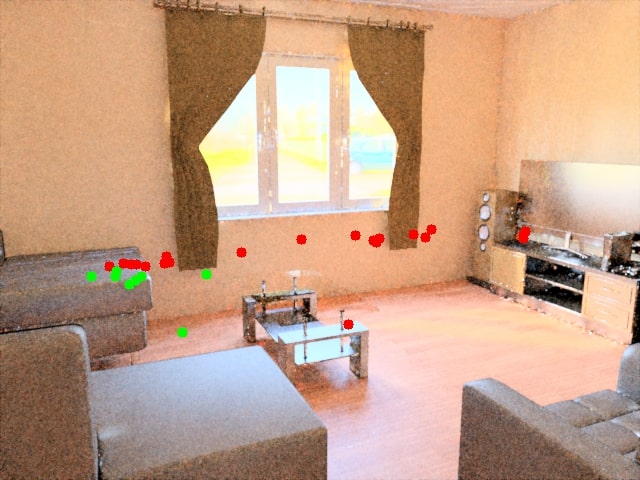}
\includegraphics[width=.24\linewidth]{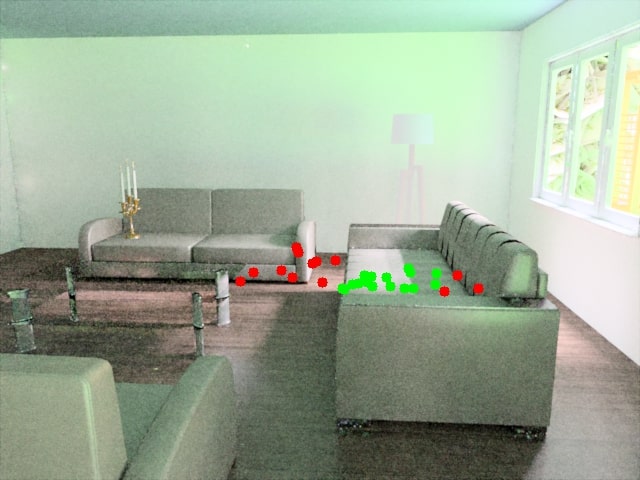}

\caption{ \textbf{Sampled locations by the ``where'' module.} In each scene, we show 30 locations sampled by our ``where'' module. For visualization purpose, we color pelvis joint locations for standing poses and sitting poses red and green respectively.}
\label{fig:where_result}
\vspace{-.2cm}
\end{figure}

We visualize the sampled locations conditioned on each scene image in Fig.~\ref{fig:where_result}. As shown in this figure, our ``where'' module (a) understands the scene and predicts reasonable locations for sitting poses around an affordable object or locations on correct height for standing poses. (b) generates multiple locations given a single scene image.

\subsection{The \emph{What} Module: Pose Gestures Prediction}
\label{sec:what}
%
%In this section, we discuss the design of our \emph{what} module, which is jointly trained with the \emph{where} module described above. 
The \emph{what} module takes pelvis joint coordinates $(x,y)$, depth $d$ and pose class $p_c$ predicted by the \emph{where} module as well as a scene image $I$ as inputs, and learns to predict coordinates and depth of each joint in $p\in\mathbb{R}^{3\times 17}$, so that the generated pose $p$ can align well with its surrounding context. 
%
% Similar to the \emph{where} module, this also requires the \emph{what} module not only understand scene context but also be able to sample different poses given same conditions.
In other words, the \emph{what} module needs to understand the scene context, and be able to sample poses conditioned on it.
Similarly, we model the pose appearance distribution with a conditional VAE, which is represented as:
\begingroup\small
\begin{eqnarray}
\label{eq:whatVAE}
    &\log(P(S|R,I)) - KL(Q(z|S,R,I)||P(z|S,R,I))\\ \nonumber &= E_{z\sim Q}(log{ P(P|z,R,I)}) - KL(Q(z|S,R,I)||P(z|R,I)),
\end{eqnarray}\endgroup
where $S$ represents the coordinates and depth $\{x,y,d\}$ for each joint, $R$ denotes $\{x,y,d, p_c\}$ predicted by the \emph{where} module. 
Other symbols follow those in Section~\ref{sec:where}.

\noindent \textbf{The structure and objectives of the \emph{what} module.} 
Our \emph{what} module shares similar structure as the \emph{where} module (Fig.~\ref{fig:arch}), except that the inputs are pose location, scene context and pose class, and the outputs are the coordinates and depth for each joint.

Similar to the \emph{where} module, the  \emph{what} module 
%The loss of \emph{what} module is similar to the \emph{where} module, which
contains three losses: a Euclidean loss on estimated joint coordinates and depth $L_{mse} = \norm{S^{\ast} - S}$, a KL-divergence loss $L_{kld} = KL[Q(z|\mu(R,I), \sigma(R,I))||\mathcal{N}(0, 1)]$, and a geometry loss $L_{geo} = \norm{M_eM_i[x_j^{\ast},y_j^{\ast},d_j^{\ast}] - M_eM_i[x_j,y_j,d_j]}$,  where $[x_j,y_j,d_j]$ are pixel coordinates and depth for joint $j$. 
While our goal is to model the shape of poses through modeling the joint distribution of joints ${S}$, 
the final objective is same as in Equation~\ref{eq:where_loss}.
% \slnote{need to point the difference between the mse for where and what}
% \xlnote{I guess we have use S in the Lmse loss, so it should be different from the mse loss of the ``where'' module.}

\subsection{The Geometry-Aware Discriminator}
\label{sec:joint}
%
%In this section, we show the two modules can be jointly trained, so that the location of the pelvis can be adjusted according to the supervision signal from the generated poses.
% the \emph{Where} module can be adjusted by the supervision signals coming from the \emph{What} module.
% One crucial advantage of our model is its ability to hallucinate the spatial relationship between human and its surrounding objects in 3D scenes by estimating the depth of pelvis joint. Accurate pelvis joint depth is key for the \emph{what} module to predict suitable poses that satisfy the geometry constraints in a scene. 
% %
% However, it is a challenging task for the \emph{where} module to regress accurate depth due to difficulties in balancing between multiple losses. Thus we propose to train the \emph{where} and \emph{what} module together so that the they can benefit each other and be optimized jointly.
%
% The joint training of \emph{where} and \emph{what} module benefits our model in two ways. First, using the pelvis location as a differentiable link, the \emph{what} module is able to backpropagate gradients back to the \emph{where} module to enforce it generating more feasible pelvis locations. More importantly, it enables us to introduce adversarial training which potentially benefits the two modules together by using a discriminator that is aware of the scene geometry (i.e. depth image of the scene).
%
In this work, we aim to generate poses in 3D scenes that follow physical rules in the scene, which requires our model to properly hallucinate the 3D scene geometry merely from a 2D image. 
To this end, in addition to including the depth value of each pose during training, we introduce a geometry-aware discriminator that further regularizes the \emph{where} and \emph{what} module simultaneously to generate poses that obey geometry rules in the scene. 
% \xlnote{maybe discriminator indicates training? so we do not need to tell specifically this is only used during training.}
%
%To further improve geometry feasibility of generated poses at inference stage, we introduce a conditional discriminator into our unsupervised path (see Figure~\ref{fig:arch} (b)). 
%
%This is based on the intuition that knowing both depth of each pose joint and scene object, one can easily infer the spatial relationship between objects and humans (i.e. occlusion). 
%

As shown in Fig.~\ref{fig:arch}(b), the discriminator takes generated poses and scene depth images as inputs and learns to discriminate between geometrically feasible (real) \emph{vs.}\ unfeasible (fake) pairs.
However, it is challenging for the discriminator to associate the discrete depth value of each joint to a scene depth map (i.e., the depth of each point between two connected joints is not modeled).
% 3D spatial location of each joint with its numerical coordinates and depth 
% (e.g. knowing which joints should be connected or the depth of points between two joints), 
Thus we first train a network which converts coordinates and depth of each joint to a ``depth heat map'' (Fig.~\ref{fig:arch}(b)), where each pixel is either the depth of a point between two joints or $-1$ for background pixels. Details about the network are available in the appendix. We then feed this ``depth heat map'' together with the scene depth image into the discriminator. 
%Our final adversarial objective $L_{adv}(G,D)$ is ${\mathop{\mathbb{E}}}_{c,p_r}[\log{D(F(p_r),c)}] + {\mathop{\mathbb{E}}}_{c,p_z}[\log{(1-D(F(p_z),c))}]$,
Our final adversarial objective is:
\begingroup\small
\begin{eqnarray}
 \label{eq:advloss}
 \begin{aligned}
     L_{adv}(G,D) =\,  &{\mathop{\mathbb{E}}}_{c,p_r}[\log{D(F(p_r),c)}] + \\ & {\mathop{\mathbb{E}}}_{c,p_z}[\log{(1-D(F(p_z),c))}]
 \end{aligned}
 \end{eqnarray}\endgroup

\noindent where $G$ and $D$ represent the pose prediction model and the discriminator model, $F$  represents a pre-trained CNN that converts joint coordinates and depth to the ``depth heat map'' described above, $p_r$ and $p_z$ denote ground truth and generated poses, $c$ denotes the depth image of the scene.

We note that both the geometry-aware discriminator as well as the geometrically feasible/unfeasible labels are utilized only during training.
During testing, only the the part shown in Fig.~\ref{fig:arch}(a) is needed to support single image conditioned generation, which makes the algorithm easy to be adapted to many application scenarios.

\vspace{-2mm}
\section{Experimental Results}
\vspace{-1mm}
\begin{figure*}
	\centering
	\begin{tabular}{c@{\hspace{0.005\linewidth}}c@{\hspace{0.005\linewidth}}c@{\hspace{0.005\linewidth}}c@{\hspace{0.005\linewidth}}c@{\hspace{0.005\linewidth}}c@{\hspace{0.005\linewidth}}c}

	    \includegraphics[height = .11\linewidth, width = .16\linewidth]{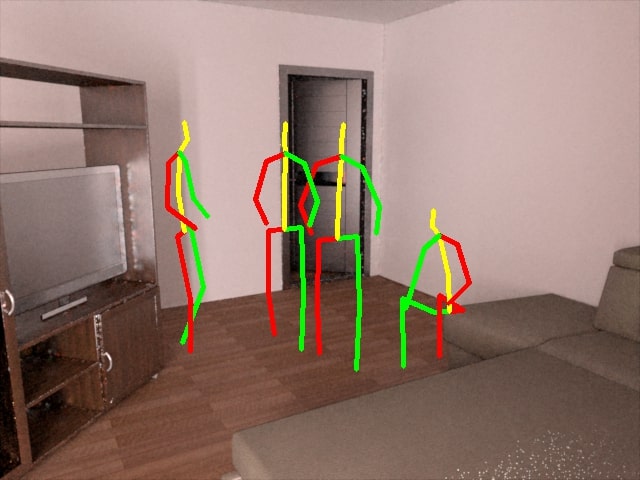} &
    	\includegraphics[height = .11\linewidth, width = .16\linewidth]{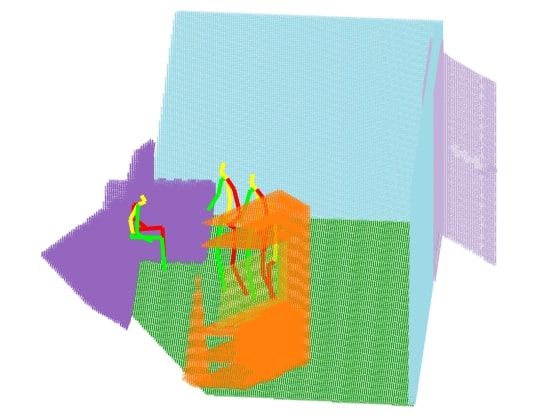} &
    	\includegraphics[height = .11\linewidth, width = .16\linewidth]{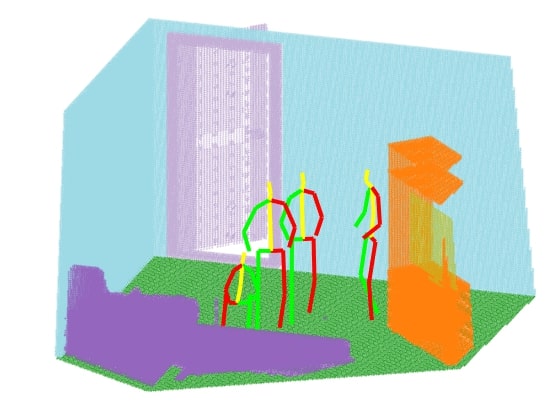} &
    	\includegraphics[height = .11\linewidth, width = .16\linewidth]{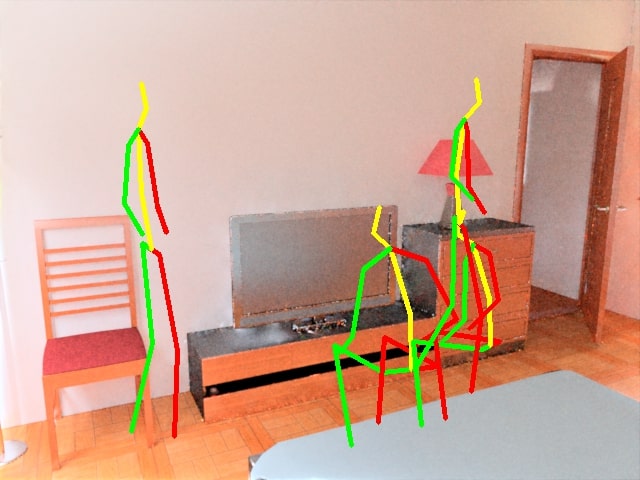} &
    	\includegraphics[height = .11\linewidth, width = .16\linewidth]{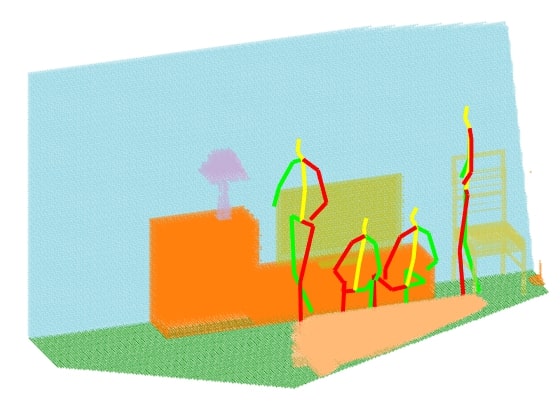} &
    	\includegraphics[height = .11\linewidth, width = .16\linewidth]{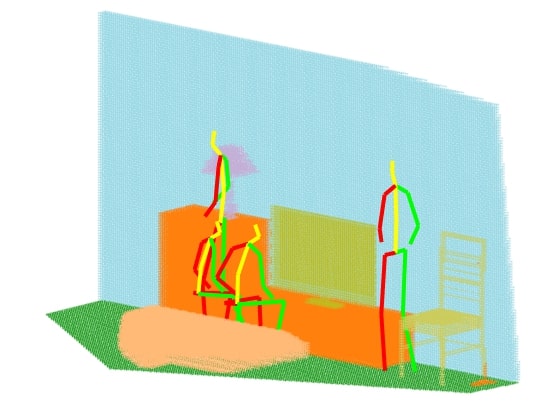} \\

    	\includegraphics[height = .11\linewidth, width = .16\linewidth]{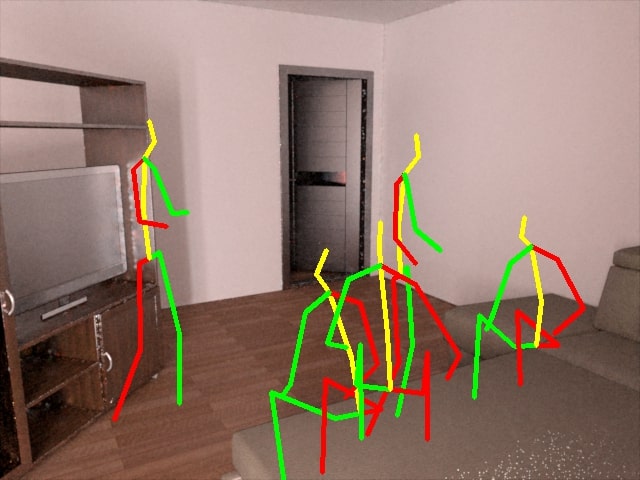} &
    	\includegraphics[height = .11\linewidth, width = .16\linewidth]{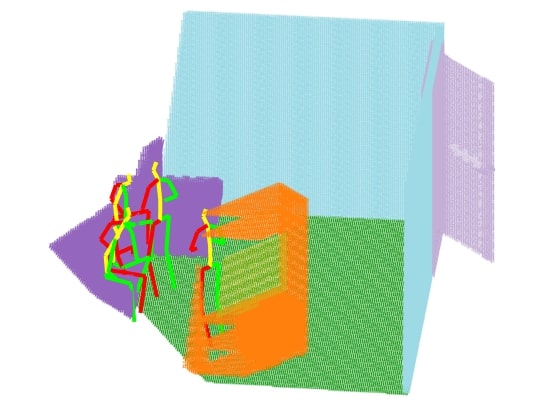} &
    	\includegraphics[height = .11\linewidth, width = .16\linewidth]{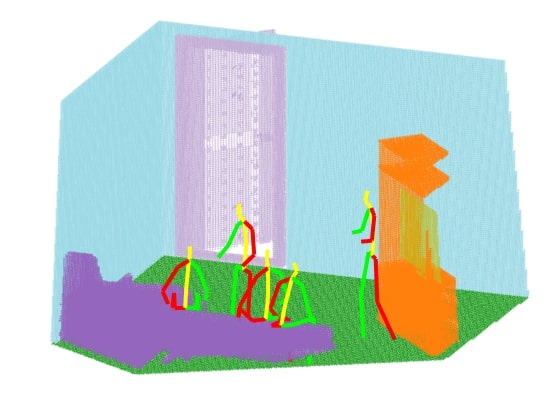} &
    	\includegraphics[height = .11\linewidth, width = .16\linewidth]{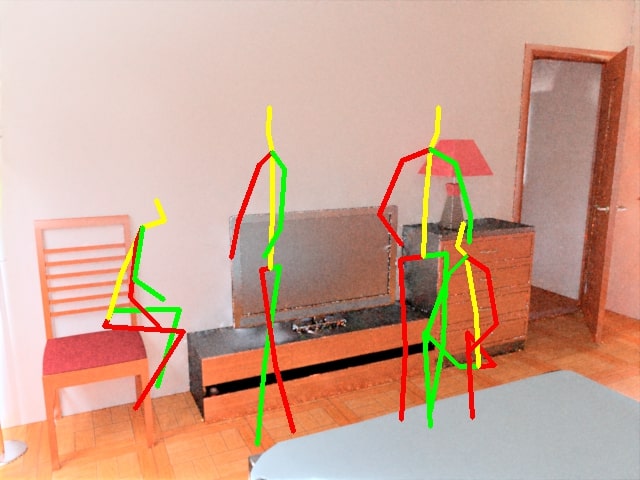} &
    	\includegraphics[height = .11\linewidth, width = .16\linewidth]{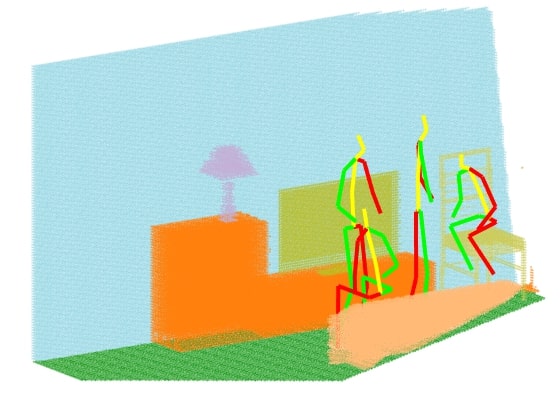} &
    	\includegraphics[height = .11\linewidth, width = .16\linewidth]{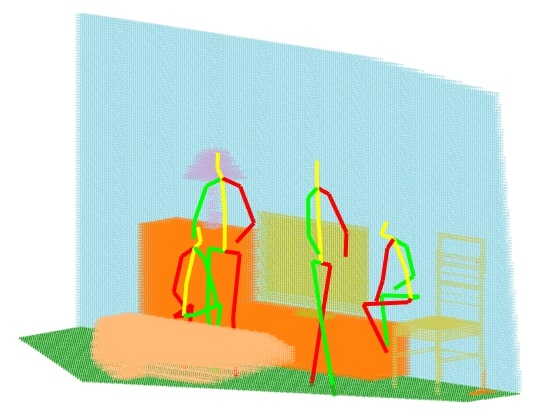} \\

    	\includegraphics[height = .11\linewidth, width = .16\linewidth]{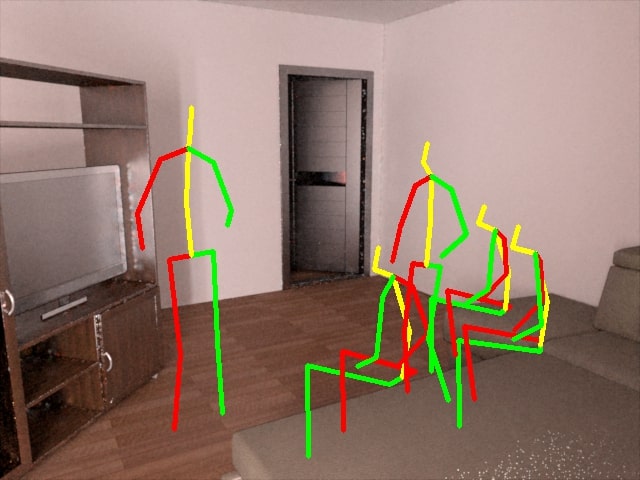} &
    	\includegraphics[height = .11\linewidth, width = .16\linewidth]{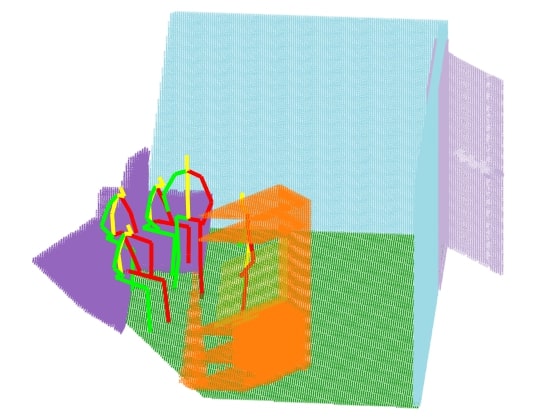} &
    	\includegraphics[height = .11\linewidth, width = .16\linewidth]{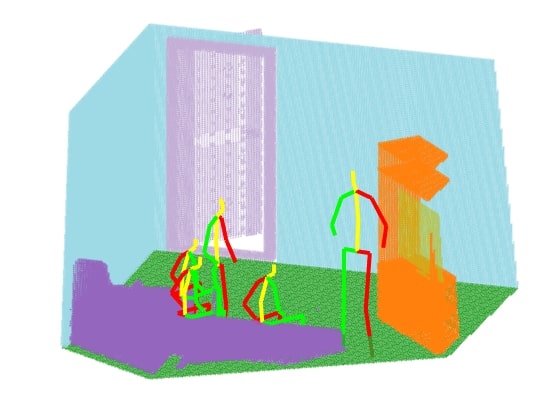} &
    	\includegraphics[height = .11\linewidth, width = .16\linewidth]{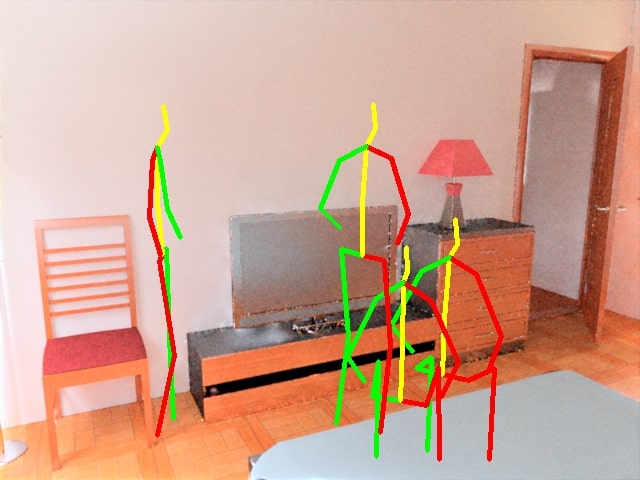} &
    	\includegraphics[height = .11\linewidth, width = .16\linewidth]{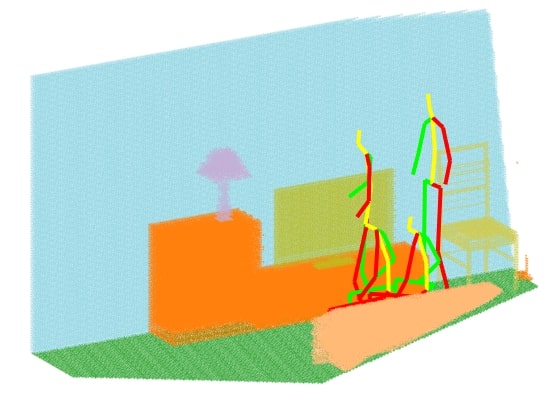} &
    	\includegraphics[height = .11\linewidth, width = .16\linewidth]{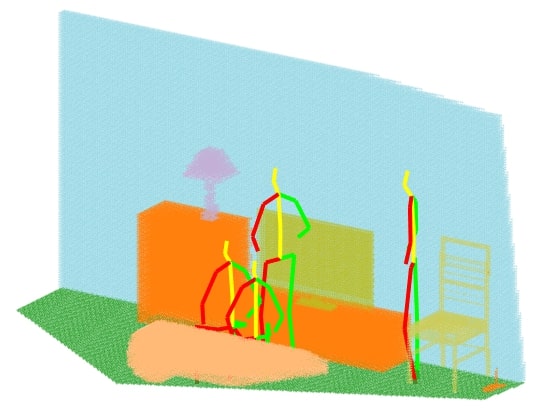} \\

	\end{tabular}
	
	\caption{\textbf{Generated poses by our model.} The three rows show generated poses by models that take a RGB, RGB-D or depth map as input. For each scene, the first column illustrates pose projections in 2D scene images, and the last two columns illustrate poses in scene voxels visualized from different views.}
	\label{fig:where_what}
\end{figure*}

\begin{figure}[t]
\centering
\includegraphics[width=1\linewidth]{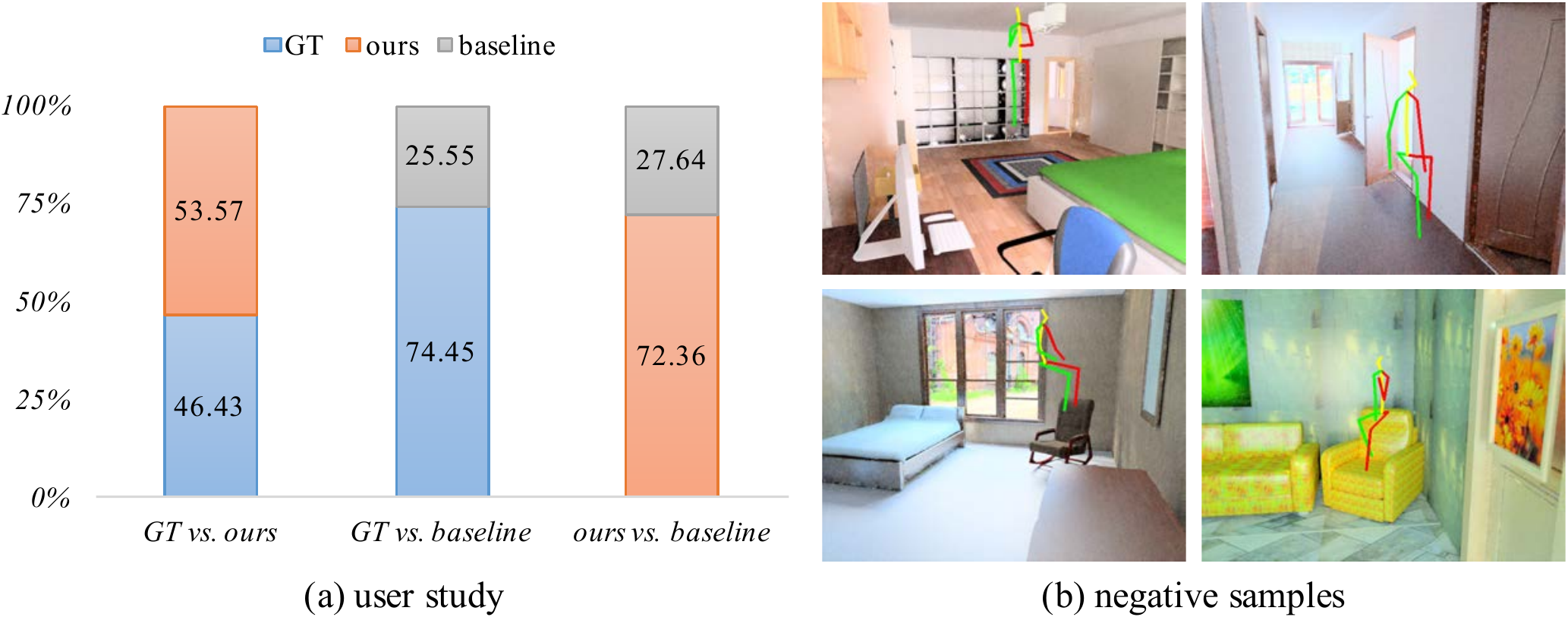}
\caption{\textbf{Semantic plausibility evaluation.} (a) User study results. Each subject is asked to select the more reasonable pose through pairwise comparisons. The number indicates the percentage of preference on that comparison pair. ``GT'' means ground truth poses. (b) Manually annotated negative pose samples that are either impossible (column 1) or uncommon (column 2) in an indoor environment.
}
\label{fig:user_study}
\vspace{-.2cm}
\end{figure}

In this section, we first introduce the details of our synthesized dataset and the quantitative evaluation metrics in Section~\ref{sec:metric}. 
Then, we present the experimental results of our affordance prediction model in Section~\ref{sec:where_what}, as well as the ablation studies to understand how the main modules of the proposed algorithm contribute in Section~\ref{sec:ablation}. 
Finally, we compare the proposed method with the  state-of-the-art affordance prediction method~\cite{Wang_affordanceCVPR2017} in Section~\ref{sec:what_eval}.

\label{sec:eval}
\subsection{Dataset Synthesis and Evaluation Metrics}
\label{sec:metric}
\noindent \textbf{Dataset synthesis.} 
As described in Section~\ref{sec:poseproduce}, we use the Sitcom dataset~\cite{Wang_affordanceCVPR2017} for pose prediction in images and map the generated poses into the scene voxels in the SUNCG dataset~\cite{zhang2017physically,song2017semantic} for 3D pose affordance correction.
In total, we apply the synthesizer to generate $1.5$ million poses in $13,774$ SUNCG scenes. We use $13,074$ scenes for training and $700$ scenes for evaluation. 
%Fig~\ref{fig:suncg_gt} shows some samples of synthesized poses in a scene voxel and their projections in scene images.\\

\vspace{1mm}
\noindent \textbf {Quantitative evaluation metrics.} The primary goal of this paper is to model 3D human affordance by generating human poses that are \emph{semantically plausible} and  \emph{physically feasible} in a given scene. %
The semantic plausibility describes how reasonable a generated pose looks in an indoor environment. We design two ways to evaluate it.
%
%First, we follow~\cite{Wang_affordanceCVPR2017} and manually annotate negative poses that are semantically impossible or uncommon in indoor environment as shown in Fig~\ref{fig:user_study}(b). 

First, we train a \emph{pose authenticity classifier} to determine whether a generated pose is plausible.
To train the classifier, we collect the ground truth poses from our synthesizer in Section~\ref{sec:poseproduce} as positive samples, and manually annotate the negative samples following~\cite{Wang_affordanceCVPR2017}. As shown in Fig.~\ref{fig:user_study}(b), the negative pose samples are either impossible or uncommon to appear in an indoor environment. 
In total, we collect $18,000$ pose samples in different scenes for training, and $1,400$ pose samples for evaluation. Both the training and the testing dataset contain an equal number of positive and negative poses.
Our trained \emph{pose authenticity classifier} achieves a classification accuracy as high as 86\% on the testing dataset, and is ready to be used to test the plausibility of a pose, i.e., to check if a pose looks like a natural human pose in the given scene context. 
We define the ratio of poses that are classified as positive by the \emph{pose authenticity classifier} as \enquote{semantic score}. High semantic scores indicate that the model is able to understand the scene semantics to generate plausible poses in an indoor environment.

\begin{figure}[h]
\centering
\includegraphics[width=.46\linewidth]{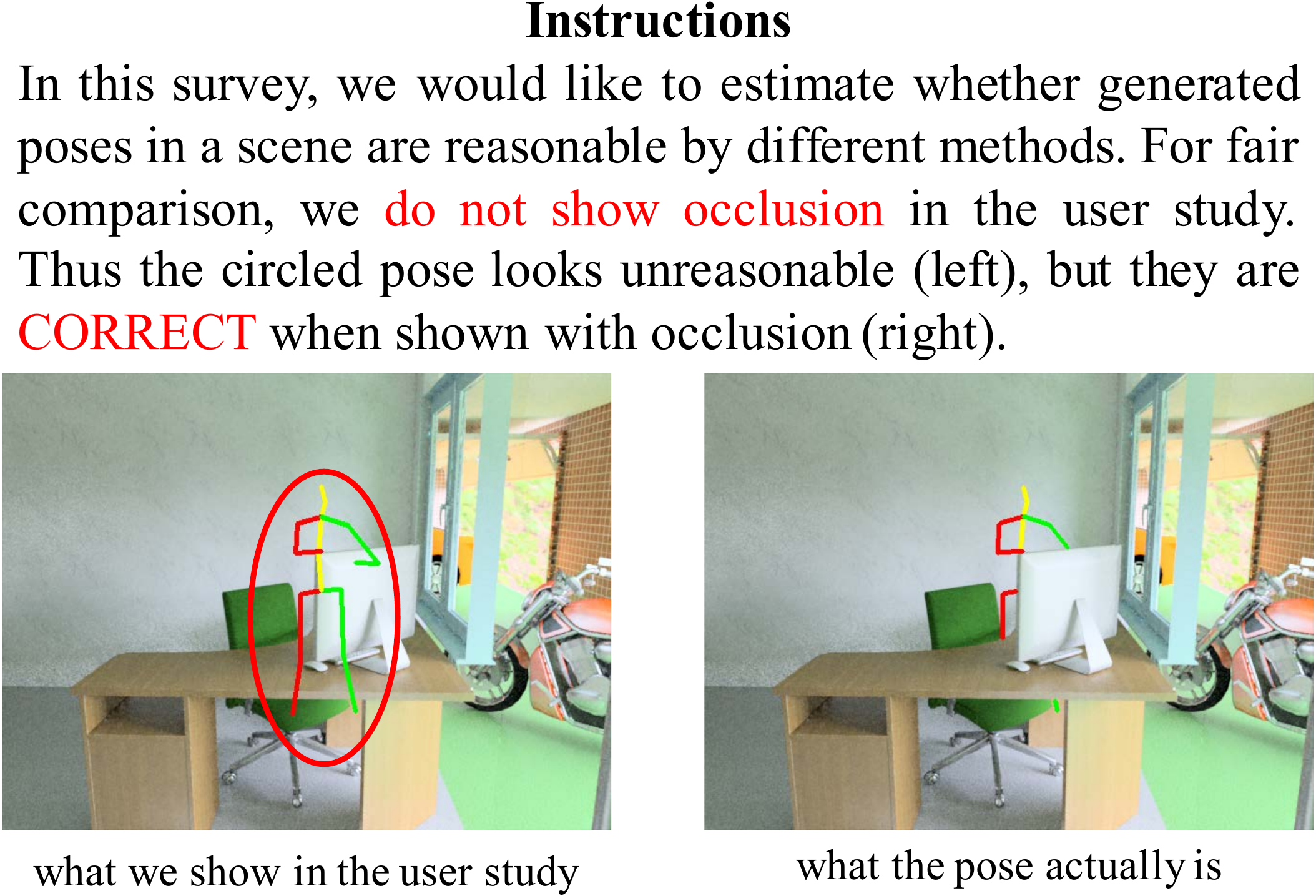}
\includegraphics[width=.46\linewidth]{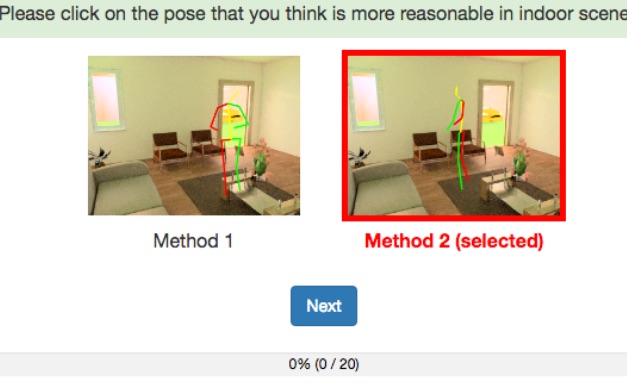}
\caption{ \textbf{User study instructions and interface.} (Left) Instructions, (Right) User interface.}
\label{fig:user_study_interface}
\vspace{-.2cm}
\end{figure}

Second, we conduct a user study to let humans to determine how authentic the generated poses look like.
Given a pair of poses sampled from ground truth poses and generated poses, either by the baseline method~\cite{Wang_affordanceCVPR2017} or our method, in the same scene, a user is asked to select the pose that is more reasonable in an indoor environment. Fig.~\ref{fig:user_study_interface} shows the instructions and web UI.
Note that since we focus on visual plausibility, both the generated/ground truth poses and the scenes for user study are projected and displayed as 2D images, which can be compared with~\cite{Wang_affordanceCVPR2017}.

%
%We compare the proposed method against the state-of-the-art method~\cite{Wang_affordanceCVPR2017}. 
%
%%One is from the ground truth poses, and the other is either either from our method or from the baseline method. 
%
%We then ask the user to select the pose that is more reasonable in an indoor environment. 
%We collect 400 votes from 20 users and present the result in Fig.~\ref{fig:user_study}(a), which shows that our method is not only able to generate poses that are more realistic compared to the baseline model, but also indistinguishable from ground truth poses.

Finally, to check if a generated pose violates the geometric rules in a scene, we map it into the corresponding scene voxel, and check if the pose satisfies the free space constraint and support constraint as discussed in Section~\ref{sec:3Dgc}.
We re-utilize the constraints as our evaluation criteria, by defining the ratio of poses that satisfy both constraints as \emph{geometry score}. To be specific, for a standing pose, it satisfies the support constraint if the feet of the pose is within 8 voxel units (each voxel unit is 0.02 meter) of the floor. For a sitting pose, it satisfies the support constraint if there is an affordable surface (with $T_s >= 100$ as discussed in Section~\ref{sec:3Dgc}) within 8 voxel units of the pose. Furthermore, a pose that intersects less than or equal to 5 voxels (i.e. $T_f <= 5$) is considered satisfying the free space constraint. 
High geometry scores indicate that the model can hallucinate the 3D geometry and obey the geometry rules in the scene.

\vspace{-2mm}
\subsection{3D Affordance Prediction}
\label{sec:where_what}
We visualize the generated poses by our \emph{where} and \emph{what} module with different input modalities in Fig.~\ref{fig:where_what}. 
We present quantitative evaluations in Table~\ref{tab:where_what}. 
For each model, we generate $3,500$ poses and calculate the semantic as well as geometry score over these poses.
%As discussed in Section~\ref{sec:}, our model can predict both the locations of human models and the coordinates of joints describing their gestures in a scene. 
Note that the previous work~\cite{Wang_affordanceCVPR2017} only focuses on predicting pose gestures at given locations. 
For a fair comparison, we combine the location heat map prediction model introduced in section~\ref{sec:2daffordance}, with the pose generator from~\cite{Wang_affordanceCVPR2017} as our baseline model. 
Furthermore, since the baseline model is not able to predict the pose depth values, to calculate the geometry score described in Section~\ref{sec:metric}, we adopt the strategy as introduced in Section~\ref{sec:mapping} to estimate the pose depth and map the poses into the 3D scene. 

Even with a single RGB image as input, our method achieves $19.47\%$ higher semantic score, and $19.02\%$ higher geometry score than the baseline model (see Table~\ref{tab:where_what}(b) and (c)). 
The results indicate that our model is able to understand both the context and moreover, the geometry of a scene.
In addition, we generate 50 poses in different scenes and conduct the user study discussed in Section~\ref{sec:metric}. In total, we collect 400 votes from 20 users and present the result in Fig.~\ref{fig:user_study}(a).
According to the user study result, the poses generated by our method are not only more reasonable than poses predicted by the baseline method, but also indistinguishable from the ground truth poses. 

Furthermore, we show that our pose prediction model can be further improved by including depth information of the scene. Specifically, we train two variants of our model that take a RGB-D or a depth map as input and present their performance in Table~\ref{tab:where_what}. From this table, we can see that including depth information of the scene constantly improve the geometry score of the pose prediction model under different experimental settings. Similar observations can also be found in Fig.~\ref{fig:where_what}, where the sitting pose generated by the model that takes a RGB image as input floats above the sofa (column 3, row 1), while the sitting pose generated by the model that takes a RGB-D or depth map as input aligns well with the sofa (column 3, row 2 and 3).

\begin{table*}
  \centering
  {\caption{\textbf{Quantitative evaluation of our affordance prediction model.} We show comparisons of our model with three different input modalities against the baseline model  described in Section~\ref{sec:where_what} in (b) and (c). Additionally, we show the performance of different variants of our model in (d) to (f) as discussed in Section~\ref{sec:ablation}. }\label{tab:where_what}. }
  {\scriptsize
  \begin{tabular}{l|l|lll|lll|lll|lll}
  \hline
    \multirow{2}*{(a) Metric} & \multirow{2}*{(b) Baseline} & \multicolumn{3}{c|}{(c) Ours} & \multicolumn{3}{c|}{(d) Ours w/o adversarial} & \multicolumn{3}{c|}{(e) Ours w/o joint training} & \multicolumn{3}{c}{(f) Ours w/o geometry loss}\\
    & & RGB&RGB-D&Depth & RGB&RGB-D&Depth & RGB&RGB-D&Depth & RGB&RGB-D&Depth\\ \hline
    
    %{semantic score} & 72.53 & 92.00 & 91.51 & 90.54 &92.11 & 89.23 & 91.34 & 83.94 & 75.83 & 80.03 &90.03 & 89.03 & 88.06\\
    
    %{geometry score} & 59.47 & 78.49 & 84.06 & 84.00 &77.38 & 81.46 & 84.47 & 79.28 & 85.43 & 85.40 &73.90 & 82.43 & 82.23\\
    
    {semantic score} & 72.53 & \textbf{91.69} & 91.14 & 89.86 &90.17 & 91.6 & 89.31 & 83.34 & 81.40 & 77.09 & 89.74 & 88.40 & 88.11\\
    
    {geometry score} & 23.25 & 66.40 & 71.17 & \textbf{72.11} & 62.71 &72.00 & 70.91 & 46.46 & 71.37 & 60.83 & 56.11 & 66.40 & 63.77\\
			 \hline
  \end{tabular}}
\end{table*}

\vspace{-2mm}
\subsection{Ablation Studies}
\label{sec:ablation}
%In this section, we discuss the contribution of different designs in our approach.\\
%\begin{table}
%\centering
%  {\caption{\textbf{Ablation study.} We compare the contribution of different desings in our approach. All models evaluated here take RGB image as input. See Section~\ref{sec:ablation} for details.}\label{tab:ablation}}
%  {\scriptsize
%  \begin{tabular}{lllll}
%  \hline
%    \multirow{2}{*}{Model} & \multirow{2}{*}{final design} & without & without & without \\
%    & & adversarial & fine-tune & geo loss \\ \hline
%    {semantic score} & 92.00 & 92.11 & 83.94 & 90.03\\
%    {geometry score} & 78.49 & 77.38 & 79.28 & 73.9\\
%			 \hline
%  \end{tabular}}
%\end{floatrow}
%\end{table}
\noindent \textbf{A single model for affordance learning.} We conduct a baseline method to show that a single, straightforward generative network does not work for modeling complex joint distributions -- 
we use a single VAE to encode 2D scene, pose locations and gestures, where all the other settings remain the same.
We obtain semantic and geometry scores of $76.23$ and $52.94$ when taking RGB images as inputs (Table~\ref{tab:where_what} (b)), which are much worse than the proposed method (Table~\ref{tab:where_what} (c)).

\noindent \textbf{Joint training.}
First, we evaluate our model without joint training the \emph{where} and \emph{what} module. Table~\ref{tab:where_what}(c) \textit{vs.} (e) shows the significant contribution of joint training for the semantic score. Without it, the semantic score reduces by $8.06\%$ when taking a RGB image as input. 
We observe that although the model without joint training present higher geometry score, many of the generated locations have wrong depth values, which lead to unreasonably small poses that do not collide with other objects. 
% Note that the geometry score of the model without joint training is higher because the inaccurate depth leads to unreasonably smaller poses, which are less prone to collision with other objects.

\vspace{1mm}
\noindent \textbf{Adversarial training.} 
Hallucinating 3D geometry purely based on 2D information is a challenging task. Thus we propose to use a geometry-aware discriminator which conditions on the depth map of a scene and learns to discriminate generated poses from ``ground truth'' poses (see Section \ref{sec:joint}). Table~\ref{tab:where_what}(c) \textit{vs.} (d) shows the effectiveness of adversarial training. 
With adversarial training, our model is able to generate poses that better obey the rules of geometry in a scene (higher geometry score).

\vspace{1mm}
\noindent \textbf{Geometry loss.} 
A pose that looks plausible in a 2D context may still violate the rules of geometry when mapped into the 3D scene. 
Thus, to encourage our model to generate poses that are consistent with the geometry of the 3D world, we minimize the Euclidean distance between predicted poses and ground truth poses in the world coordinate space.
% we map generated poses back into the 3D scene for Euclidean loss calculation.
%
Table~\ref{tab:where_what}(c) \textit{vs.} (f) demonstrates the contribution of the geometry loss. Without it, the geometry score drops by 4.59\% when taking a RGB image as input.
% which shows its effectiveness in training  models that are better at hallucinating 3D geometry in scenes.
%\noindent \textbf{Input modality.}
%Intuitively, depth information of a scene should assist our model to generate poses better obey geometry constraints. Table~\ref{tab:3dWhat_eval} and Table~\ref{tab:where_what} verifies this intuition, i.e., when taken RGBD or depth image as inputs, our model achieves higher geometry score. Similar observation can be found in Figure~\ref{fig:where_what}, the generated sitting pose given only RGB information (row 1) looks plausible under 2D context. However, it barely touches the sofa when mapped back into 3D scene. In contrast, all sitting poses align well with the surface of sofa when given RGBD/depth (row 2/3) information as inputs. \\
%\noindent \textbf{Class reconstruction in \textit{where} module.}
%We also evaluate a variant of our model when removing the center related branch in both \emph{where} and \emph{what} module. Table~\ref{tab:ablation} column 2 \textit{vs.} column 5 shows that removing this branch slightly drops the geometry score slightly by 0.71\%.\\
\subsection{Comparison with State-of-the-Art}\label{sec:what_eval}

\begin{figure}[t]
	\centering
	\includegraphics[width=8.5cm,scale=0.5]{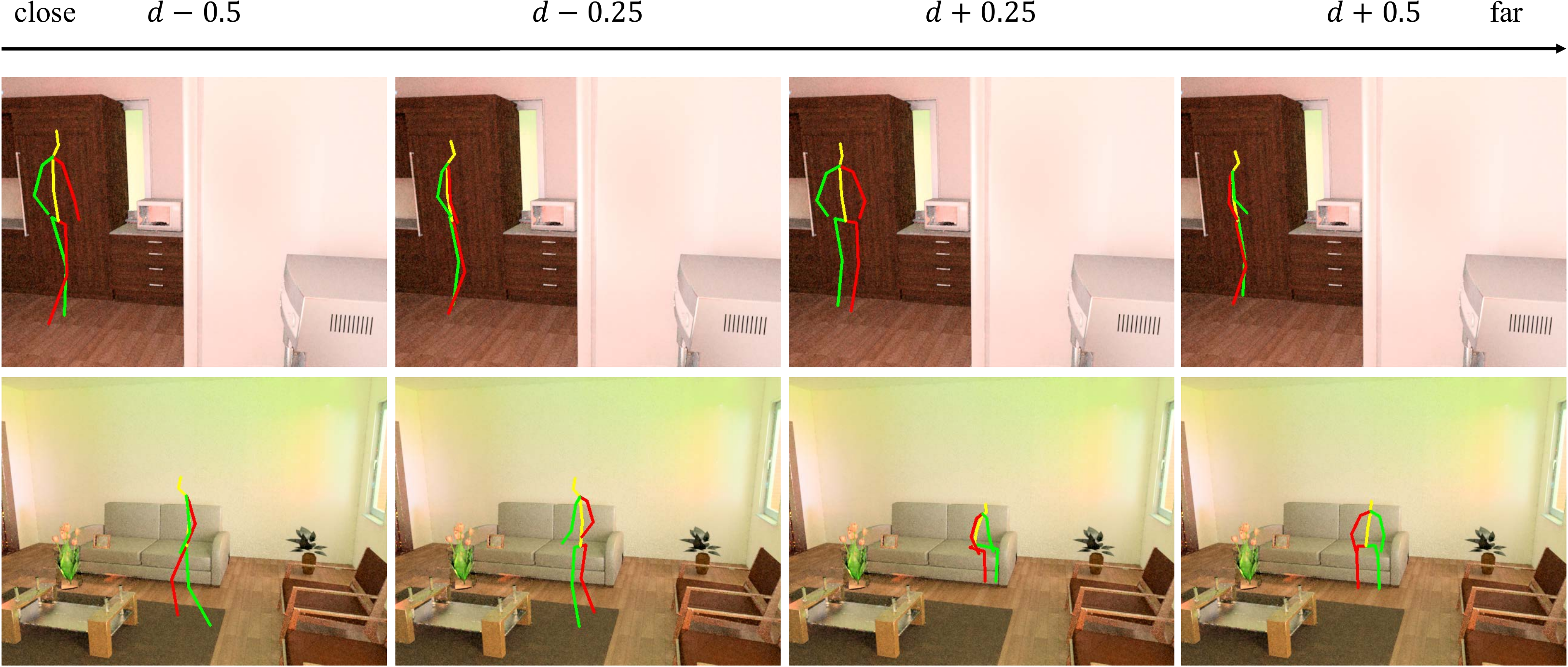} 
	
	\caption{\textbf{Depth interpolation.} In the first scene (top row), closer poses are smaller than farther poses. In the second scene (bottom row), the generated poses change from standing to sitting when depth varies from $d-0.5$ to $d+0.5$, which corresponds well with the sofa in the back of the scene.}
	\label{fig:depth_interp}
	\vspace{-.2cm}
\end{figure}

\begin{figure}[t]

\centering

\begin{tabular}{c@{\hspace{0.005\linewidth}}c@{\hspace{0.005\linewidth}}c@{\hspace{0.005\linewidth}}}
    	\includegraphics[height = .23\linewidth, width = .33\linewidth]{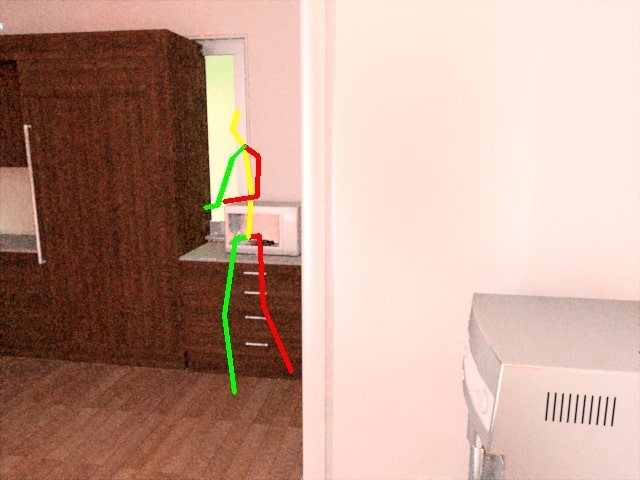} &
    	\includegraphics[height = .23\linewidth, width = .33\linewidth]{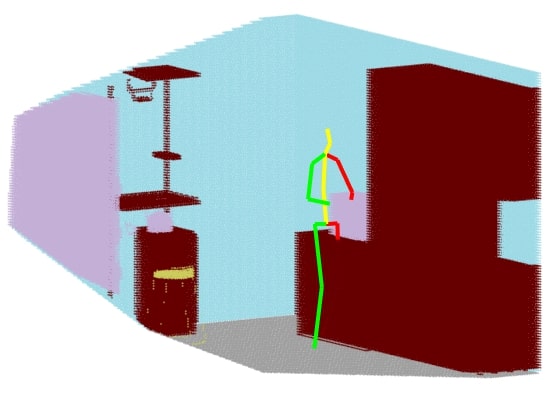} &
    	\includegraphics[height = .23\linewidth, width = .33\linewidth]{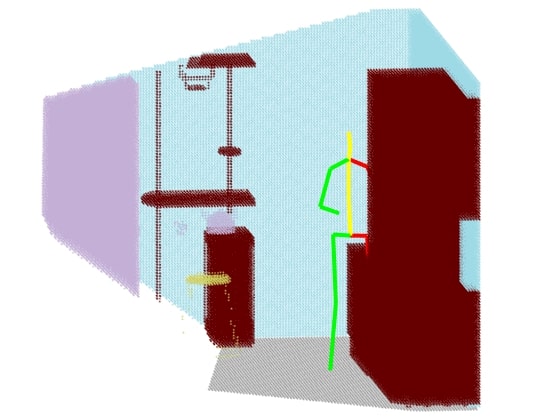} \\
		
		\includegraphics[height = .23\linewidth, width = .33\linewidth]{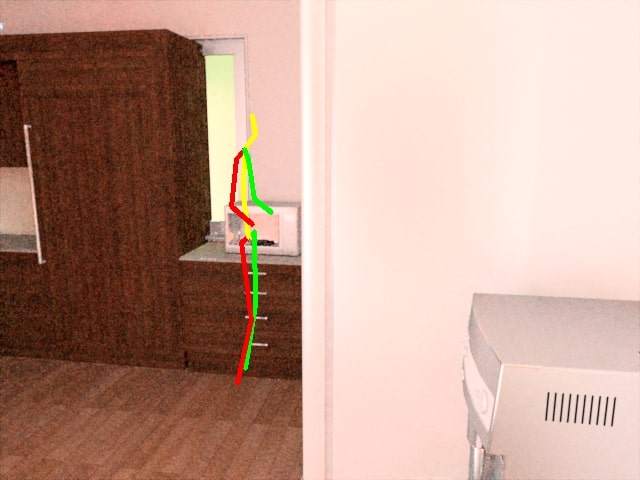} &
    	\includegraphics[height = .23\linewidth, width = .33\linewidth]{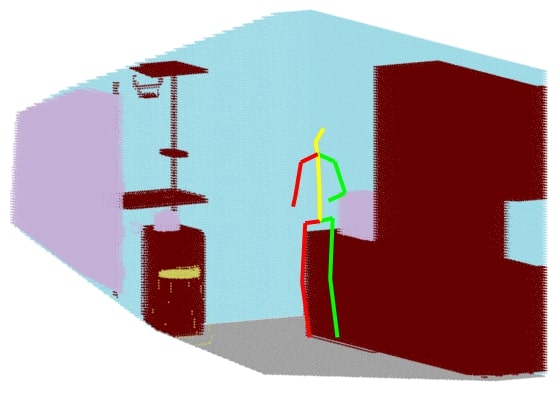} &
    	\includegraphics[height = .23\linewidth, width = .33\linewidth]{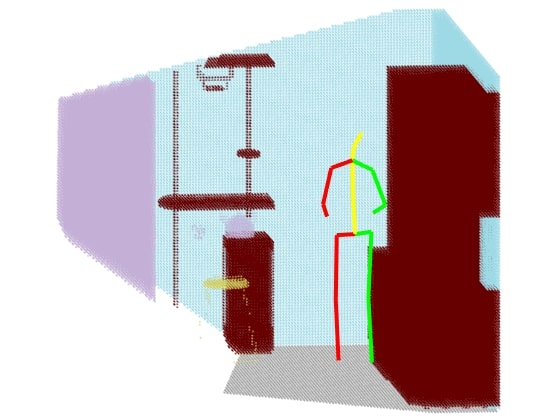} \\

	\end{tabular}

\caption{\textbf{Pose generation at given locations.} We show poses generated by the baseline method~\cite{Wang_affordanceCVPR2017} (top row) and our method (bottom row) at given locations. The first column shows pose projections in scene images, and the last two columns show generated poses in 3D voxels visualized from two different views.}
\label{fig:whatCom}
\vspace{-.2cm}
\end{figure}

In this section, we follow the experimental settings by Wang et al.~\cite{Wang_affordanceCVPR2017} and only focus on pose generation at given locations, i.e., the \emph{what} module. To have a fair comparison, %different from the \emph{what} module introduced in section~\ref{sec:what} which depends on the \emph{3D pelvis coordinate $(x,y,d)$} predicted by the \emph{where} module, here 
we train a \emph{what} module that takes the same inputs as~\cite{Wang_affordanceCVPR2017}, i.e., the \emph{2D pelvis coordinates $(x,y)$} and predicts the coordinates as well as depth for each joint. 
We train the model in~\cite{Wang_affordanceCVPR2017} on the SUNCG dataset with the synthesized poses for the ease of comparison. This model takes the \emph{2D pelvis coordinates $(x,y)$} as our model but only predicts 2D coordinates of each joint. 
%
% Meanwhile, 
% We train our models on our generated dataset from scratch to avoid the domain gap.
%which takes a scene image, 2D pelvis coordinate $(x,y)$ as inputs and predicts a suitable pose based on scene context. 
%
Table~\ref{tab:3dWhat_eval} shows the quantitative scores of these two models. 
Note that we use similar method to calculate geometry score for the baseline method discussed in  Section~\ref{sec:where_what}. As shown in the table, our model achieves $6.66\%$ higher geometry score, indicating that our model performs favorably in generating poses that obey the physical rules in the scene. The same observation can also be found in Fig.~\ref{fig:whatCom}. 
Though given the same location, both the poses generated by our model and the baseline model appear plausible in the 2D image, only our generated pose is geometrically valid when mapped into the 3D scene.

\begin{table}
\centering

  \caption{\textbf{Quantitative evaluation of the  \emph{what} module.} We show comparisons between the baseline model~\cite{Wang_affordanceCVPR2017} and our model with three different input modalities.}
  \label{tab:3dWhat_eval}
  {\scriptsize
  \begin{tabular}{l|l|lll}
  \hline
   \multirow{2}*{Model} & \multirow{2}*{Baseline} & \multicolumn{3}{c}{Ours} \\
    & & RGB&RGB-D&Depth \\ \hline
       
    %{semantic score} & 87.77 & 88.60 & 88.89 & \textbf{90.54} \\
   
    %{geometry score} & 77.17 & 83.83 & 86.26 & \textbf{88.00} \\
    {semantic score} & 91.29 & 91.43 & \textbf{91.86} & 90.86 \\
   
    {geometry score} & 56.29 & 78.43 & 82.00 & \textbf{84.00} \\
	\hline		 
	
  \end{tabular}
  }
%\end{floatrow}
\end{table}

A 2D coordinate $(x,y)$ in a 2D scene image may correspond to multiple locations in the 3D scene with different depth values. 
A model that is able to hallucinate the geometry of a scene should be able to predict different poses at the same $(x,y)$ location with different depth values.
To inspect whether such geometry knowledge has been learned by our \emph{what} module properly, we train another model that only depends on 3D pose locations and scene images.
We particularly remove the pose class $p_c$ in order to eliminate any clue that may indicate the geometrical information.
Other settings are the same as the \emph{what} model described in Section~\ref{sec:where_what}. 
During testing, we fix pelvis coordinates and the input scene image while interpolating depth between $d-0.5$ to $d+0.5$, where $d$ is the ground truth pelvis depth. As we can see in Fig.~\ref{fig:depth_interp}, our model is able to generate poses with different scales and actions that well align with the scene according to different depth values, indicating its ability to hallucinate the 3D geometry of a scene properly.

\vspace{-2mm}
\subsection{Failure Cases}
\begin{figure}[t]
	\centering
	\begin{tabular}{c@{\hspace{0.005\linewidth}}c@{\hspace{0.005\linewidth}}c@{\hspace{0.005\linewidth}}c@{\hspace{0.005\linewidth}}c}
	    \includegraphics[height = .17\linewidth, width = .24\linewidth]{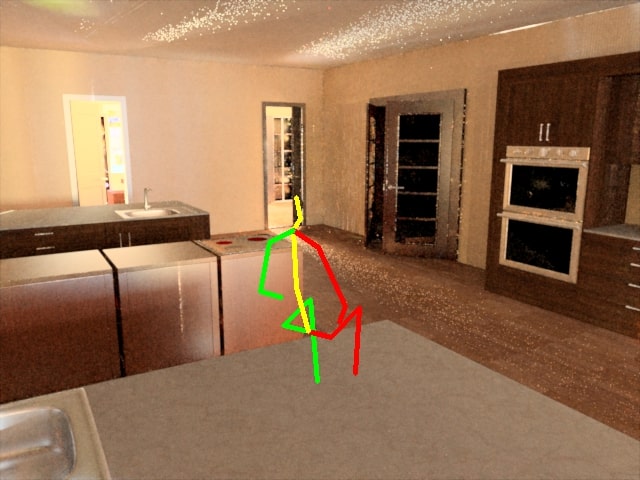} &
    	\includegraphics[height = .17\linewidth, width = .24\linewidth]{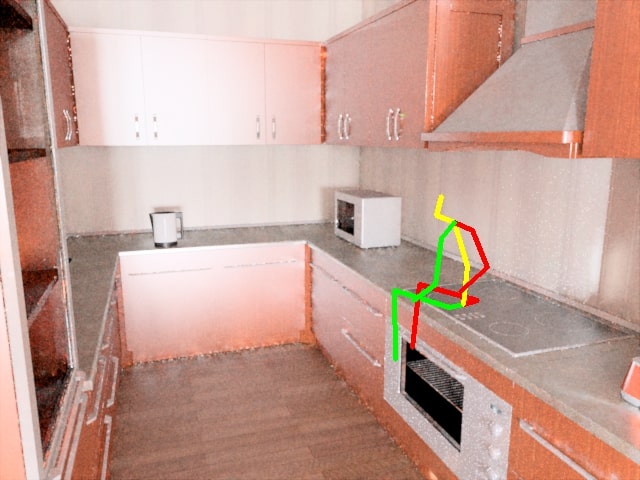} &
    	\includegraphics[height = .17\linewidth, width = .24\linewidth]{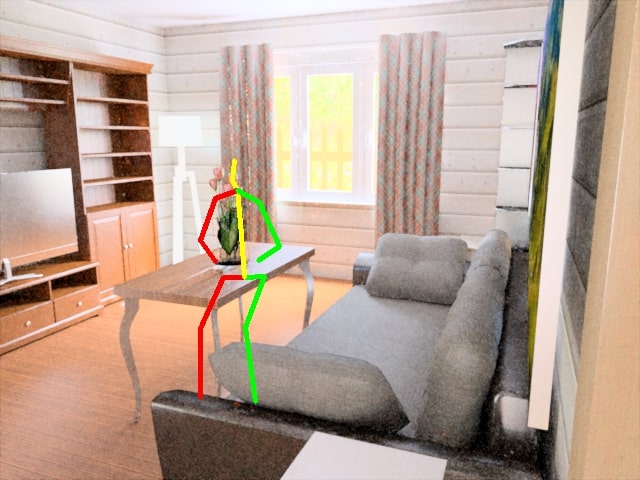} &
		\includegraphics[height = .17\linewidth, width = .24\linewidth]{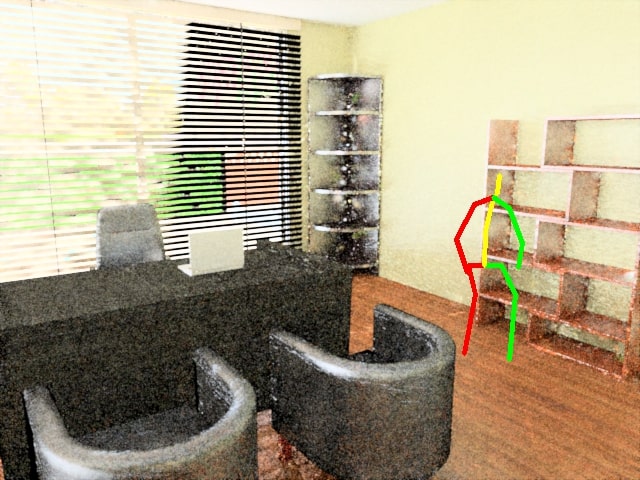} \\
		
		\includegraphics[height = .17\linewidth, width = .24\linewidth]{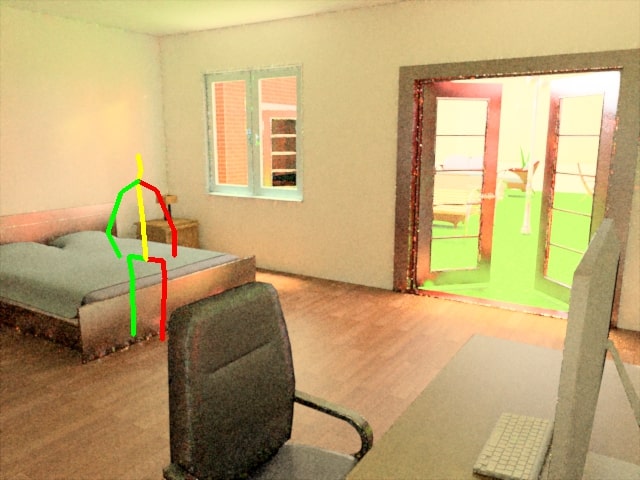} &
    	\includegraphics[height = .17\linewidth, width = .24\linewidth]{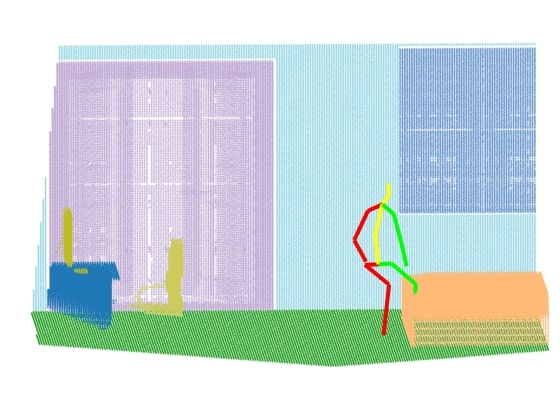} &
    	\includegraphics[height = .17\linewidth, width = .24\linewidth]{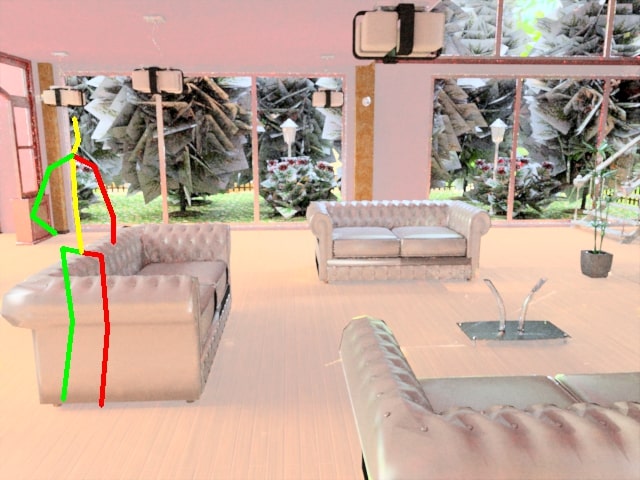} &
		\includegraphics[height = .17\linewidth, width = .24\linewidth]{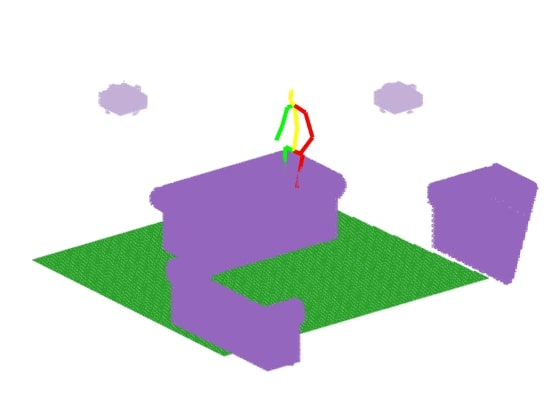} \\

	\end{tabular}
	
	\caption{\textbf{Failure cases.} (Top) Semantic failure: failing to predict socially acceptable poses. (Bottom) Geometric failure: incorrectly hallucinating geometric information of the scene.
	}
	\label{fig:failure}
\end{figure}

Fig.~\ref{fig:failure} shows some failure cases. We mainly have two types of failure cases: (a) generated poses do not align well with the semantic context due to wrong semantic understanding of the scene (e.g., mistakenly sitting on the cabinet) (b) generated poses do not obey geometric rules (e.g., colliding with the objects in a scene). These are caused by a failure of object functionality understanding or 3D geometry hallucination based on 2D information, i.e., ~\emph{reasoning}, which is an interesting open problem for future research.
\vspace{-2mm}

\section{Conclusion}
In this work, we propose to predict \emph{where} and \emph{what} human poses can be put in 3D scenes using a two stage pipeline. We develop a 3D pose synthesizer that can produce millions of ground truth poses in 3D scenes automatically by fusing semantic and geometric knowledge from the Sitcom dataset~\cite{Wang_affordanceCVPR2017} and a 3D scene dataset~\cite{song2017semantic,zhang2017physically}. Then we learn an end-to-end generative model that predicts both locations and gestures of human poses that are semantically plausible and geometrically feasible. Experimental results demonstrate the effectiveness of our proposed method against the stage-of-the-art human affordance prediction method.

\section*{Acknowledgement} We thank Soumyadip Sengupta and Jinwei Gu for providing the SUNCG-PBR dataset.

{\small
\bibliographystyle{ieee}
\bibliography{affordance3d}
}
\clearpage
\begin{appendix}

%We provide additional experiments and details on the original submission. 
%
%Specifically, we explain how to map 2D poses annotated by Wang et al.~\cite{Wang_affordanceCVPR2017} to 3D poses from the Human3.6M dataset~\cite{h36m_pami} in Section~\ref{sec:2d23d}. 
%
%We give detailed derivation on how to estimate pose depth values given generated 2D poses in Section~\ref{sec:depth_est}. 
%
%Then in Section~\ref{sec:pose_in_image}, we discuss the details about our pose location heat map prediction model in the pose synthesizer. 
%
%We further discuss the details w.r.t\ our \emph{where} and \emph{what} modules, as well as the geometry-aware discriminator in Section~\ref{sec:tech_details}. 
%
%Finally, we show more experimental results in Section~\ref{sec:results}.

\section{From 2D Pose to 3D Pose (Section~\ref{sec:2daffordance})}
\begin{figure}[h]
\centering
\includegraphics[width=1\linewidth]{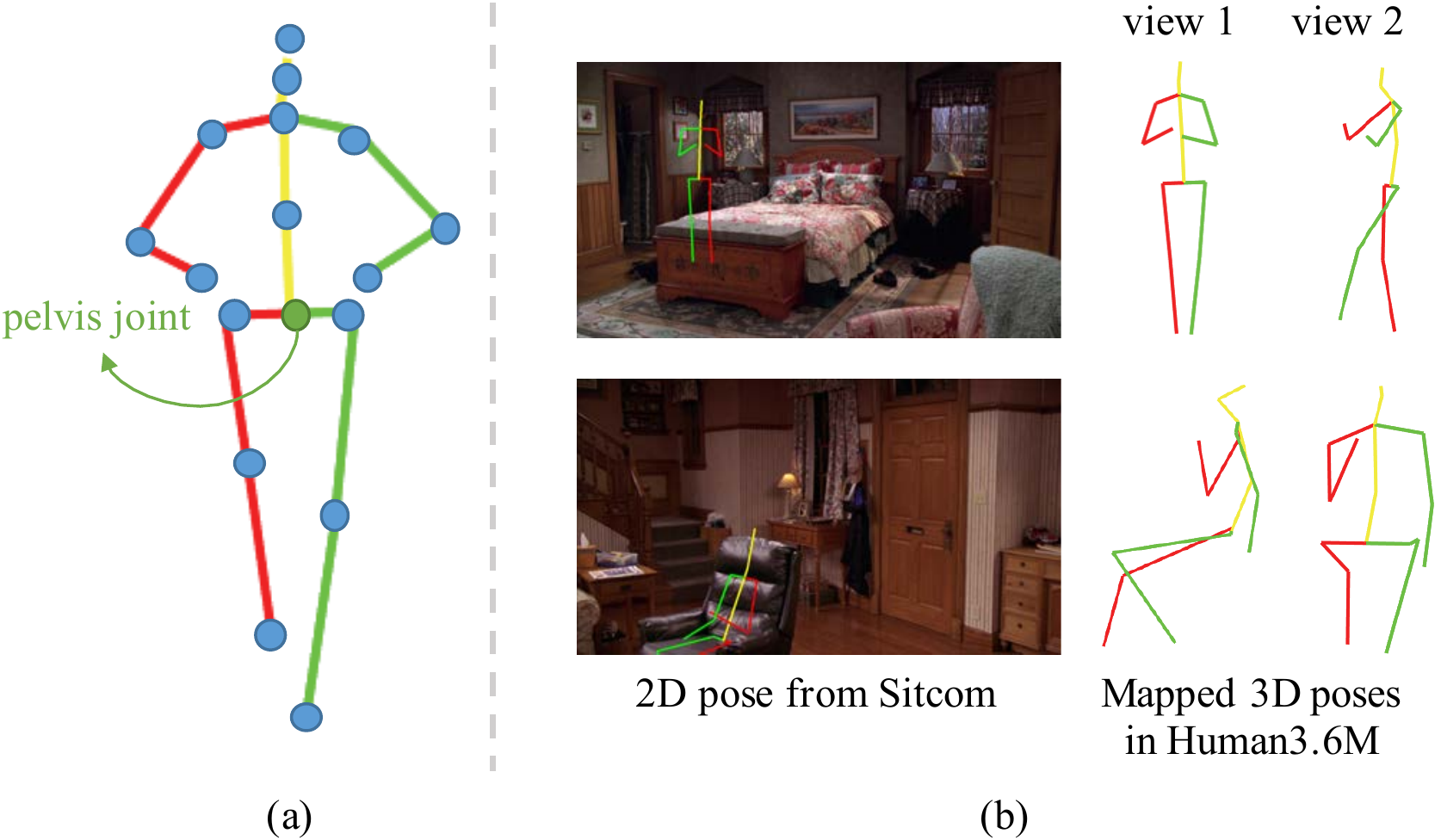}
\caption{(a) We represent each pose using the coordinates of 17 joints. We represent the pose location using the coordinates of the pelvis joint (green dot). (b) We map each 2D pose from~\cite{Wang_affordanceCVPR2017} to a 3D pose in the Human3.6M dataset~\cite{h36m_pami}. The last two columns show corresponding 3D pose of the 2D pose in column 1 visualized from two different views.
}
\label{fig:2d23d}
\end{figure}

\label{sec:2d23d}
As discussed in Section ~\ref{sec:2daffordance}, we map 2D poses annotated by Wang et al.~\cite{Wang_affordanceCVPR2017} to 3D poses in the Human3.6M dataset~\cite{Ionescu36mpami}. This is carried out by first rotating each 3D pose by $\theta$ radian uniformly sampled from $[-\pi,\pi]$, then projecting it onto the $xy$ plane. 
For each 2D pose, we search for its nearest neighbor with minimal Euclidean distance among all projected 2D poses and take the corresponding 3D pose as its 3D mapping.
Fig.~\ref{fig:2d23d}(b) shows examples of mapped 3D poses of 2D poses.

In this work, we represent pose location using the pelvis joint coordinates as shown in Fig.~\ref{fig:2d23d}(a). For pose gesture representation, we use 17 joint coordinates, resulting a 34 dimensional vector for 2D poses and a 51 dimensional vector for 3D poses.

\section{Mapping Poses into 3D Scenes (Section~\ref{sec:mapping})}
\label{sec:depth_est}
\begin{figure}[h]
\centering
\includegraphics[width=1\linewidth]{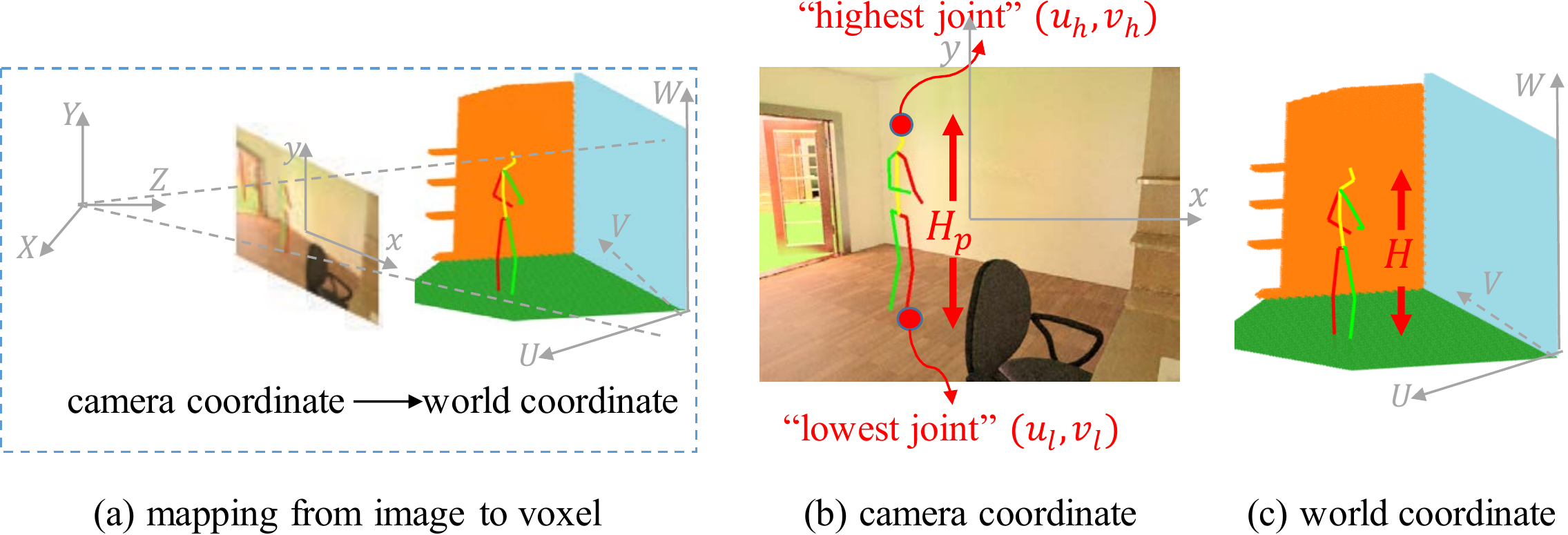}
\caption{(a) Mapping from pixel coordinate system to world coordinate system. (b) Illustrations of human height $H_p$ in pixel coordinates, ``highest joint'' and ``lowest joint'' in the pose. (c) Illustrations of human height in world coordinate system.}
\label{fig:mapping}
\end{figure}

We present more details of how to estimate the depth of a pose (denoted as $d$), given the generated human pose on the image and the approximated human height in the real world, as described in Section~\ref{sec:mapping}. 
We sample human height $H$ in the real world from a Gaussian distribution, i.e., $\mathcal{N}(1.65,0.1)$ for standing poses and $\mathcal{N}(1.20,0.1)$ for sitting poses. 
We denote the 2D coordinates of the ``highest joint'' (usually the head joint) and the ``lowest joint'' (usually the one of the foot joint) as $(u_h,v_h)$ and $(u_l,v_l)$ as shown in Fig.~\ref{fig:mapping}. 
In addition, we denote camera intrinsic matrix $M_i$ and extrinsic matrix $M_e$ as:
\begingroup\small
\begin{align}
\label{eq:derivation}
&M_i = \begin{pmatrix}
    f & 0 & o_x\\
    0 & f & o_y\\
    0 & 0 & 1
\end{pmatrix}
&M_e = \begin{pmatrix}
    r_{11} & r_{12} & r_{13} & t_1\\
    r_{21} & r_{22} & r_{23} & t_2\\
    r_{31} & r_{32} & r_{33} & t_3
\end{pmatrix}\\ \nonumber
\end{align}\endgroup
%\vspace{-3mm}
The world coordinates of joint $(u_h,v_h)$ is calculated by:
\begingroup\small
\begin{align}
\label{eq:world}
\begin{pmatrix}
X_h\\
Y_h\\
Z_h\\
\end{pmatrix} =
\begin{pmatrix}
    r_{11} & r_{12} & r_{13} & t_1\\
    r_{21} & r_{22} & r_{23} & t_2\\
    r_{31} & r_{32} & r_{33} & t_3 
\end{pmatrix}
\begin{pmatrix}
x_{ch}\\
y_{ch}\\
d \\
1
\end{pmatrix}
\end{align}\endgroup
where $(x_{ch},y_{ch},d,1)$ is the camera coordinate of $(u_h,v_h)$, which is calculated by:
\begin{align}
\label{eq:camera}
    &x_{ch} = \frac{(u_h-o_x)d}{f},
    &y_{ch} = \frac{(v_h-o_y)d}{f}
\end{align}
From (\ref{eq:world}) we have $Z_h = r_{31}x_{cl}+r_{32}y_{cl}+r_{33}d+t_1$. 
Similarly, we have the $Z$ coordinate $Z_l = r_{31}x_{cl}+r_{32}y_{cl}+r_{33}d+t_1$ to represent the ``lowest joint''. 
Given the human height $H$ in real world, we have $ H = Z_h - Z_l = r_{31}(x_{ch} - y_{cl}) + r_{32}(y_{ch} - y_{cl})$. 
By substituting (\ref{eq:camera}) into this equation, we have $H = \frac{r_{31}d}{f}(u_h-u_l)+\frac{r_{32}d}{f}(v_h-v_l)$. 
Note that $(u_h - u_l)$ and ($v_h-v_l$) are the pose height $H_p$ and width $W_p$ in the pixel coordinate system as shown in Figure~\ref{fig:2d23d}(b), thus we can calculate pose depth by $d = \frac{H\times f}{r_{31}\times W_p+r_{32}\times H_p}$. 
Specifically, for the SUNCG dataset~\cite{zhang2017physically,song2017semantic}, $r_{32}=0$ for all scenes, we simplify the depth estimation equation above as $d=\frac{H\times f}{r_{32}\times H_p}$, as concluded in Section~\ref{sec:mapping}.

\section{Location Prediction in 2D Scene Images (Section~\ref{sec:2daffordance})}
\label{sec:pose_in_image}
\begin{figure}[h]
\centering
\includegraphics[width=1\linewidth]{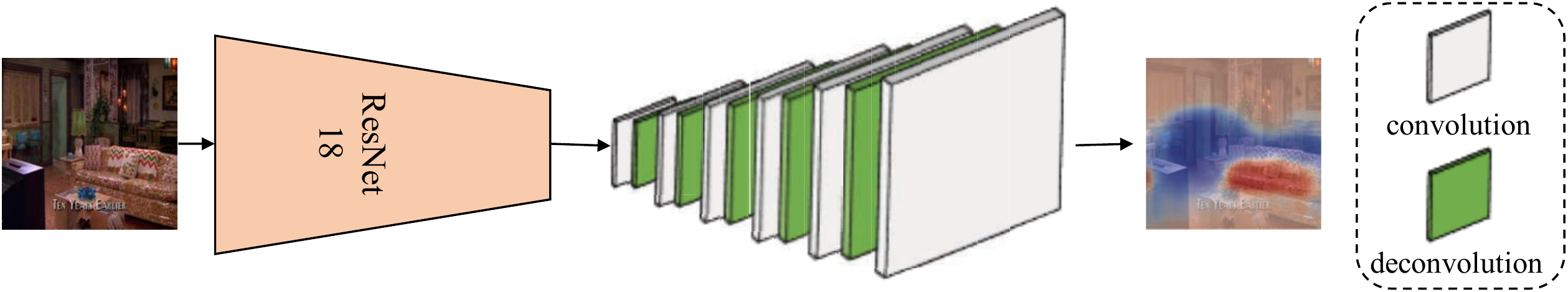}
\caption{Architecture of the location prediction model. The encoder is a 18-layer ResNet~\cite{he2016deep} without the last two fully connected layers. The decoder is a 11-layer CNN with an extra Softmax layer at the end to normalized the generated heat map.}
\label{fig:hp}
\end{figure}

\begin{figure*}[t]
	\centering
\begin{tabular}{c@{\hspace{0.005\linewidth}}c@{\hspace{0.005\linewidth}}c@{\hspace{0.005\linewidth}}c@{\hspace{0.005\linewidth}}c@{\hspace{0.005\linewidth}}c@{\hspace{0.005\linewidth}}}
    	\includegraphics[height = .11\linewidth, width = .18\linewidth]{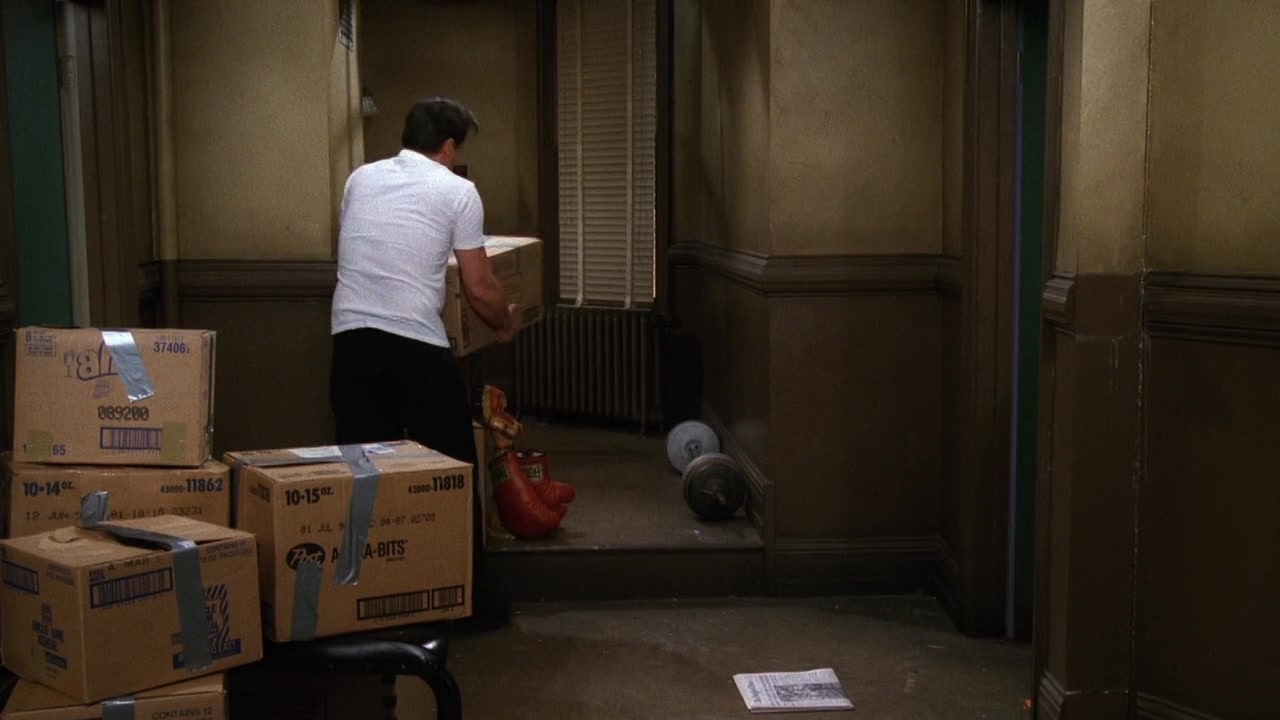} &
    	\includegraphics[height = .11\linewidth, width = .18\linewidth]{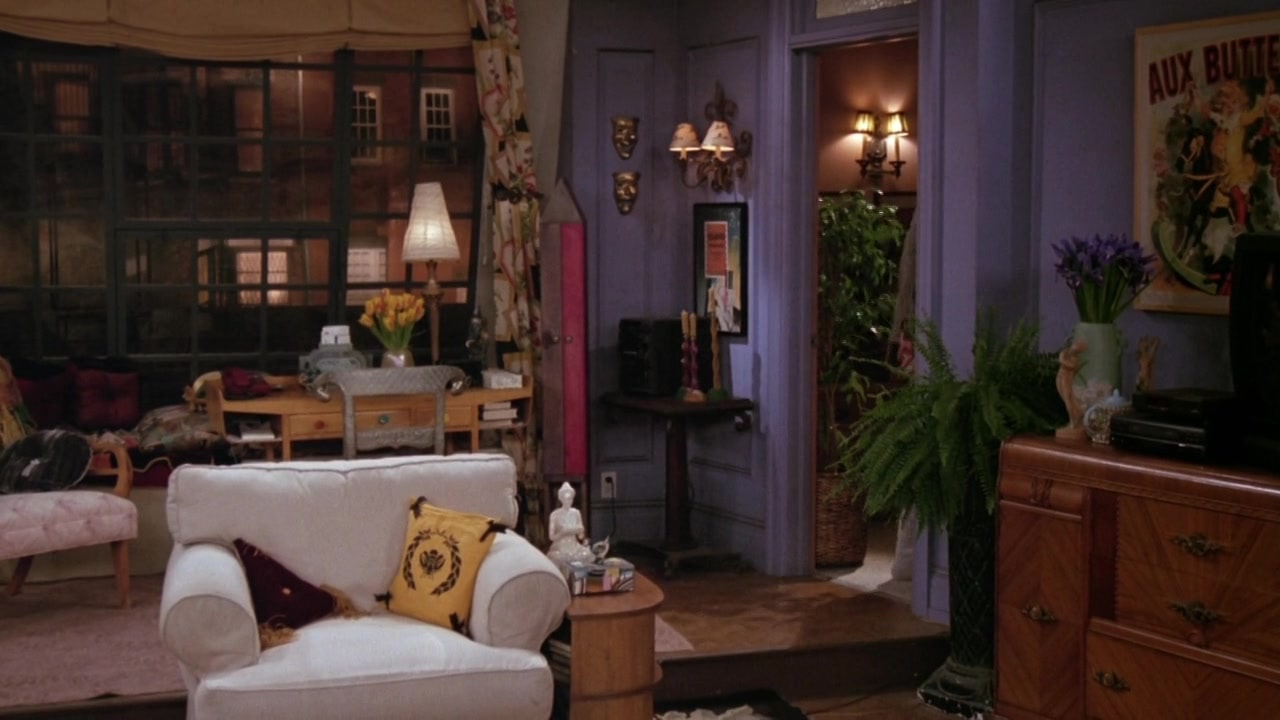} &
    	\includegraphics[height = .11\linewidth, width = .18\linewidth]{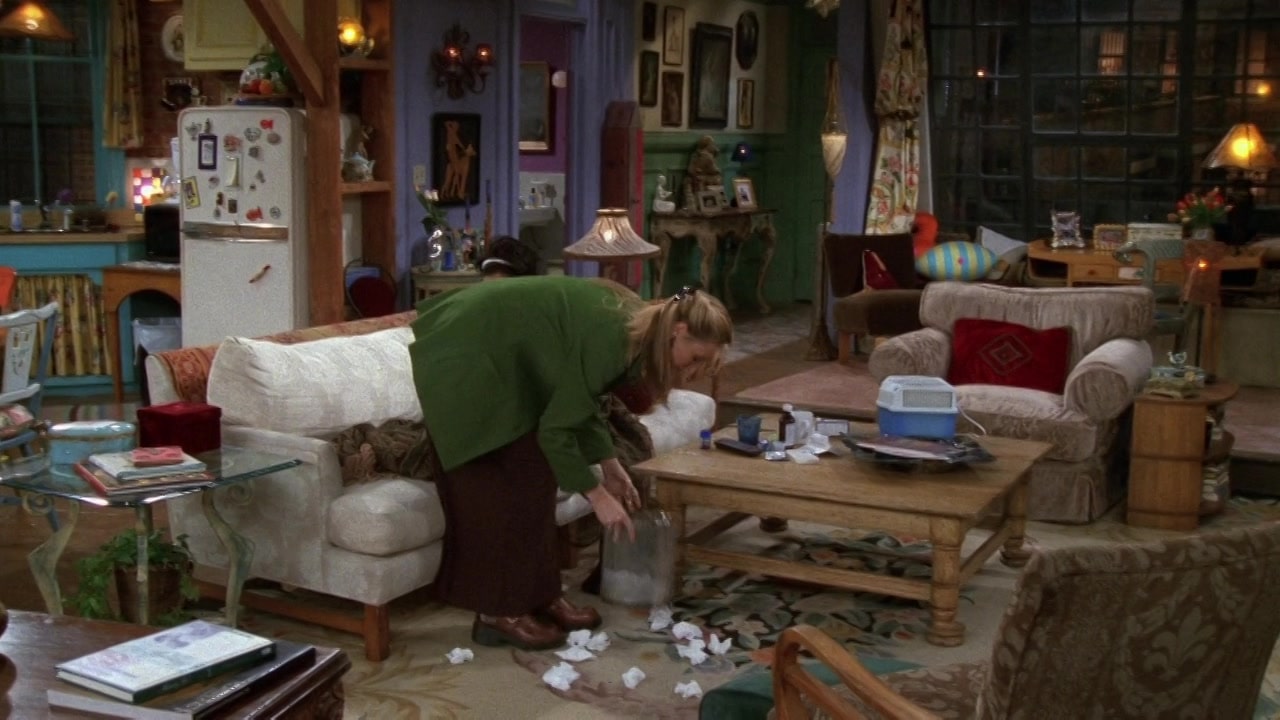} &
		\includegraphics[height = .11\linewidth, width = .15\linewidth]{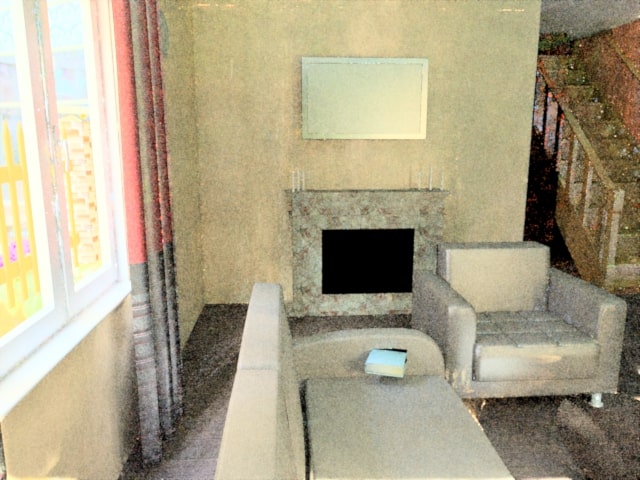} &
		\includegraphics[height = .11\linewidth, width = .15\linewidth]{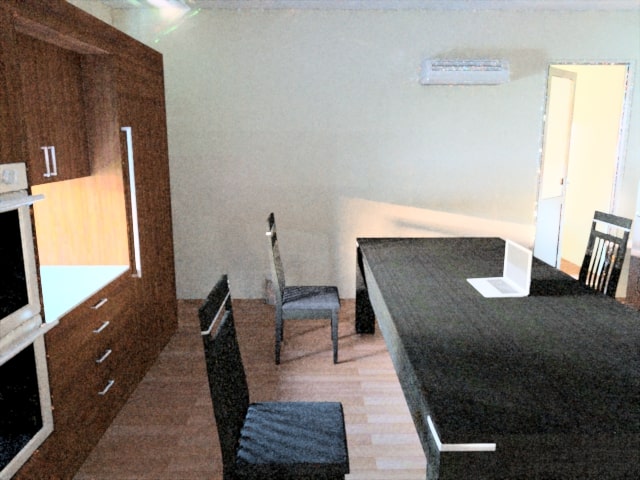} &
		\includegraphics[height = .11\linewidth, width = .15\linewidth]{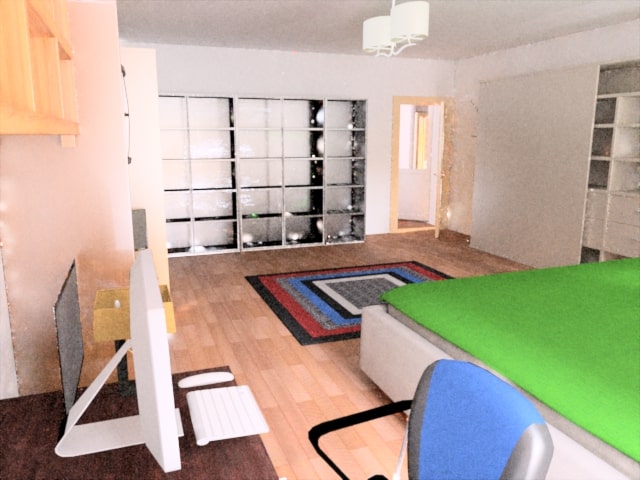}  \\
		
		\includegraphics[height = .11\linewidth, width = .18\linewidth]{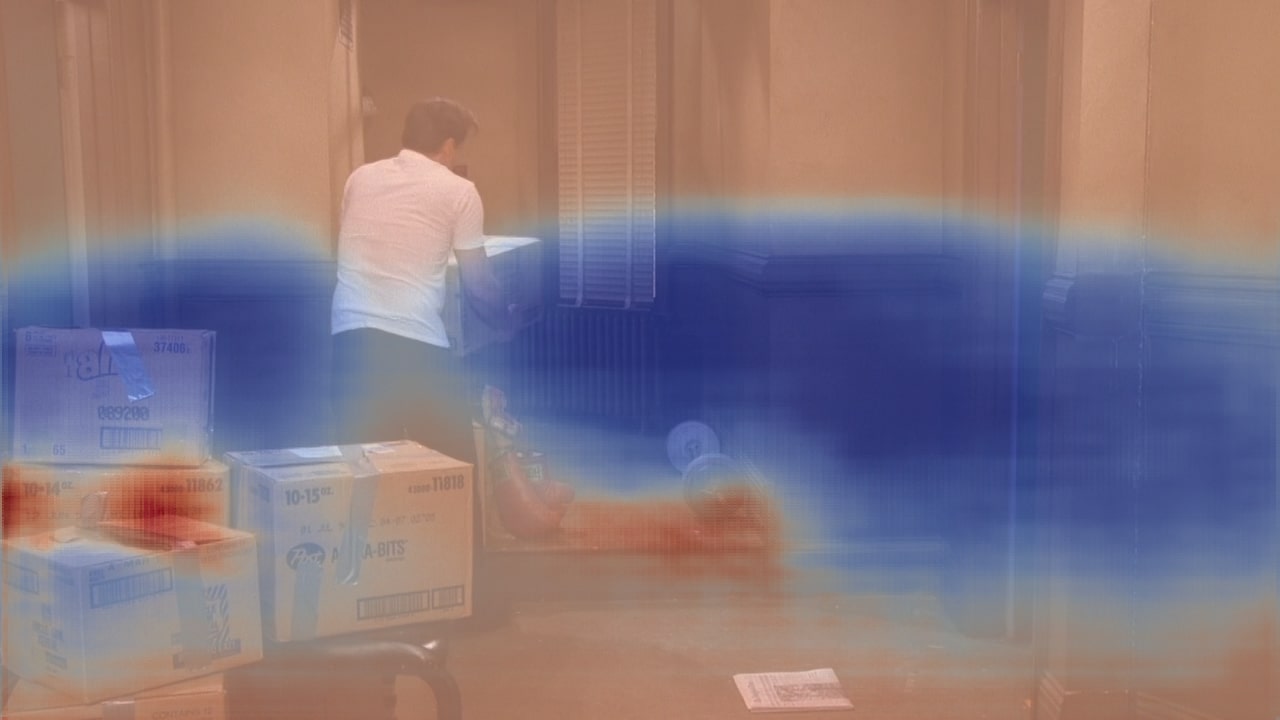} &
    	\includegraphics[height = .11\linewidth, width = .18\linewidth]{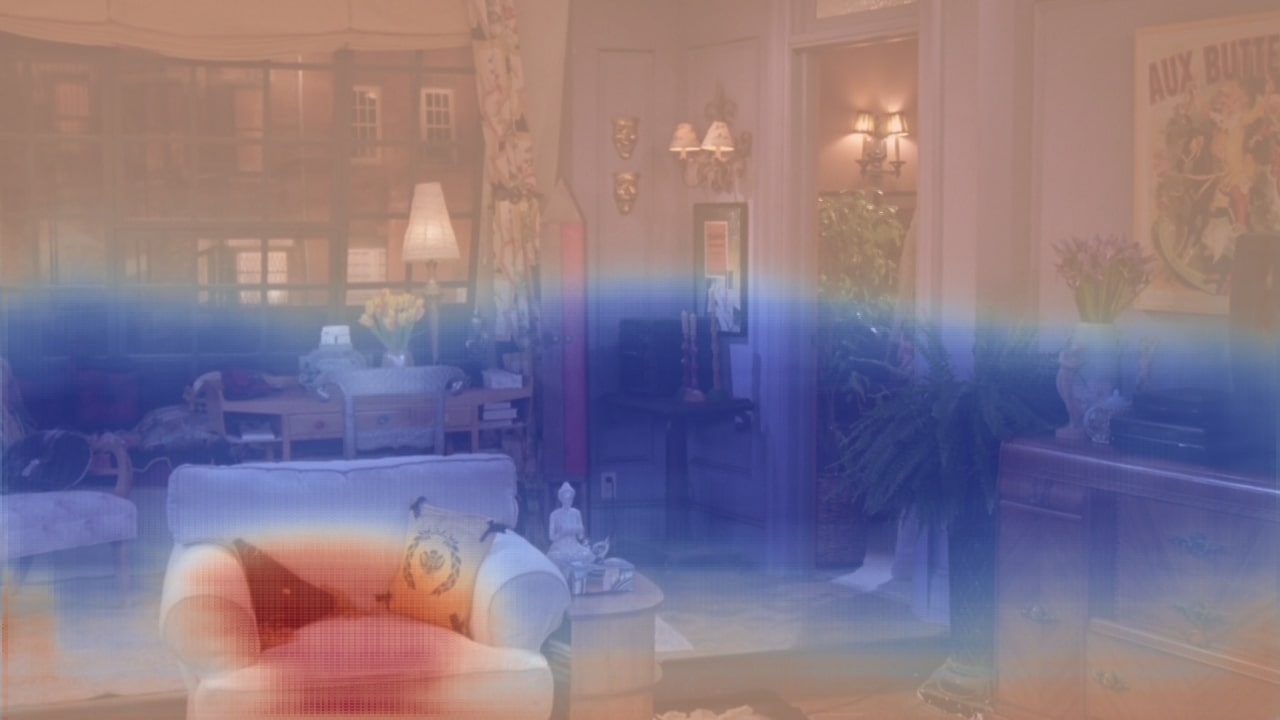} &
    	\includegraphics[height = .11\linewidth, width = .18\linewidth]{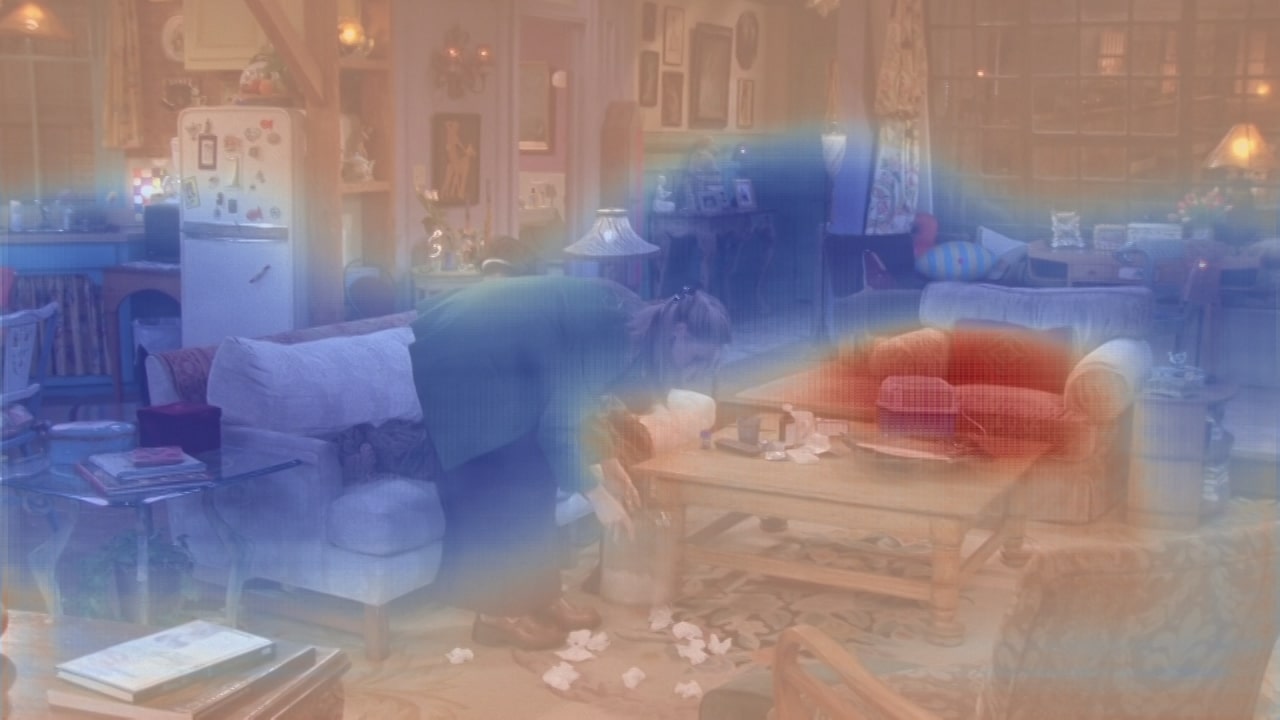} &
		\includegraphics[height = .11\linewidth, width = .15\linewidth]{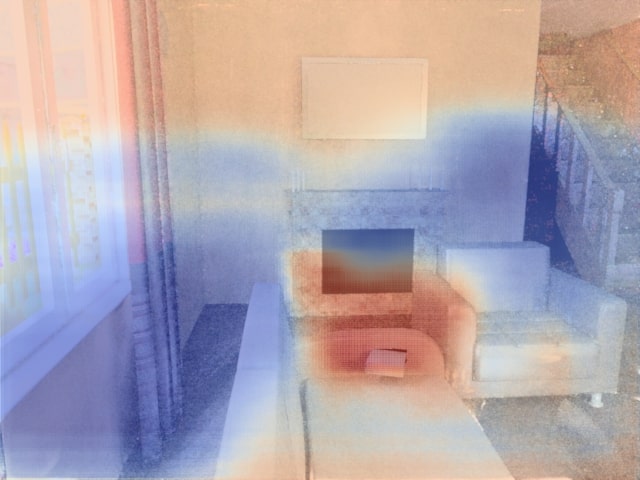} &
		\includegraphics[height = .11\linewidth, width = .15\linewidth]{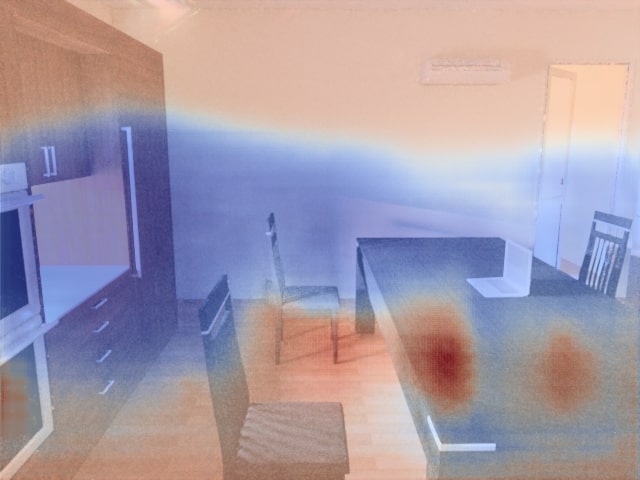} &
		\includegraphics[height = .11\linewidth, width = .15\linewidth]{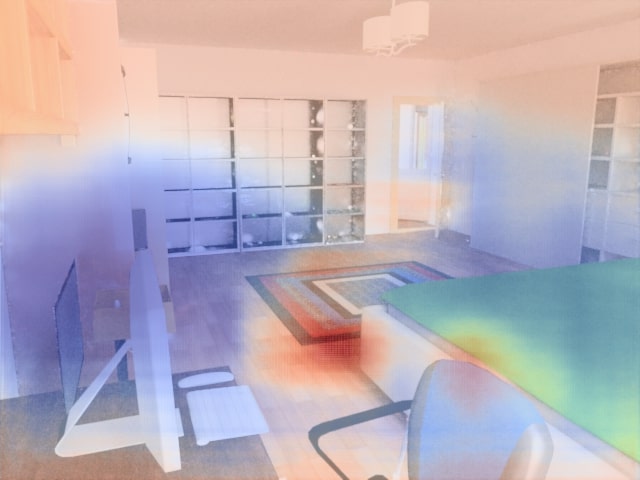} \\
		\includegraphics[height = .11\linewidth, width = .18\linewidth]{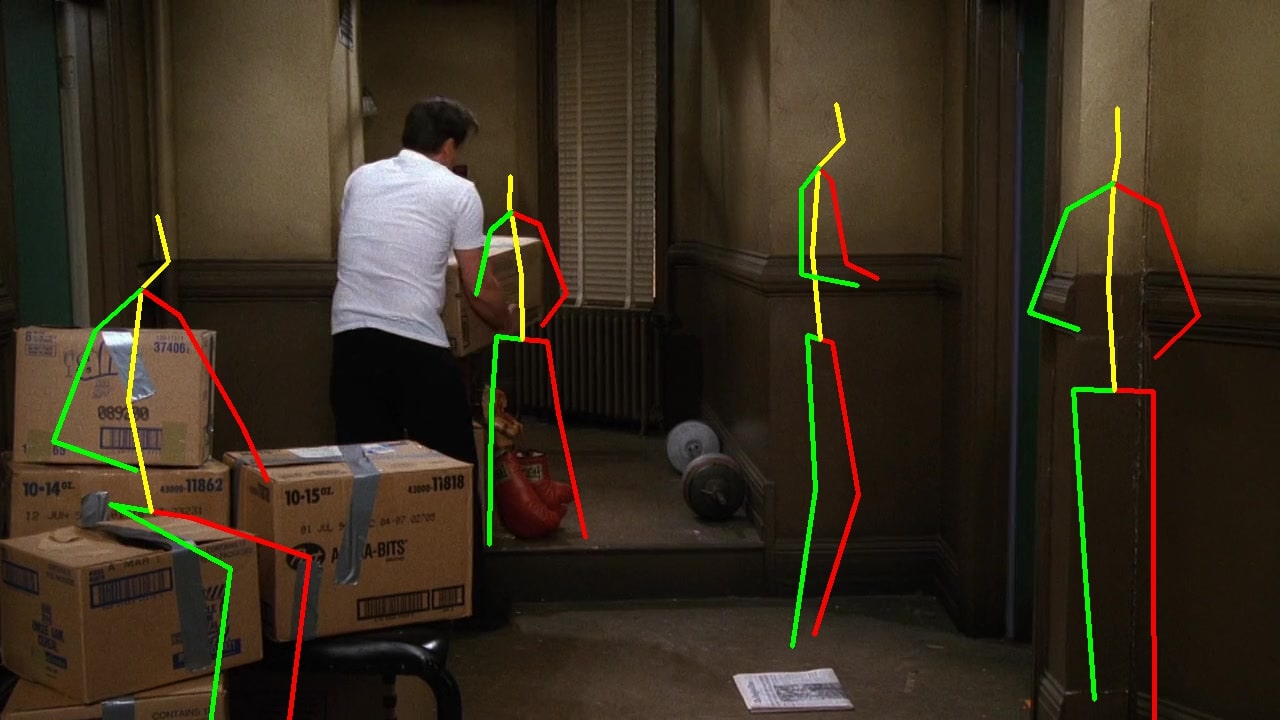} &
    	\includegraphics[height = .11\linewidth, width = .18\linewidth]{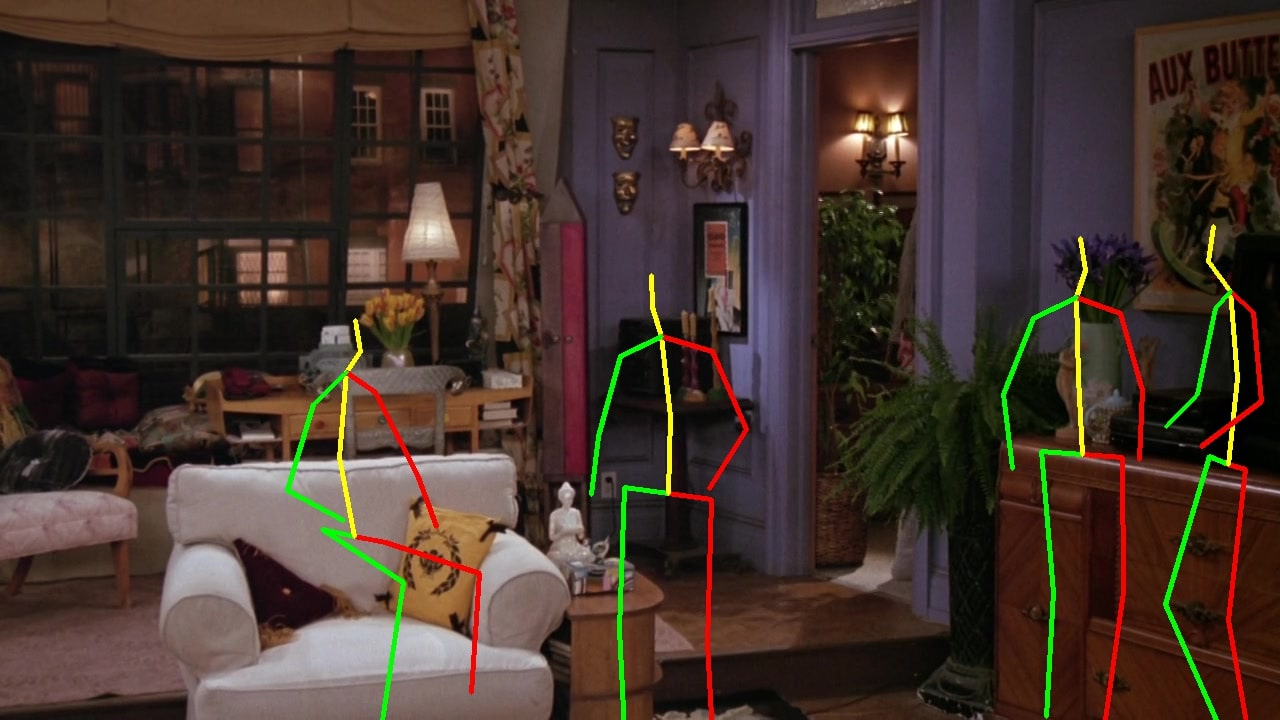} &
    	\includegraphics[height = .11\linewidth, width = .18\linewidth]{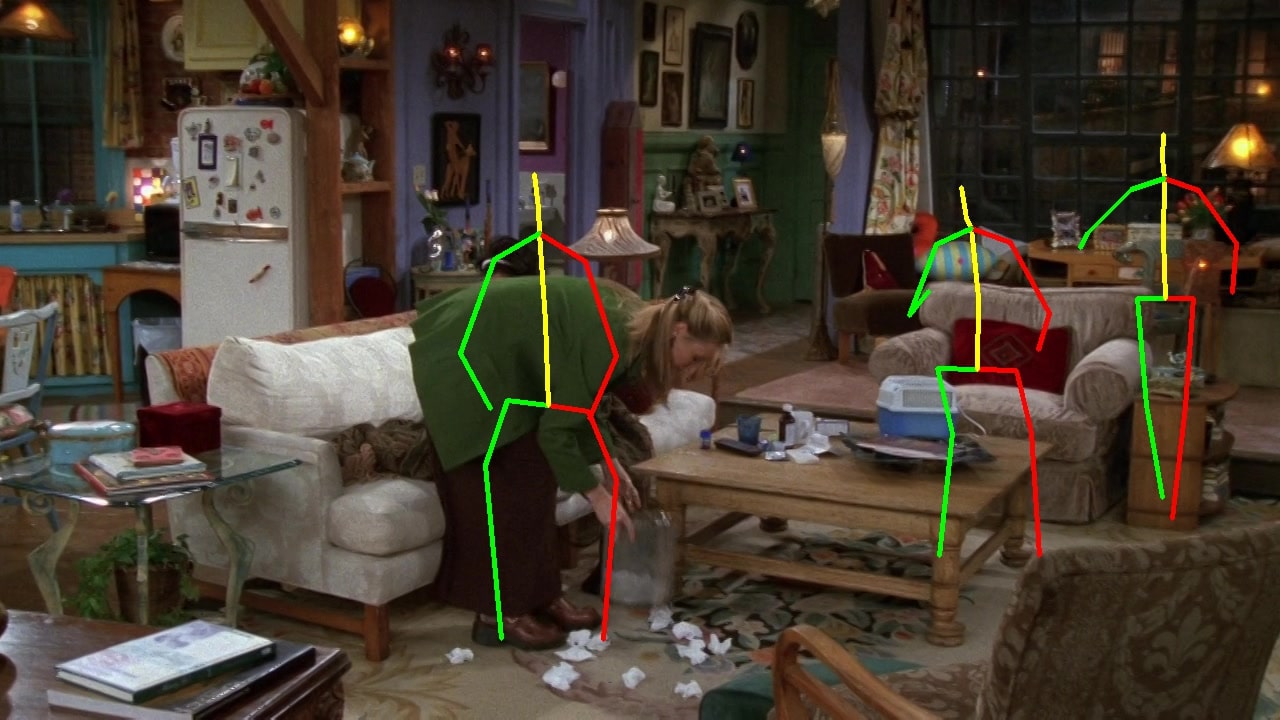} &
		\includegraphics[height = .11\linewidth, width = .15\linewidth]{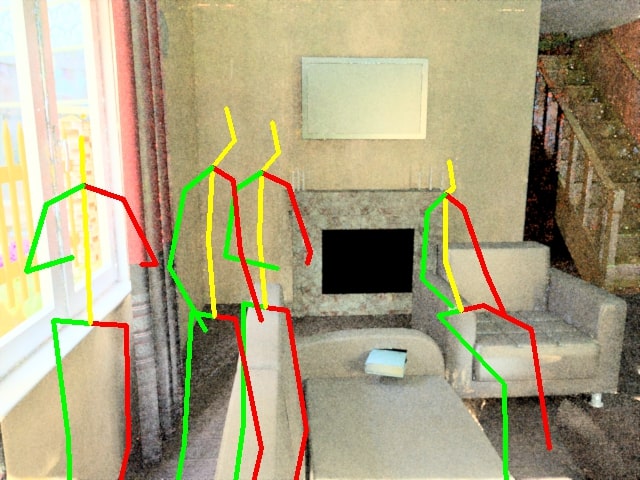} &
		\includegraphics[height = .11\linewidth, width = .15\linewidth]{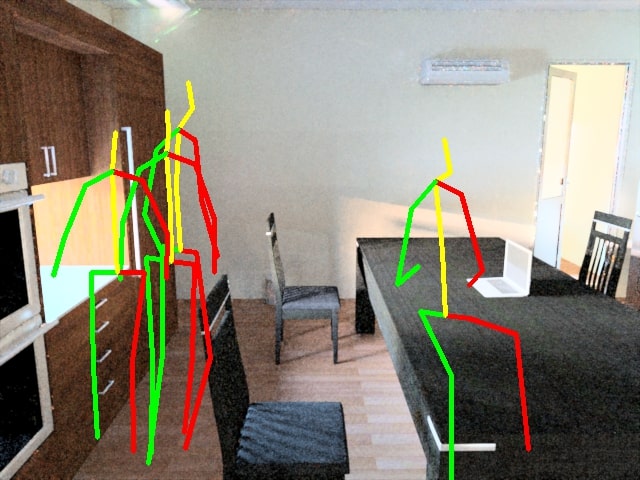} &
		\includegraphics[height = .11\linewidth, width = .15\linewidth]{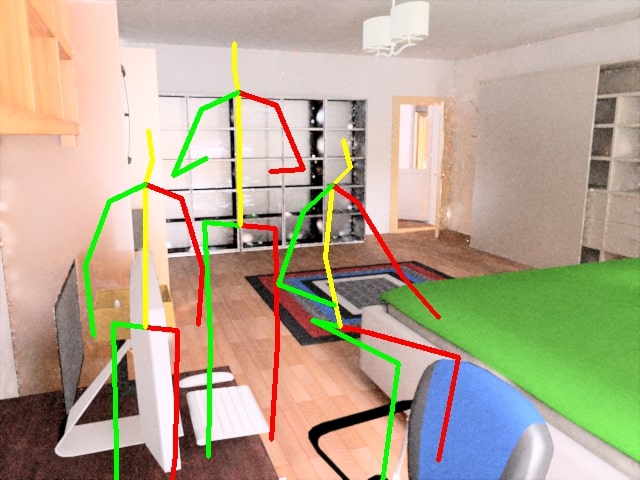}
		
	\end{tabular}
	
	\caption{Predicted pose location heat maps and sampled poses. First three columns show results on Sitcom and last three columns show results on SUNCG. For visualization purpose, we summarize area suitable for sitting pose location (shown as red area), area suitable for standing pose location (shown as blue) as well as area not suitable for any human poses (shown as light yellow) in predicted heat maps. All poses shown in the bottom row are projections of 3D poses generated by our model.
	}
	\label{fig:stage1}
\end{figure*}

Fig.~\ref{fig:hp} illustrates the structure of our 2D pelvis location prediction model, as discussed in Section~\ref{sec:2daffordance}. 
Fig.~\ref{fig:stage1} shows predicted heat maps and poses for the Sitcom~\cite{Wang_affordanceCVPR2017} and the SUNCG dataset. 
We train the heat map prediction model for $5,000$ iterations using the Adam~\cite{kinga2015method} solver. 
For data augmentation, we randomly crop a $384\times 384$ patch from a $448\times 448$ image, we set batch size to $100$ and learning rate to $0.001$. For pose generation at given locations, we use the same model as~\cite{Wang_affordanceCVPR2017}. Note that instead of predicting 2D poses, we directly predict 3D poses obtained via 2D to 3D pose mapping, as described in Appendix~\ref{sec:2d23d} and Section~\ref{sec:2daffordance}.

\section{More Details for 3D Pose Prediction (Section~\ref{sec:3daffordance})}
\label{sec:tech_details}
\begin{figure*}[h]
\centering
\includegraphics[width=0.9\linewidth]{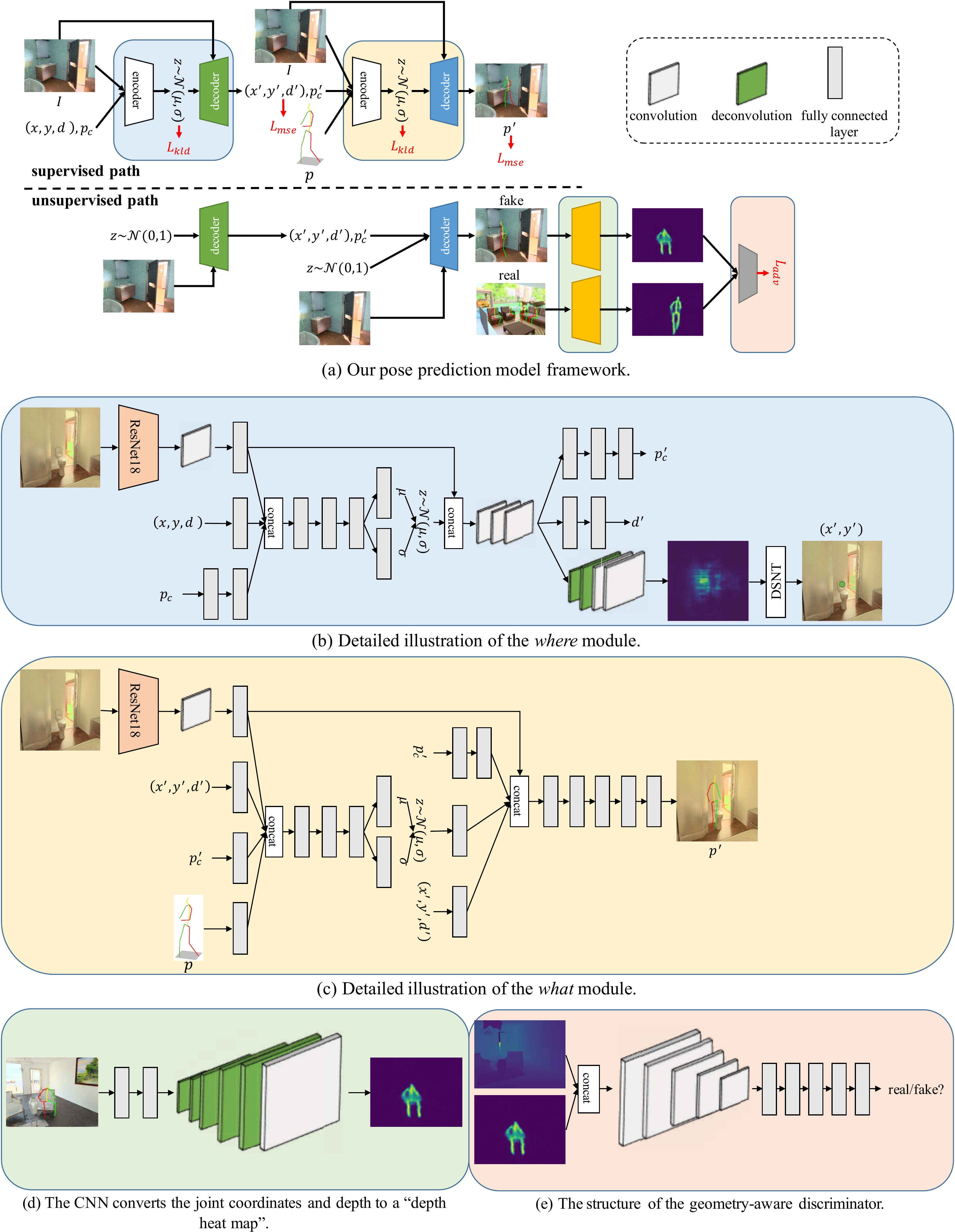}
\caption{Detailed structure of each block in our pose prediction model. We show the overview of our pose prediction model, including the \emph{supervised} and \emph{unsupervised} path explained in Appendix~\ref{sec:tech_details} in Fig. (a). Detailed structure of each block is illustrated in (b), (c), (d) and (e) respectively, same blocks are filled with same background color.}
\label{fig:arch_detail}
\end{figure*}

\paragraph{The \emph{where} and \emph{what} modules.}
We first train the \emph{where} and \emph{what} module discussed in Section~\ref{sec:where} and~\ref{sec:what} for $80000$ iterations using the Adam~\cite{kinga2015method} solver. 
Specifically, we set the batch size as $100$ and the learning rate as $0.0001$. 
Then, we connect the two modules and jointly finetune them with the geometry-aware discriminator, as introduced in Section~\ref{sec:joint}, for another $50000$ iterations.
We adopt the similar training strategy as Lee et al.~\cite{donghoon2018} and only use the discriminator to regularize an \emph{unsupervised path} for both modules, i.e., the discriminator is used to regularize the distributions of generated poses that coming from the random noises, instead of interacting with the VAE block in a direct manner.
We observe that such network architecture brings significant improvement to the generated results. 
% we only take the generated pose from random noise and scene image as a fake pose. 
%
Fig.~\ref{fig:arch_detail} (a) shows the detailed structure of our \emph{supervised} and \emph{unsupervised} path and Fig.~\ref{fig:arch_detail} (b), (c) shows the detailed structure of our \emph{where} and \emph{what} module.

\paragraph{Geometry-aware discriminator.} 
As discussed in Section~\ref{sec:joint}, we propose a geometry-aware discriminator to further regularize the generator to generate poses that obey the rules of geometry in a scene. However, it is challenging for the discriminator to associate joint coordinates, i.e., a 3-dimensional tensor, with the image. 
Therefore, we first train a CNN to convert the coordinates and depth of joints, into a ``depth heat map'' that has the same dimension as the input image.
Fig.~\ref{fig:arch_detail}(d) illustrates the structure of this CNN. We train the CNN for $5000$ iterations using the Adam~\cite{kinga2015method} solver with a learning rate of $0.0002$. Fig.~\ref{fig:arch_detail}(e) further shows the detailed structure of our geometry-aware discriminator.

\begin{figure*}[t]
	\centering
	\begin{tabular}{c@{\hspace{0.005\linewidth}}c@{\hspace{0.005\linewidth}}c@{\hspace{0.005\linewidth}}c@{\hspace{0.005\linewidth}}c@{\hspace{0.005\linewidth}}}
	    \includegraphics[height = .13\linewidth, width = .18\linewidth]{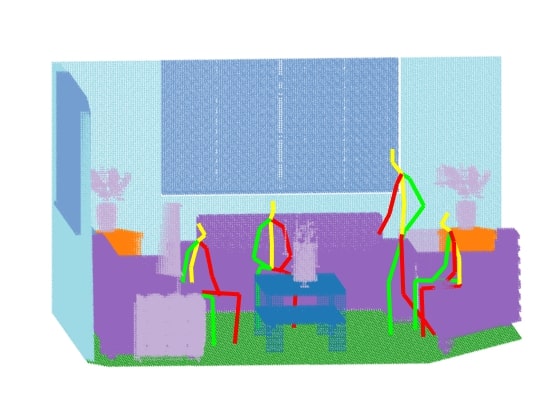} &
    	\includegraphics[height = .13\linewidth, width = .18\linewidth]{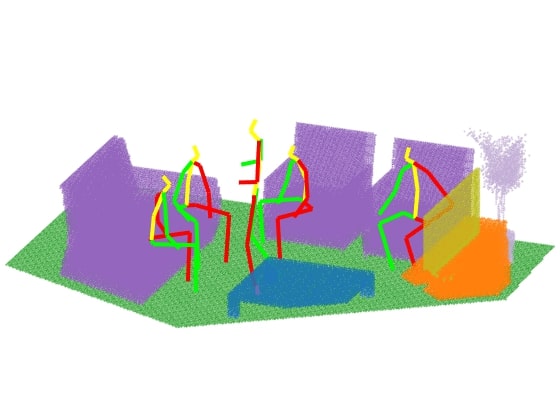} &
    	\includegraphics[height = .13\linewidth, width = .18\linewidth]{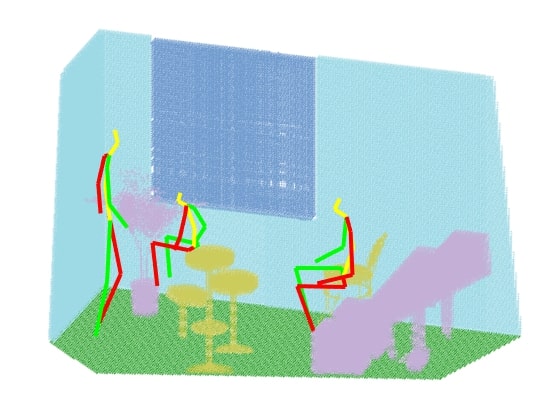} &
		\includegraphics[height = .13\linewidth, width = .18\linewidth]{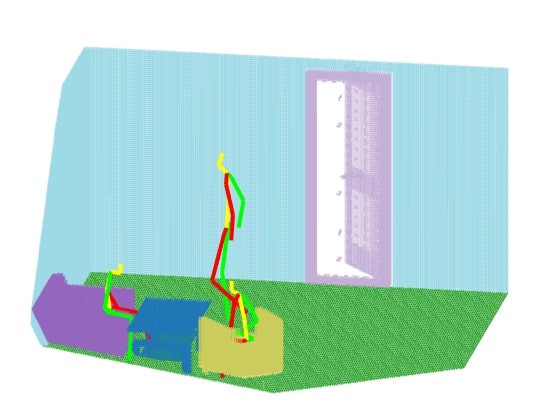} 
		\includegraphics[height = .13\linewidth, width = .18\linewidth]{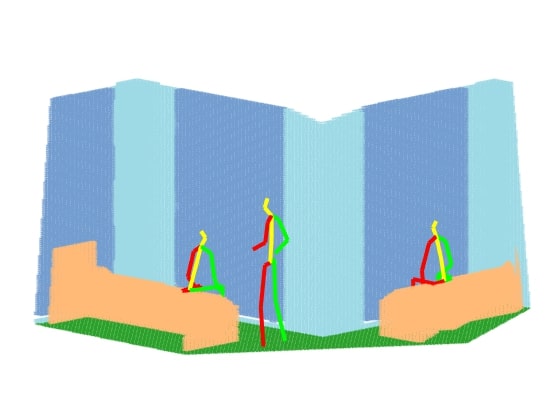} &\\
		
    	\includegraphics[height = .13\linewidth, width = .18\linewidth]{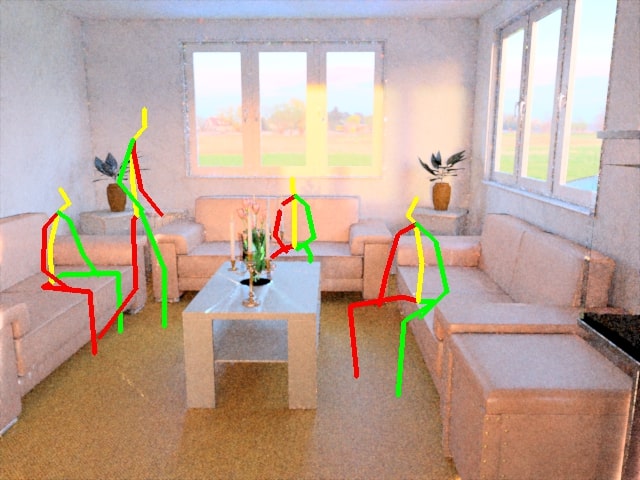} &
   	\includegraphics[height = .13\linewidth, width = .18\linewidth]{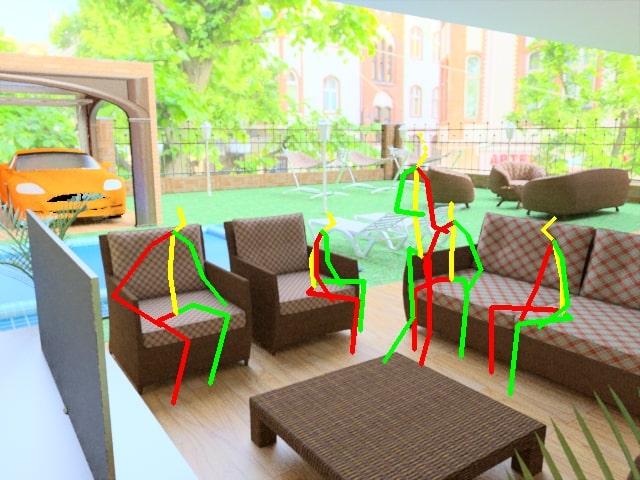} &
    	\includegraphics[height = .13\linewidth, width = .18\linewidth]{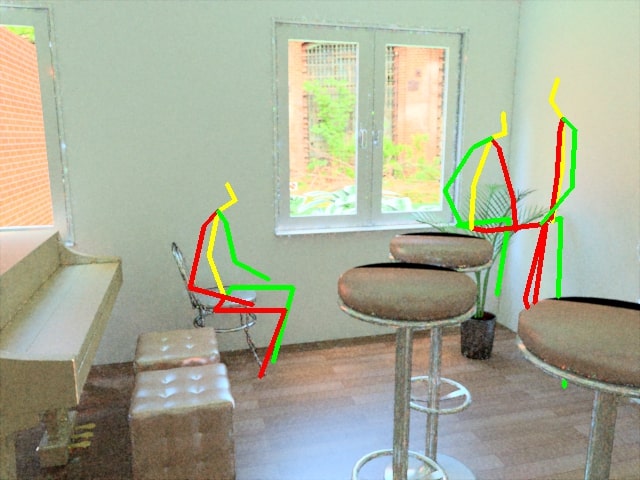} &
		\includegraphics[height = .13\linewidth, width = .18\linewidth]{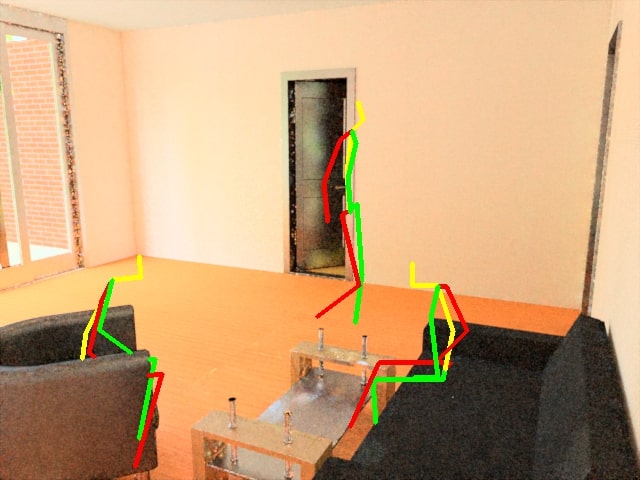} 
		\includegraphics[height = .13\linewidth, width = .18\linewidth]{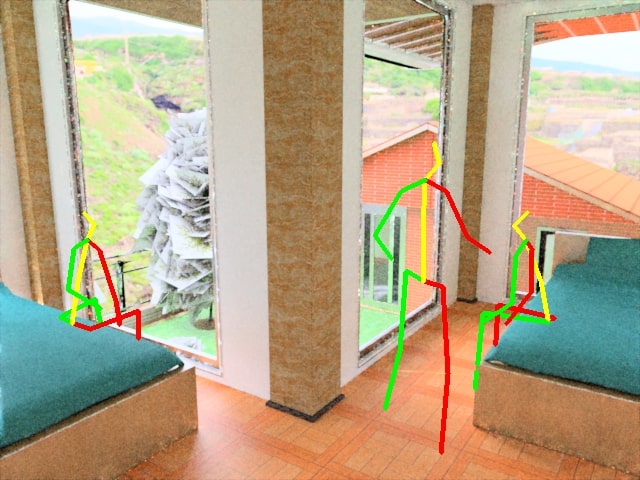} &\\
		
	\end{tabular}
	
	\caption{Samples of synthesized poses. (Top) Sample poses shown in 3D scene geometry, and (Bottom) rendered images of the corresponding scenes. Note that the generated poses contain information about occlusion in the scene.
	}
	\label{fig:suncg_gt}
\end{figure*}

\begin{figure*}[t]
	\centering
\begin{tabular}{c@{\hspace{0.005\linewidth}}c@{\hspace{0.005\linewidth}}c@{\hspace{0.005\linewidth}}c@{\hspace{0.005\linewidth}}c@{\hspace{0.005\linewidth}}c@{\hspace{0.005\linewidth}}}
    	\includegraphics[height = .11\linewidth, width = .18\linewidth]{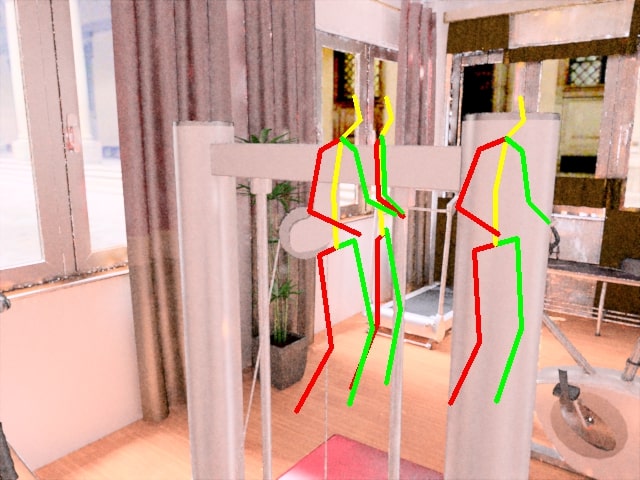} &
    	\includegraphics[height = .11\linewidth, width = .18\linewidth]{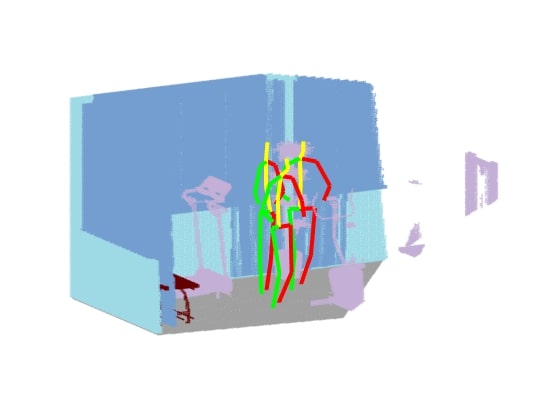} &
    	\includegraphics[height = .11\linewidth, width = .18\linewidth]{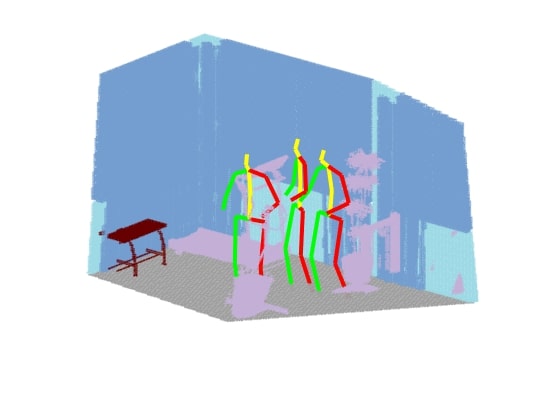} &
		\includegraphics[height = .11\linewidth, width = .15\linewidth]{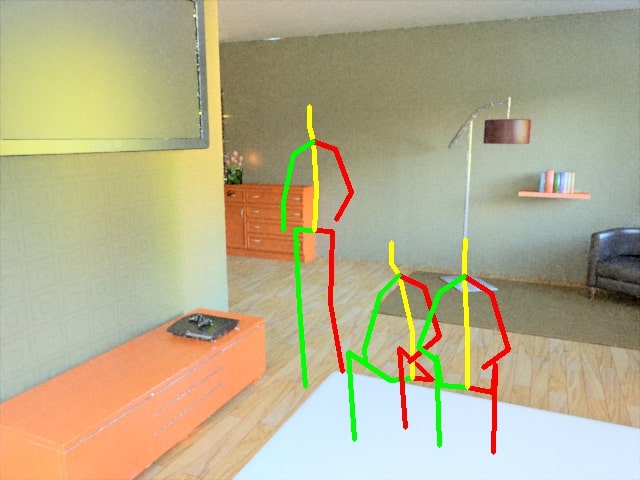} &
		\includegraphics[height = .11\linewidth, width = .15\linewidth]{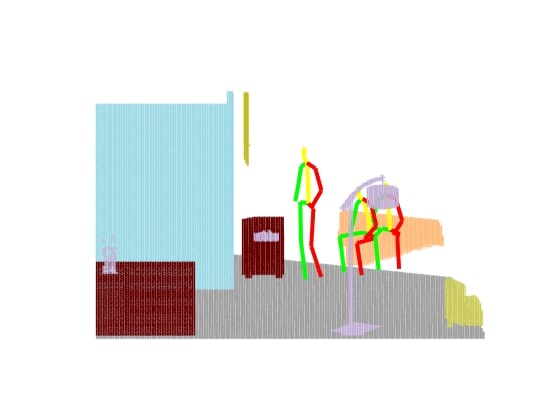} &
		\includegraphics[height = .11\linewidth, width = .15\linewidth]{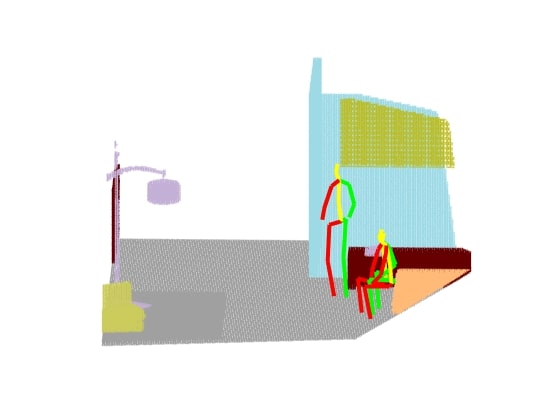}  \\
		
		\includegraphics[height = .11\linewidth, width = .18\linewidth]{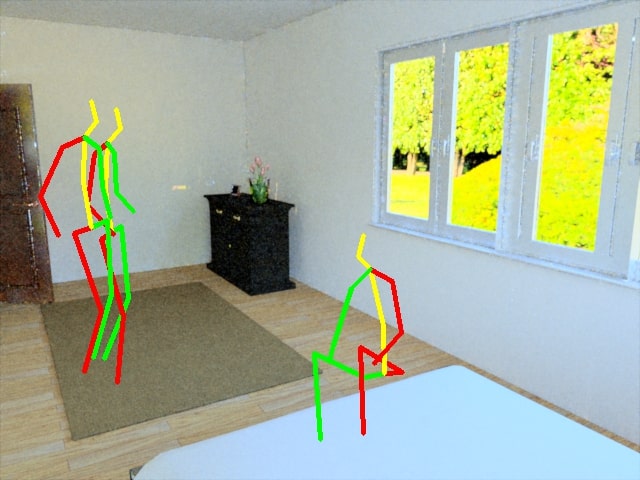} &
    	\includegraphics[height = .11\linewidth, width = .18\linewidth]{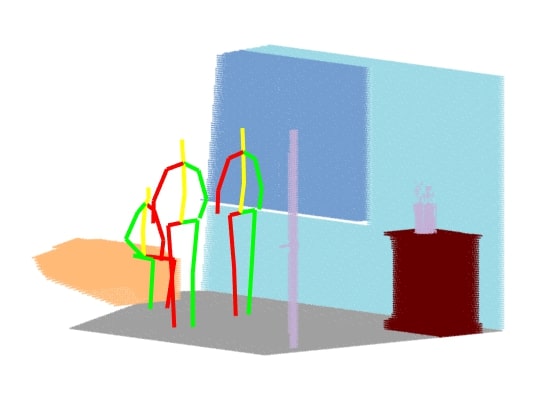} &
    	\includegraphics[height = .11\linewidth, width = .18\linewidth]{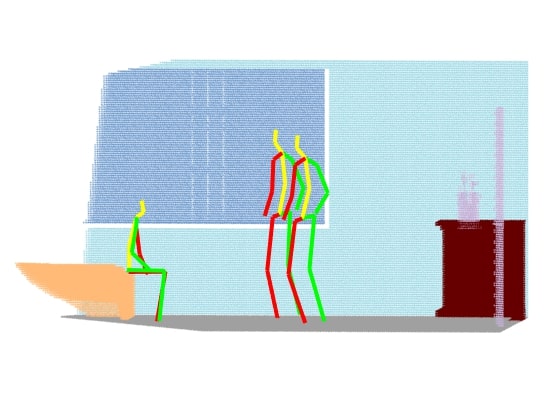} &
		\includegraphics[height = .11\linewidth, width = .15\linewidth]{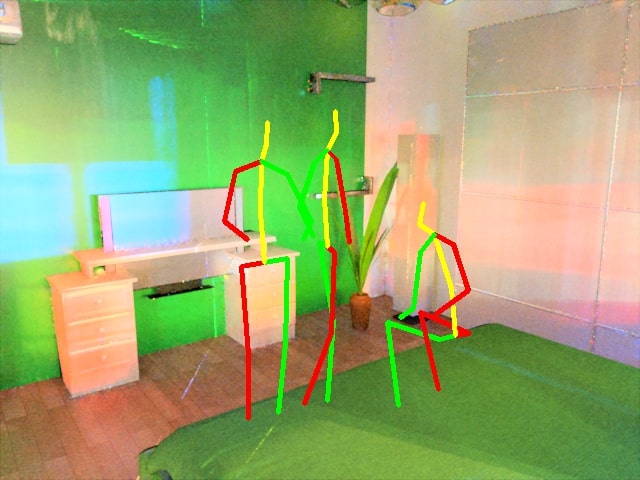} &
		\includegraphics[height = .11\linewidth, width = .15\linewidth]{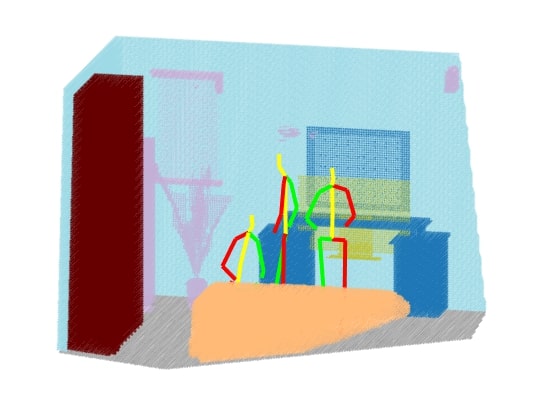} &
		\includegraphics[height = .11\linewidth, width = .15\linewidth]{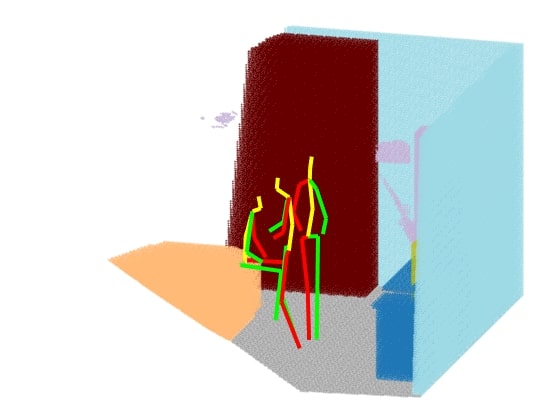} \\
		\includegraphics[height = .11\linewidth, width = .18\linewidth]{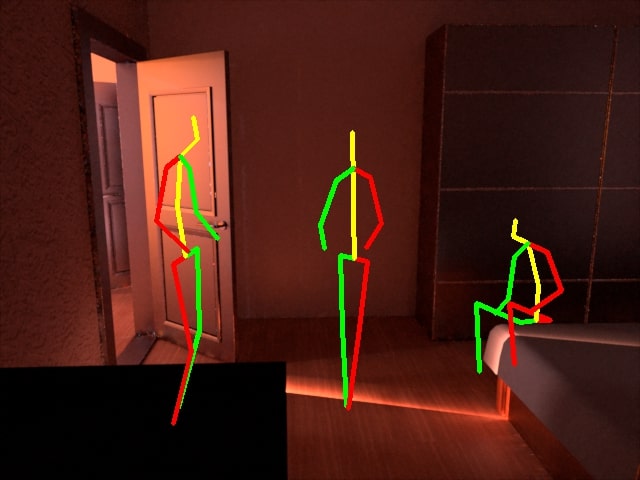} &
    	\includegraphics[height = .11\linewidth, width = .18\linewidth]{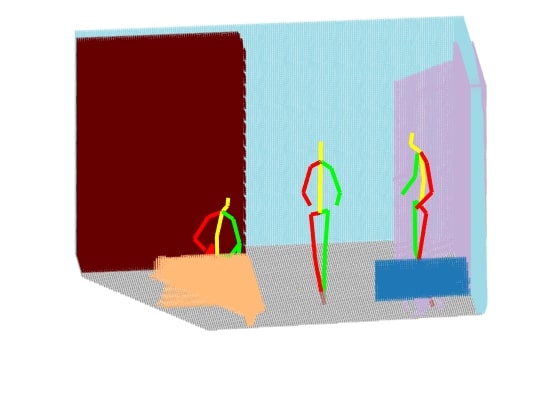} &
    	\includegraphics[height = .11\linewidth, width = .18\linewidth]{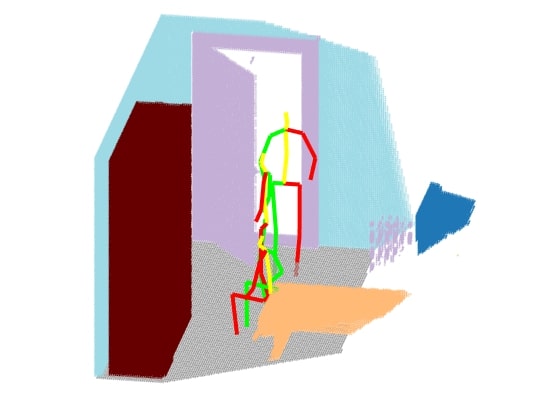} &
		\includegraphics[height = .11\linewidth, width = .15\linewidth]{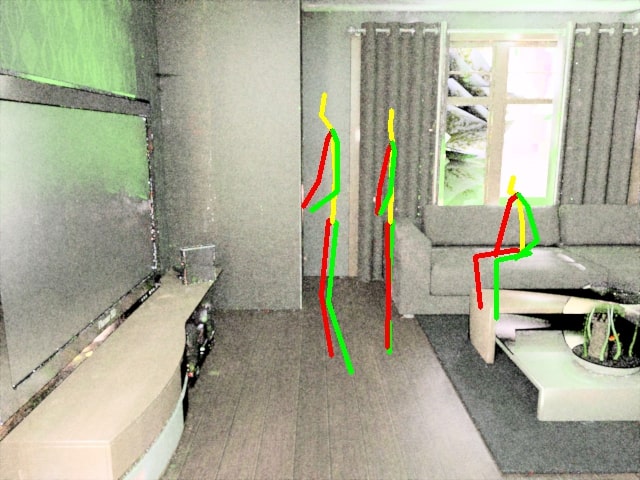} &
		\includegraphics[height = .11\linewidth, width = .15\linewidth]{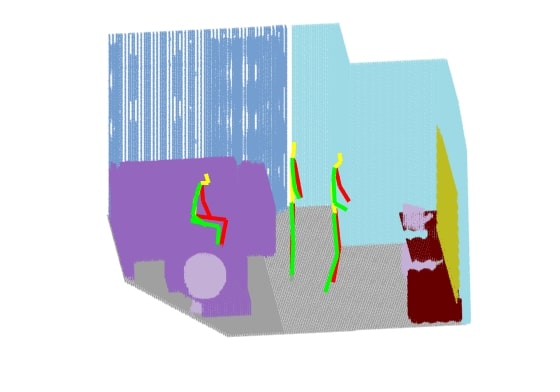} &
		\includegraphics[height = .11\linewidth, width = .15\linewidth]{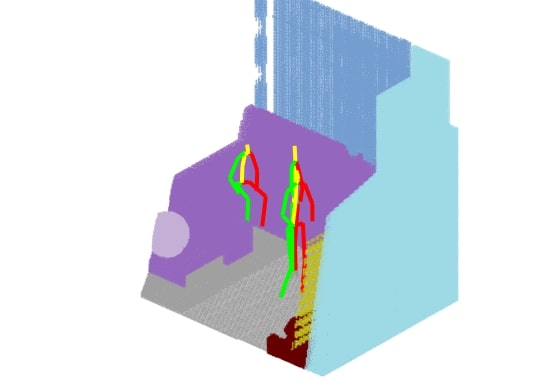}
		
	\end{tabular}
	
	\caption{Generated poses by our pose prediction model. We show generated poses in images (first column) and voxels visualized from two different views (last two columns) for each scene. Poses are generated by our model which takes a single depth image as input.}
	\label{fig:generation}
\end{figure*}

\vspace{-2mm}
\section{Additional Experimental Results}
We show \textbf{synthesized poses} in scene images and voxels in Fig.~\ref{fig:suncg_gt}. More results of \textbf{generated poses} in images and scene voxels are shown in Fig.~\ref{fig:generation}. Note that in this work we use the SUNCG-PBR dataset by Sengupta et al.~\cite{Senguta19}. Despite noise introduced by the rendering process, our pose prediction model is still able to predict plausible poses. 

\end{appendix}

\end{document}